\documentclass{article} % For LaTeX2e
\usepackage{iclr2020_conference,times}

\usepackage{amsmath,amsfonts,bm}

\def\eqref#1{equation~\ref{#1}}
\def\1{\bm{1}}

\DeclareMathAlphabet{\mathsfit}{\encodingdefault}{\sfdefault}{m}{sl}
\SetMathAlphabet{\mathsfit}{bold}{\encodingdefault}{\sfdefault}{bx}{n}

\usepackage{hyperref}
\usepackage{url}

\usepackage{mathtools}

\usepackage{pgfplots}

\usepackage{multirow}

\usepackage{latexsym}
\usepackage{amssymb}
\usepackage{amsmath}
\usepackage{subcaption}
\usepackage[textsize=tiny]{todonotes}
\usepackage{comment}
\usepackage{marginnote}

\usepackage{booktabs}
\usepackage{bm}
\usepackage{colortbl}
\usepackage{color}
\usepackage{xcolor}
\usepackage{soulpos}

\usepackage{algorithm}
\usepackage[noend]{algpseudocode}
\usepackage{amsthm}

\usepackage{array}
\newcolumntype{H}{>{\setbox0=\hbox\bgroup}c<{\egroup}@{}}

\author{Yujia Bao$^{\dagger*}$, Menghua Wu$^{\dagger}$\thanks{Equal
contribution.} , Shiyu
Chang$^{\ddagger}$, Regina Barzilay$^{\dagger}$\\
$^\dagger$Computer Science and Artificial Intelligence Lab, MIT\\
$^\ddagger$MIT-IBM Watson AI LAB, IBM Research\\
\texttt{\{yujia,rmwu,regina\}@csail.mit.edu},
\texttt{\{shiyu.chang\}@ibm.com} \\
}

\iclrfinalcopy % Uncomment for camera-ready version, but NOT for submission.

\newcommand{\set}[1]{\left\{ #1 \right\}}

\newcommand{\norm}[1]{\left\| #1\right\|}
\newcommand{\cD}{\mathcal{D}}
\newcommand{\cT}{\mathcal{T}}
\newcommand{\cY}{\mathcal{Y}}

\newcommand{\cS}{\mathcal{S}}
\newcommand{\cQ}{\mathcal{Q}}
\newcommand{\cP}{\mathcal{P}}
\newcommand{\tx}[1]{\text{#1}}
\newcommand{\f}[2]{\frac{#1}{#2}}

\newcommand{\RR}{\mathbb{R}}
\newcommand{\lto}{\leftarrow}

\newcommand{\beq}{\begin{equation}}
\newcommand{\eeq}{\end{equation}}
\definecolor{blue2}{RGB}{105,200,255}
\definecolor{orange2}{RGB}{185,155,255}
\definecolor{pink2}{RGB}{150,120,230}

\definecolor{bar1}{RGB}{164, 122, 255}
\definecolor{bar2}{RGB}{122, 131, 255}
\definecolor{bar3}{RGB}{99, 203, 255}
\definecolor{bar4}{RGB}{87, 217, 186}

\newcommand{\prob}[1]{\mathrm{P}\left(#1\right)}

\definecolor{grey}{rgb}{0.8,0.8,0.8}
\newcommand{\hlr}[2]{\setlength{\fboxsep}{0.3pt}\colorbox{red!#2}{\rule[-.05\baselineskip]{0pt}{.7\baselineskip}{#1}}}

\newtheorem{theorem}{Theorem}

\newtheorem{innercustomthm}{Theorem}
\newenvironment{customthm}[1]
  {\renewcommand\theinnercustomthm{#1}\innercustomthm}
  {\endinnercustomthm}

\title{Few-shot Text Classification with\\Distributional Signatures}

\date{}

\begin{document}
\maketitle
\begin{abstract}

In this paper, we explore meta-learning for few-shot text classification.
Meta-learning has shown strong performance in computer vision, where low-level patterns are transferable across learning tasks.
However, directly applying this approach to text is challenging--lexical features
highly informative for one task may be insignificant for another.
Thus, rather than learning solely from words, our model also leverages their
distributional signatures, which encode pertinent word occurrence patterns.
Our model is trained within a meta-learning framework to map these signatures
into attention scores, which are then used to weight the lexical representations
of words. We demonstrate that our model consistently outperforms prototypical
networks learned on lexical knowledge~\citep{snell2017prototypical} in both few-shot text classification and relation classification by a significant margin across six benchmark datasets
(20.0\% on average in 1-shot classification).\footnote{Our 
code is available at  \url{https://github.com/YujiaBao/Distributional-Signatures}.}

\end{abstract}

\section{Introduction}

In computer vision, meta-learning has emerged as a promising methodology for learning in a low-resource regime.  Specifically, the goal is to enable an algorithm to expand to new classes for which only a few training instances are available. These models learn to generalize in these low-resource conditions by recreating such training episodes from the data available. Even in the most extreme low-resource scenario--a single training example per class--this approach yields 99.6\% accuracy on the character recognition task~\citep{sung2018learning}.

\begin{figure}[ht]
    \centering
    \begin{minipage}[t]{0.5\linewidth}
      \centering\tiny
      % style={transform shape} applies scaling to each node, including text
      \begin{tikzpicture}[xscale=0.9, yscale=0.9, every node/.style={transform shape}]
    %    \draw (0.15,-1.45) -- (0.05,-1.45) -- (0.05,1.4) -- (0.15,1.4);
    %    \draw (0.3,1.6) -- (0.3,1.7) -- (7.65,1.7) -- (7.65,1.6);
        % zoom drawing
        \node[anchor=south] at (0,0) {
          \includegraphics[height=3.2cm]{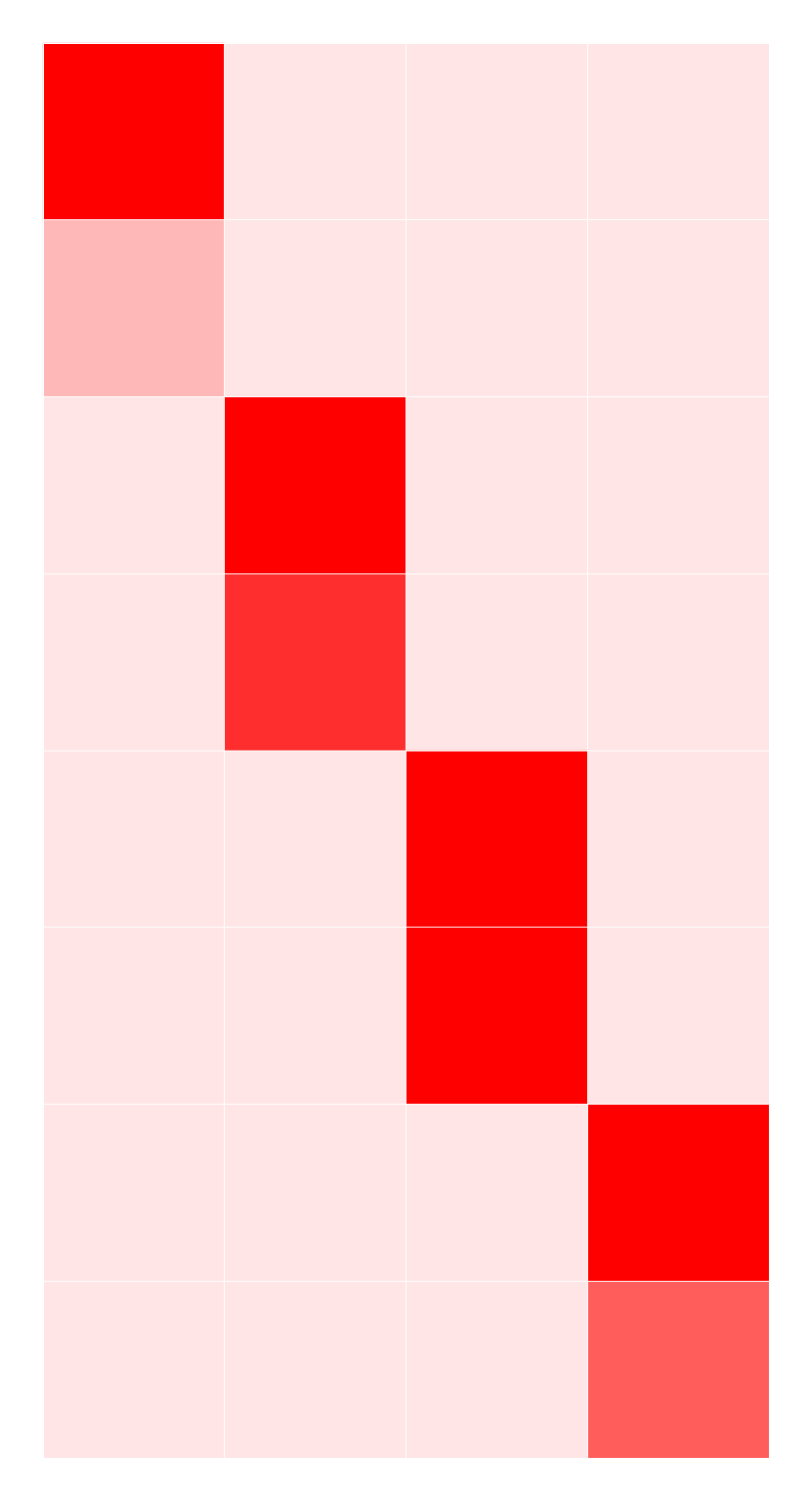}};
        % zoom labels
        \foreach \label[count=\x] in {politics,wellness,entertainment,travel} {
          \node[anchor=west,rotate=90] at (-1+\x*0.4,3.3) {\emph{\label}};
        }
        \foreach \label[count=\y] in {looney,travel,trailer,netflix,cancer,sleep,gop,trump} {
          \node[anchor=east] at (-0.8,-0.05+\y*0.38) {\label};
        }

        % full drawing
        \node[anchor=south] at (3.6,-0.04) {
          \includegraphics[height=4.9cm,width=4cm]{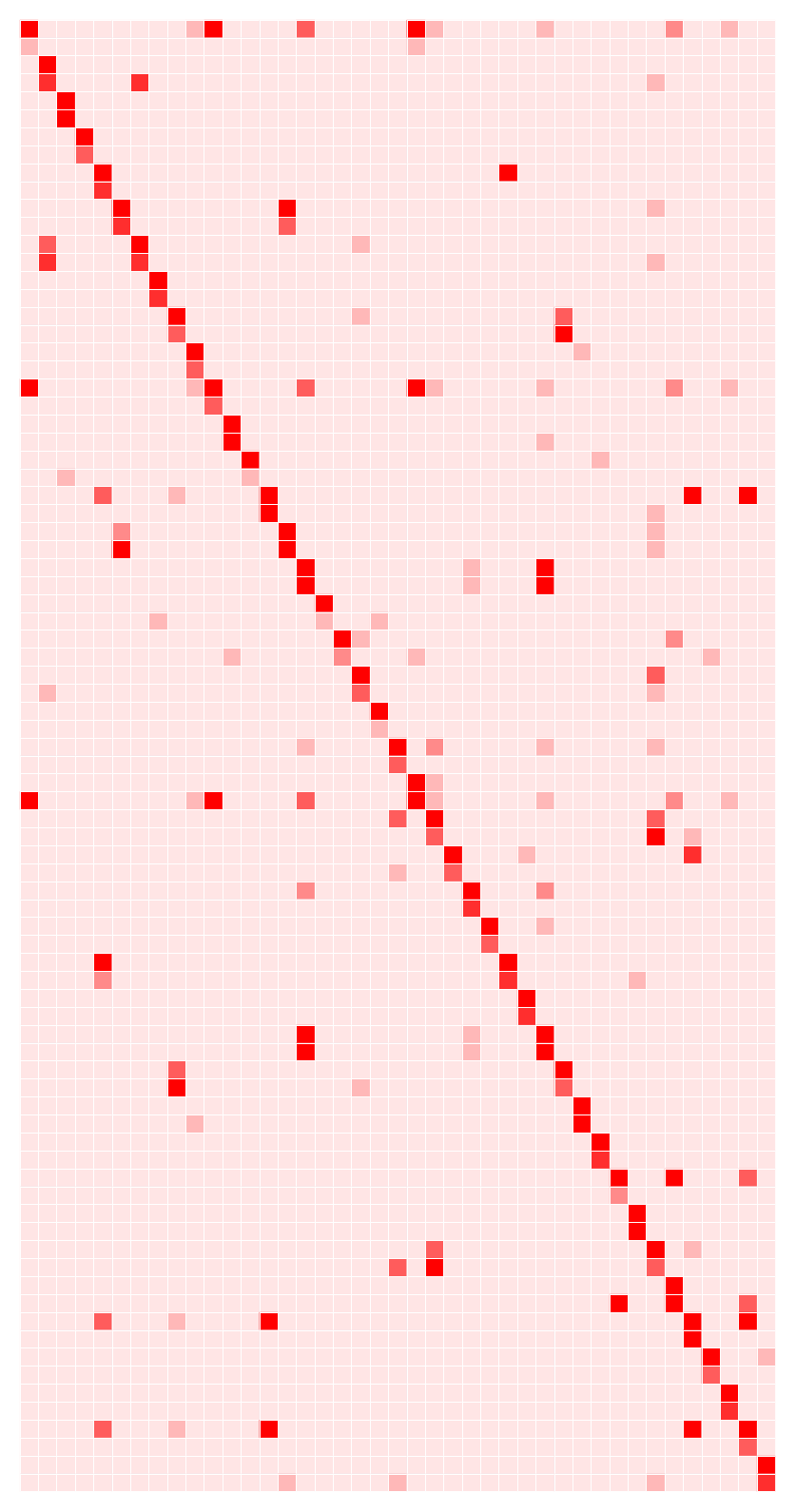}};
        % axis labels
        \node[rotate=90,anchor=east] at (1.5,1.1) {\footnotesize words};
        \node[anchor=south] at (3.6,4.9) {\footnotesize \emph{classes}};

        % selection box
        %% box large
        %\draw (1.7,4.32) rectangle (2.05,4.87);
        \draw (1.7,4.41) rectangle (2.07,4.87);
        %% lines connect
        \draw[dotted,line width=0.8pt] (1.7,4.87) -- (0.78,3.17);
        \draw[dotted,line width=0.8pt] (1.7,4.41) -- (0.78,0.18);
        %% box zoom
      \end{tikzpicture}
    \captionof{figure}{
    Different classes exhibit different word distributions
    in HuffPost headlines.
    We compute the local mutual information (LMI)~\citep{evert2005statistics}
    between words and classes.
    For each class, we include its top 2 LMI-ranked words.
    Darker colors indicate higher LMI.
    }
    \label{fig:intro_heatmap}
    \end{minipage}
    ~
    \begin{minipage}[t]{0.48\linewidth}
        \small
        \renewcommand{\hlr}[2]{\setlength{\fboxsep}{.3pt}\colorbox{red!#2}{\rule[-1mm]{0pt}{3.5mm}{#1}}}
        \centering
        \vspace{-4.45cm}
        \begin{tabular}{p{0.87\linewidth}}
            %\toprule
            \emph{Prototypical network (seen class)}\\
            %\arrayrulecolor{grey}  % choose color
            \midrule
            \hlr{this}{3} \hlr{gorgeous}{10} \hlr{grandma}{100} \hlr{proves}{6} \hlr{beauty}{27} \hlr{has}{5} \hlr{no}{4} \hlr{expiration}{0} \hlr{date}{13}  \\
            \arrayrulecolor{black}
            \vspace{0.01cm}
            \emph{Prototypical network (unseen class)}\\
            \midrule
            \hlr{this}{4} \hlr{gorgeous}{7} \hlr{grandma}{12} \hlr{proves}{3} \hlr{beauty}{17} \hlr{has}{2} \hlr{no}{15} \hlr{expiration}{0} \hlr{date}{46}  \\
            \arrayrulecolor{black}
            \vspace{0.01cm}
            \emph{Our model (unseen class)}\\
            \midrule
            \hlr{this}{0} \hlr{gorgeous}{59} \hlr{grandma}{100} \hlr{proves}{3} \hlr{beauty}{32} \hlr{has}{0} \hlr{no}{1} \hlr{expiration}{0} \hlr{date}{10}  \\
        \end{tabular}

        \captionof{figure}{
        Visualization of word importance
        on example from class \emph{fifty} in HuffPost headlines.
        Top: \emph{fifty} is seen during meta-training;
        prototypical network~\citep{snell2017prototypical}
        finds important words.
        Middle: \emph{fifty} is unavailable during meta-training;
        it fails to generalize.
        Bottom: Our model identifies key words for unseen classes.
    }
    \label{fig:intro}
    \end{minipage}
    \vspace{-5mm}
\end{figure}

Given this strong empirical performance, we are interested in employing meta-learning frameworks in NLP. The challenge, however, is the degree of transferability of the underlying representation learned across different classes. In computer vision, low-level patterns (such as edges) and their corresponding representations can be shared across tasks. However, the situation is different for language data where most tasks operate at the lexical level. Words that are highly informative for one task may not be relevant for other tasks. Consider, for example, the corpus of HuffPost headlines, categorized into 41 classes.  Figure~\ref{fig:intro_heatmap} shows that words highly salient for one class do not play a significant role in classifying others. Not surprisingly, when meta-learning is applied directly on lexical inputs, its performance drops below a simple nearest neighbor classifier.
% \footnote{See Section~\ref{sec:results} for details.}
The inability of a traditional meta-learner to zoom-in on important features is
further illustrated in Figure~\ref{fig:intro}:
when considering the target class \emph{fifty}
(lifestyle for middle-aged),
% \footnote{The class ``\emph{fifty}'' describes lifestyle for the middle-aged.}
the standard prototypical network~\citep{snell2017prototypical} attends to uninformative words like ``date,'' while downplaying highly predictive words such as ``grandma.''

%\begin{table}[t]
%  \scriptsize
%  \centering
%  \begin{tabular}{lllll}
%    \toprule
%    Graphics & Baseball & Electronics & Guns     & Atheism  \\
%    \midrule
%    \midrule
%    image        &baseball     &circuit      &gun          &god          \\
%    graphics     &game         &wire         &guns         &atheism      \\
%    jpeg         &year         &ground       &fbi          &atheists     \\
%    images       &team         &wiring       &firearms     &keith        \\
%    gif          &players      &voltage      &atf          &livesey      \\
%    format       &games        &battery      &batf         &morality     \\
%    file         &hit          &copy         &weapons      &religion     \\
%    3d           &braves       &amp          &people       &moral        \\
%    ftp          &runs         &electronics  &waco         &islamic      \\
%    color        &pitcher      &audio        &cdt          &say          \\
%    \bottomrule
%  \end{tabular}
%  \caption{Top 10 unigrams ranked by local mutual
%  information~\citep{evert2005statistics} in 5 classes from 20
%  Newsgroups.}\label{tab:20news}
%\end{table}

In this paper we demonstrate that despite these variations,
we can effectively transfer representations across classes and thereby enable learning in a low-resource regime.
%Instead of directly computing on word representations,
%our method considers the \emph{distributional signatures} of words.
%These signatures characterize underlying word distributions,
%which exhibit consistent behavior across classification tasks.
Instead of directly considering words,
our method utilizes their \emph{distributional signatures},
characteristics of the underlying word distributions,
%Instead of directly considering words,
%by computing on their lexical representations,
%our method utilizes their \emph{distributional signatures}.
%These distributional signatures
%depict the characteristics of the underlying word distributions,
which exhibit consistent behaviour across classification tasks.
Within the meta-learning framework,
these signatures enable us to transfer attention across tasks,
which can consequently be used to weight the lexical representations of words.
One broadly used example of such distributional signatures is tf-idf weighting, which explicitly specifies word importance in terms of its frequency in a document collection, and its skewness within a specific document.

Building on this idea, we would like to learn to utilize distributional signatures in the context of cross-class transfer. In addition to word frequency, we assess word importance with respect to a specific class. This latter relation cannot be reliably estimated of the target class due to the scarcity of labeled data. However, we can obtain a noisy estimate of this indicator by utilizing the few provided training examples for the target class, and then further refine this approximation within the meta-learning framework.
We note that while the representational power of
distributional signatures
is weaker than that of their lexical counterparts,
meta knowledge built on distributional signatures are better able to generalize.

Our model consists of two components.
The first is an \emph{attention generator},
which translates distributional signatures
into attention scores that reflect word importance for classification.
Informed by the attention generator's output,
our second component, a \emph{ridge regressor},
quickly learns to make predictions
after seeing only a few training examples.
The attention generator is shared across all episodes,
while the ridge regressor is trained from scratch for each individual episode.
The latter's prediction loss provides supervision
for the attention generator.
Theoretically,
we show that the attention generator
is robust to word-substitution perturbations.

We evaluate our model on five standard text classification datasets~\citep{lang1995newsweeder, lewis2004rcv1, Lewis1997Reuters21578TC, he2016ups, huffpost} and one relation classification dataset~\citep{han2018fewrel}. Experimental results demonstrate that our model
delivers significant performance gains over all baselines.
For instance, our model outperforms prototypical networks by 20.6\% on average
in one-shot text classification and 17.3\% in one-shot relation classification.
In addition, both qualitative and quantitative analyses confirm that our model
generates high-quality attention for unseen classes.

\section{Related Work}

\noindent\textbf{Meta-learning }
Meta-learning has been shown to be highly effective
in computer vision,
where low-level features are transferable across classes.
Existing approaches include learning
a metric space over input features~\citep{koch2015siamese,vinyals2016matching,
snell2017prototypical,sung2018learning},
developing a prior over the optimization procedure~\citep{ravi2016optimization,
finn2017model,nichol2018reptile,antoniou2018train},
and exploiting the relations between classes~\citep{garcia2017few}.
These methods have been adapted with some success
to specific applications in NLP,
including machine translation~\citep{gu2018meta},
text classification~\citep{yu2018diverse, geng2019few,
guo2018multi,jiang2019attentive}
and relation classification~\citep{han2018fewrel}.
Such models primarily build meta-knowledge
on lexical representations.

However, as our experiments show,
there exist innate differences
in transferable knowledge between image data and language data,
and lexicon-aware meta-learners
fail to generalize on standard multi-class classification datasets.
% Furthermore, \citep{chen2018a} assumes a resource-rich setting for multi-task learning across domains (each with 1000+ examples), while our model is designed for processing new classes with limited examples. \citep{yu2018diverse} adapts a single task (sentiment classification) across domains, rather than solving new classification tasks.

\begin{figure*}[t]
  \centering
  \footnotesize
  \begin{tikzpicture}[xscale=0.85,yscale=0.85]
  %\fontfamily{qag}  % comment out to mix times + qag

    \newcommand{\drawstack}[3][blue2]{
      \begin{scope}[minimum width=0.8cm,minimum height=0.38cm,rounded corners=2pt]
        % \node[fill=#1!30,draw=black!25] at (#2-0.125,#3+0.125) {};
        \node[fill=#1!40,draw,draw opacity=0.4] at (#2-0.0625,#3+0.0625) {};
        \node[fill=#1!50,draw=black] at (#2,#3) {};
      \end{scope}
    }

    %%%%%%% meta-training
    \node[anchor=west] at (2.75,-0.35) {a) meta-training};
    % y train bounding box
    \fill[fill=pink2!10] (0.25,2.5) rectangle (8,4);
    % all y train
    \node[anchor=west] at (0.4,3.675) {training data: all examples from $\mathcal{Y}^\tx{train}$};
    \foreach \x/\label in {1/politics,2/tech,3/travel,4/media,5.5/sports,6.5/beauty,7.5/taste} {
      \fontfamily{qag}
      %\tiny
      \fontsize{5}{4}\selectfont
      \drawstack[white]{\x}{3}
      \node at (\x,3) {\label};
    }
    \node at (4.75,3) {\dots};
    % target bounding box
    \fill[fill=pink2!10] (0.25,0) rectangle (3.75,1.625);
    % all target
    \node[anchor=north] at (2,0.75) {\tiny examples from which the};
    \node[anchor=north] at (2,0.5) {\tiny \emph{support} and \emph{query} are drawn};
    \foreach \x/\label in {1/politics,2/media,3/sports} {
      \fontfamily{qag}
      %\tiny
      \fontsize{5}{4}\selectfont
      \drawstack[orange2]{\x}{1}
      \node at (\x,1) {\label};
    }
    % source bounding box
    \fill[pink2!10] (4,0) rectangle (8,1.625);
    % all source

    \node at (6,0.45) {\fontsize{8}{4}\selectfont \emph{source pool}};
    \node at (6,0.15) {\tiny(our extension)};
    \foreach \x/\label in {4.75/tech,5.75/travel,7.25/beauty} {
      \fontfamily{qag}
      %\tiny
      \fontsize{5}{4}\selectfont
      \drawstack[blue2]{\x}{1}
      \node at (\x,1) {\label};
    }
    \node at (6.5,1) {\dots};
    % edges
    \begin{scope}
    \fontsize{8}{4}\selectfont
    \path[->]
      (0.75,2.5)    edge node[pos=.5,anchor=west] {1. sample $N$ classes} (0.75,1.625)
      (4.5,2.5) edge    node[pos=.4,anchor=west] {2. retrieve examples from}
                      node[pos=.7,anchor=west] {remaining classes} (4.5,1.625);
    \end{scope}

    %%%%%%% meta-testing
    \draw[dashed] (8.25,0) -- (8.25,4);
    \begin{scope}% [xshift=-8.25cm, yshift=-4.75cm]  % edit from emnlp submission
    \node[anchor=west] at (11,-0.35) {b) meta-testing};
    \begin{scope}[xshift=8.25cm]
      % y train bounding box
      \fill[fill=pink2!10] (0.25,2.5) rectangle (8.125,4);
      % all y train
      \node[anchor=west] at (0.4,3.675) {training data
      is used as \emph{source pool} {\tiny(our extension)}};
      \foreach \x/\label in {1/politics,2/tech,3/travel,4/media,5.5/sports,6.5/beauty,7.5/taste} {
        \fontfamily{qag}
        %\tiny
        \fontsize{5}{4}\selectfont
        \drawstack[blue2]{\x}{3}
        \node at (\x,3) {\label};
      }
      \node at (4.75,3) {\dots};
    \end{scope}
    % y target bounding box
    \fill[pink2!10] (8.5,0) rectangle (11.75,2.25);
    % all y target
    \node[anchor=west] at (8.65,2) {testing data $\mathcal{Y}^\tx{test}$};
    \foreach \x/\label in {9.25/style,10.25/culture,11.25/fifty} {
      \fontfamily{qag}
      %\tiny
      \fontsize{5}{4}\selectfont
      \drawstack[white]{\x}{1.375}
      \node at (\x,1.375) {\label};
    }
    \foreach \x/\label in {9.25/taste,10.25/arts,11.25/money} {
      \fontfamily{qag}
      %\tiny
      \fontsize{5}{4}\selectfont
      \drawstack[white]{\x}{0.375}
      \node at (\x,0.375) {\label};
    }
    \node at (10.25,1) {\scriptsize\vdots};
    % y target sampled
    %\begin{scope}[xshift=12.75cm,yshift=0.5cm]
    \begin{scope}[xshift=12.75cm]
      % y target sampled bounding box
      %\fill[fill=pink2!10] (0.25,0) rectangle (3.5,1.625);
      \fill[fill=pink2!10] (0.25,0) rectangle (3.625,2.25);
      % all target
      \node[anchor=north] at (2,1) {\tiny examples from which the};
      \node[anchor=north] at (2,0.75) {\tiny \emph{support} and \emph{query} are drawn};
      \foreach \x/\label in {1/style,2/money,3/culture} {
        \fontfamily{qag}
        %\tiny
        \fontsize{5}{4}\selectfont
        \drawstack[orange2]{\x}{1.5}
        \node at (\x,1.5) {\label};
      }
    \end{scope}
    % arrow
    \fontsize{7}{4}\selectfont
    \path[->]
      (11.75,1) edge    node[pos=0.5,above] {sample}
                        node[pos=0.5,below] {$N$ classes} (13,1);
    \end{scope}

  \end{tikzpicture}
  \caption{Episode generation.
    a) Meta-training:
    First, sample $N$ classes from $\cY^\tx{train}$.
    Then, sample the support set and the query set from the $N$ classes.
    We use examples from the remaining classes to form the source pool.
    b)
    Meta-testing:
    Select $N$ new classes from $\cY^\tx{test}$ and
    sample the support set and the query set from these $N$ classes.
    We use all examples from $\cY^\tx{train}$ to form the source pool.
  }\label{fig:gen_episode}
\vspace{-5mm}
\end{figure*}
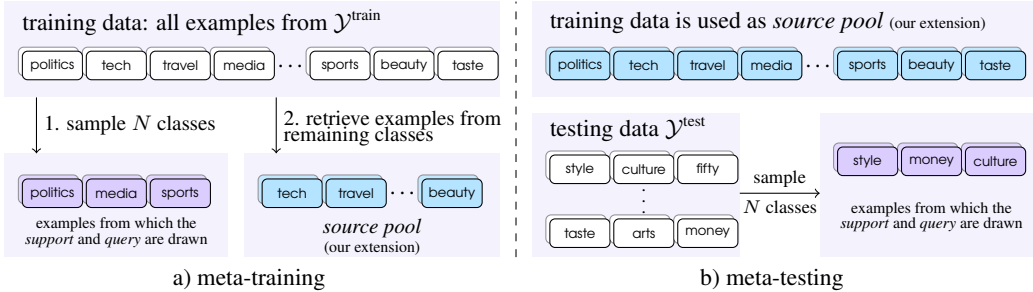

In this work, we observe that
even though salient features in text may not be transferable,
their \emph{distributional} behaviors are alike.
% CUT IF NEED SPACE 
Thus, we focus on learning the
connection between word importance and
distributional signatures.
As a result,
our model can reliably identify important features from novel classes.

\noindent\textbf{Transfer learning }
Our work is also closely related to transfer learning:
we assume access to
a large number of labeled examples from \emph{source} classes,
and we would like to identify word importance for the 
\emph{target} classification task.
Current approaches transfer knowledge from the source to the target
by either fine-tuning a pre-trained encoder~\citep{howard2018universal,Peters:2018,
radford2018improving,bertinetto2018metalearning},
or multi-task learning with a shared encoder~\citep{collobert2008unified,
liu2015representation,luong2015multi,strubell2018linguistically}.
Recently, \citet{bao2018deriving} also successfully transferred
task-specific attention through human rationales.

In contrast to these methods,
where the transfer mechanism is pre-designed,
we learn to transfer based on
the performance of
downstream tasks.
Specifically, we utilize distributional statistics
to transfer attention across tasks.
We note that while \citet{wei2017learning} and \citet{sun2018meta}
also learn transfer mechanisms for image recognition,
their methods do not directly apply to NLP.

\section{Background}

\begin{figure}[t]
  \centering
  \begin{tikzpicture}[xscale=0.65,yscale=0.8]

    % scale x and y from \NN coordinates
    \pgfmathsetmacro{\xs}{1.7}
    \pgfmathsetmacro{\ys}{1.5}

    % \drawgrid{-8}{6}{-2}{4}

    % traditional meta-learning episode box
    \draw[dashed] (-1.25,-1.1) -- (-1.25,1.75);
    \node[anchor=north west,text width=4cm] at (-9.8,-1.15) {\footnotesize a) traditional episode};

    %%%%% source pool
    % draw background + label
    \fill[pink2!10] (-1,-1.125) rectangle (5.25,1.75);
    \node at (2.125,-0.75) {\footnotesize source pool};
    \node[anchor=north west] at (0,-1.15) {\footnotesize b) our extension};
    % draw data
    \foreach \x/\label in {0/media,1.5/impact,2.5/tech} {
      \fontfamily{qag}\tiny
      \foreach \y in {0,0.85} {
        \node[draw=none,fill=blue2!50,minimum width=0.9cm,minimum height=0.5cm,rounded corners=2pt] at (\x*\xs,\y*\ys) {\label};
      }
    }
    \foreach \x in {0,1.5,2.5} {
        \node[scale=0.6] at (\x*\xs,0+0.5*\ys) {\textbf\vdots};
    }
    \foreach \y in {0,0.85} {
      \node at (0.75*\xs,\y*\ys) {\dots};}

    %%%%% support set
    % draw background + label
    \begin{scope}[xshift=-18.5cm]
      \fill[pink2!10] (5.5,-1.125) rectangle (11,1.75);
      % labels
      \node at (8.25,-0.4) {\footnotesize support set};
      \node at (8.25,-0.8) {\scriptsize ($K\times N$ examples)};
      % draw data
      \foreach \x/\label in {0/religion,1/beauty,2/games} {
        \fontfamily{qag}\tiny
        \foreach \y in {0.55555} {
          \node[draw=none,fill=orange2!50,minimum width=0.9cm,minimum height=0.5cm,rounded corners=2pt] at (6.5+\x*\xs,\y*\ys) {\label};
        }
      }
    \end{scope}

    %%%%% query set
    \begin{scope}[xshift=-12.5cm]
      \fill[pink2!10] (5.5,-1.125) rectangle (11,1.75);
      % labels
      \node at (8.25,-0.4) {\footnotesize query set};
      \node at (8.25,-0.8) {\scriptsize ($L\times N$ examples)};
      % draw data
      \foreach \x/\label in {0/religion,1/beauty,2/games} {
        \fontfamily{qag}\tiny
        \foreach \y in {0.75,0.25} {
          \node[draw=none,fill=orange2!50,minimum width=0.9cm,minimum height=0.5cm,rounded corners=2pt] at (6.5+\x*\xs,\y*\ys) {\label};
        }
      }
    \end{scope}

  \end{tikzpicture}
  \caption{Single episode with $N=3$, $K=1$, $L=2$.
    Rectangles denote input examples.
    The text inside corresponds to their labels.
    An episode contains a support set, a query set, and a source pool.
  }\label{fig:episode}
\vspace{-5mm}
\end{figure}
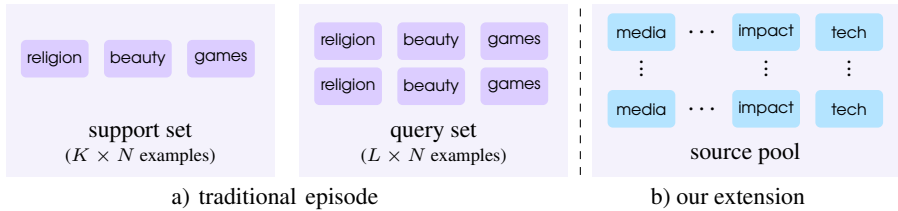

In this section,
we first summarize the standard meta-learning framework
for few-shot classification
and describe the terminology~\citep{vinyals2016matching}.
Next, we introduce our extensions to the framework.
Figures~\ref{fig:gen_episode} and~\ref{fig:episode}
graphically illustrate our framework.

\textbf{Problem statement }
Suppose we are given labeled examples
from a set of classes $\cY^\tx{train}$.
Our goal is to develop a model that
acquires knowledge from these training data,
so that we can make predictions over new
(but related) classes, for which we only have a few annotations.
These new classes belong to a set of classes $\cY^\tx{test}$,
disjoint from $\cY^\tx{train}$.

\textbf{Meta-training }
In meta-learning,
we emulate the above testing scenario during meta-training
so our model learns to quickly learn from a few annotations.
To create a single \emph{training episode},
we first sample $N$ classes from $\cY^\tx{train}$.
For each of these $N$ classes,
we sample $K$ examples as our training data
and $L$ examples as our testing data.
We update our model based on
loss over these testing data.
Figure~\ref{fig:episode}a shows an example of an episode.
We repeat this procedure
to increase the number of training episodes,
each of which is constructed over its own set of $N$ classes.
In literature,
the training data of one episode
is commonly denoted as the \emph{support set},
while the corresponding testing data
is known as the \emph{query set}.
Given the support set, we refer to
the task of making predictions over
the query set as \emph{$N$-way $K$-shot classification}.

\textbf{Meta-testing}
After we have finished meta-training,
we apply the same episode-based mechanism to test
whether our model can indeed adapt quickly to new classes.
To create a \emph{testing episode},
we first sample $N$ new classes from $\cY^\tx{test}$.
Then we sample the support set and the query set from the $N$ classes.
We evaluate the average performance
on the query set across all testing episodes.

\textbf{Our extension }
We observe that even though all examples from $\cY^\tx{train}$
are accessible throughout meta-training,
the standard meta-learning framework~\citep{vinyals2016matching}
only learns from small subsets of these data per training episode.
In contrast, our model leverages
distributional statistics over all training examples
for more robust inference.
To accommodate this adjustment,
we augment each episode with a \emph{source pool}
(Figure~\ref{fig:episode}b).
During meta-training (Figure~\ref{fig:gen_episode}a),
this source pool includes all examples
from training classes not selected for the particular episode.
During meta-testing (Figure~\ref{fig:gen_episode}b),
this source pool includes all training examples.
%Figure~\ref{fig:episode}.b shows our extensions
%to the traditional episode.

\section{Method}
\textbf{Overview }
Our goal is to improve few-shot classification performance
by learning high-quality attention
from the distributional signatures of the inputs.
%\yujia{the current is too low level}
Given a particular episode,
we extract relevant statistics from the source pool and the support set.
%we estimate general word importance
%by computing the term frequency over the source pool,
%and estimate class-specific word relevancy through the support set.
Since these statistics only roughly approximate
word importance for classification,
we utilize an \emph{attention generator} to translate them
into high-quality attention that operates over words.
This generated attention provides guidance for the downstream predictor,
a \emph{ridge regressor},
to quickly learn
from a few labeled examples.\footnote{
Note this generated attention can be applied to \emph{any}
downstream predictor~\citep{lee2019meta, snell2017prototypical}.
Details from these experiments may be found in the Appendix~\ref{app:combine}.
}

\begin{figure*}[t]
  \centering
  \includegraphics[width=\linewidth]{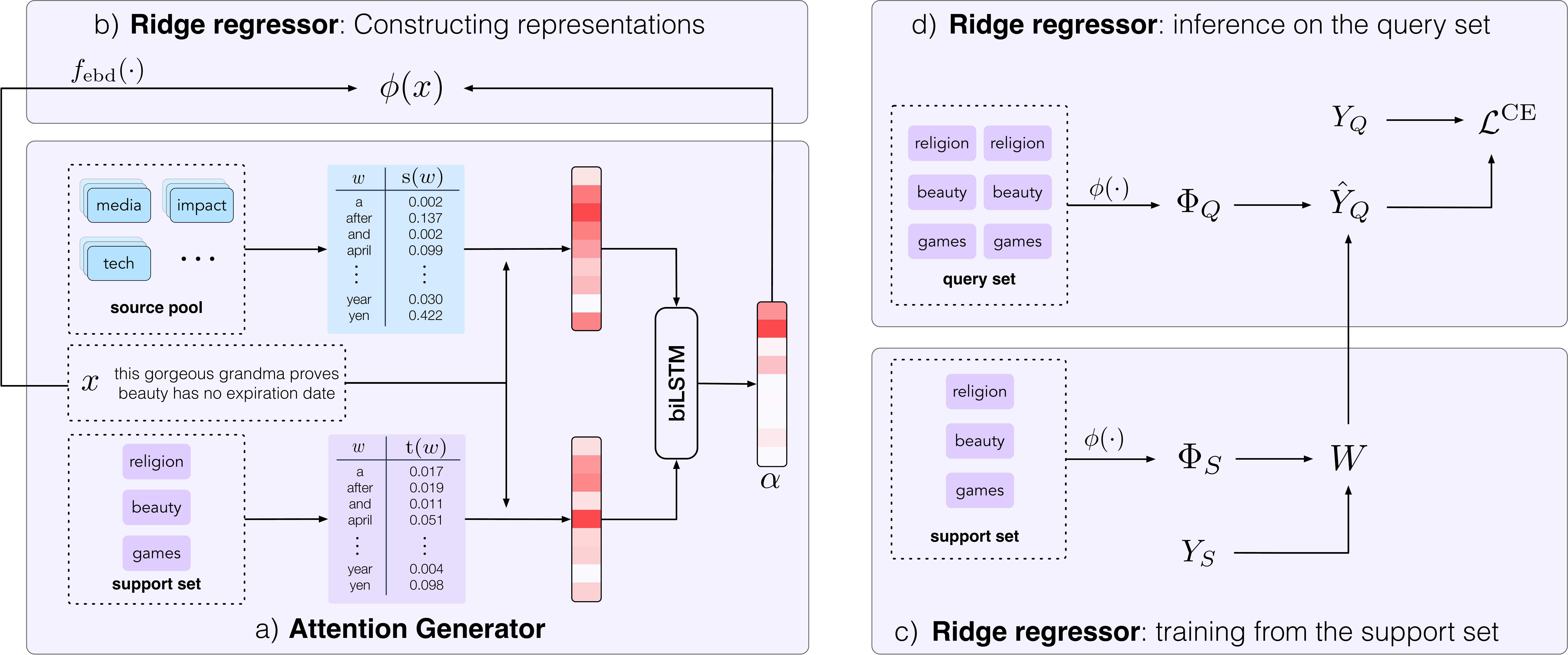}
  \caption{Illustration of our model
    for an episode with $N=3$, $K=1$, $L=2$.
    The attention generator translates
    the distributional signatures
    from the source pool and the support set into
    an attention $\alpha$ for each input example $x$ (5a).
    The ridge regressor utilizes
    the generated attention to weight the lexical representations (5b).
    It then learns from the support set (5c)
    and makes predictions over the query set (5d).
  }\label{fig:overview}
  \vspace{-5mm}
\end{figure*}

We note that the attention generator is optimized over all training episodes,
while the ridge regressor is trained from scratch for each episode.
Figure~\ref{fig:overview} illustrates
the two components of our model.

\begin{itemize}
  \item \textbf{Attention generator:}
    This module generates class-specific attention
    by combining the distributional statistics
    of the source pool and the support set (Figure~\ref{fig:overview}a).
    The generated attention provides the ridge regressor
    an inductive bias on the word importance.
    We train this module
    based on feedback from the ridge regressor (Section~\ref{sec:tag}).
  \item \textbf{Ridge regressor:}
    For each episode,
    this module constructs lexical representations
    using the attention
    derived from distributional signatures (Figure~\ref{fig:overview}b).
    The goal of this module is to make predictions over the query set,
    after learning from the support set (Figures~\ref{fig:overview}c and
    \ref{fig:overview}d).
    Its prediction loss is end-to-end differentiable with respect to
    the attention generator which leads to efficient training
    (Section~\ref{sec:r2d2}).
\end{itemize}

In our theoretical analysis,
we show that the attention generator's outputs
are invariant to word-substitution perturbations (Section 4.3).

\subsection{Attention generator}
\label{sec:tag}
The goal of the attention generator is to assess word importance
from the distributional signatures of each input example.
There are many choices of distributional signatures.
Among them, we focus on functions of unigram statistics,
which are provably robust to word-substitution perturbations
(Section 4.3).
We utilize the large source pool to
inform the model of \emph{general} word importance
and leverage the small support set to
estimate \emph{class-specific} word importance.
The generated attention will be used later to construct
the input representation for downstream classification.

It is well-documented in literature
that words which appear frequently
are unlikely to be informative~\citep{sparck1972statistical}.
Thus, we would like to downweigh frequent words
and upweight rare words.
To measure \emph{general word importance},
we select an established approach by \citet{arora2016simple}:
\begin{equation}\label{eq:s}
  \mathrm{s}(x_i) \coloneqq \f{\varepsilon}{\varepsilon + \prob{x_i}}
\end{equation}
where $\varepsilon = 10^{-3}$,
$x_i$ is the $i^\tx{th}$ word of input example $x$, and
$\prob{x_i}$ is the unigram likelihood of $x_i$
over the source pool.
%In our theoretical analysis, we show that
%as long as $\mathrm{s}(x_i)$ is some function of $\prob{x_i}$,
%the attention generator will produce 

% REMOVED FOR SPACE
%\footnote{\citet{arora2016simple} shows that
%$\mathrm{s}(\cdot)$ provides
%the maximum a posteriori estimate
%of the discourse vector governing the input.}

On the other hand,
words that are discriminative in the support set
are likely to be discriminative in the query set.
Thus, we define the following statistic to reflect
\emph{class-specific word importance}:
\begin{equation}\label{eq:t}
  \mathrm{t}(x_i) \coloneqq {\mathcal{H}(\prob{y \mid x_i})}^{-1}
\end{equation}
where the conditional likelihood $\prob{y \mid x_i}$
is estimated over the support set using
a regularized linear classifier\footnote{
See Appendix~\ref{app:softmax} for details.}
and $\mathcal{H}(\cdot)$ is the entropy operator.
We note that
$\mathrm{t}(\cdot)$ measures
the uncertainty of the class label $y$, given the word.
Thus, words that exhibit a skewed distribution will be highly weighted.

Directly applying these statistics
may not result in good performance for two reasons:
1) the two statistics contain complementary information,
and it is unclear how to combine them;
and 2) these statistics are noisy approximations to
word importance for classification.
To bridge this gap,
we concatenate these signatures and
employ a bi-directional LSTM~\citep{hochreiter1997long}
to fuse the information across the input:
$h = \mathrm{biLSTM}([\mathrm{s}(x); \mathrm{t}(x)])$.
Finally, we use dot-product attention
to predict the attention score $\alpha_i$ of word $x_i$:
\begin{equation}\label{eq:alpha}
  \alpha_i \coloneqq \f{\exp \left( v^T h_i \right) }{\sum_j \exp
  \left( v^T h_j\right) }
\end{equation}
where $h_i$ is the output of the $\mathrm{biLSTM}$ at position~$i$
and $v$ is a learnable vector.

\subsection{Ridge regressor}
\label{sec:r2d2}

Informed by the attention generator,
the ridge regressor quickly learns to make predictions
after seeing a few examples.
First, for each example in a given episode,
we construct a lexical representation
that focuses on important words,
as indicated by attention scores.
Next, given these lexical representations,
we train the ridge regressor on the support set from scratch.
Finally, we make predictions over the query set
and use the loss to
teach the attention generator to produce better attention.
%Finally, the regressor's prediction loss on the query set
%teaches the attention generator to produce better attention.

\textbf{Constructing representations }
Given that different words exhibit varying levels of importance
towards classification,
we construct lexical representations that favor pertinent words.
Specifically, we define the representation of example $x$ as
\begin{equation}
  \phi(x) \coloneqq \sum_i \alpha_i \cdot f_\tx{ebd}(x_i)
  \label{eq:phi}
\end{equation}
where $f_\tx{ebd}(\cdot)$ is a pre-trained embedding function that
maps a word into $\RR^E$.

\textbf{Training from the support set }
Given an $N$-way $K$-shot classification task,
let $\Phi_S \in \RR^{NK\times E}$
be the representation of the support set, obtained from $\phi(\cdot)$,
and $Y_S \in \RR^{NK\times N}$ be the one-hot labels.
We adopt ridge regression~\citep{bertinetto2018metalearning}
to fit the labeled support set for the following reasons:
1) ridge regression admits a closed-form solution
that enables end-to-end differentiation through the model, and
2) with proper regularization,
ridge regression reduces over-fitting on the small support set.
%3) the closed-form solution is computationally insensitive
%to the embedding dimension of the data.
Specifically, we minimize regularized squared loss
\begin{equation}\label{eq:rrloss}
  \mathcal{L}^{RR}(W) \coloneqq \norm{\Phi_S W - Y_S}_F^2
  + \lambda \norm{W}_F^2
\end{equation}
over the weight matrix $W \in \RR^{E \times N}$.
Here $\norm{\cdot}_F$ denotes the Frobenius norm, and
$\lambda > 0$ controls the conditioning of the learned transformation $W$.
The closed-form solution can be obtained as
\begin{equation}
  \label{eq:w}
  W = \Phi_S^T (\Phi_S \Phi_S^T + \lambda I)^{-1} Y_S
\end{equation}
where $I$ is the identity matrix.

\textbf{Inference on the query set }
Let $\Phi_Q$ denote
the representation of the query set.
Although we optimized for a regression objective in Eq~\eqref{eq:rrloss},
the learned transformation has been shown to work well
in few-shot classification
after a calibration step~\citep{bertinetto2018metalearning}, as
\begin{equation}\label{eq:calibration}
  \hat{Y}_Q = a \Phi_Q W + b
\end{equation}
where $a \in \RR^{+}$ and $b \in \RR$ are meta-parameters
learned through meta-training.
Finally, we apply a softmax over $\hat{Y}_Q$ to
obtain the predicted probabilities $\hat{P}_Q$.
Note that
this calibration only adjusts the temperature and scale of the softmax;
its mode remains unchanged.
During meta-training,
we compute the cross-entropy loss $\mathcal{L}^\tx{CE}$
between $\hat{P}_Q$ and the labels over the query set.
Since both $\Phi_S$ and $\Phi_Q$ depend on $\phi(\cdot)$,
$\mathcal{L}^\tx{CE}$ provides supervision for the attention generator.

\subsection{Theoretical analysis}

Working with distributional signatures
brings certified robustness against input perturbations.
Formally, let $(\cP, \cS, \cQ)$ be a single episode,
where $\cP$ is the source pool, $\cS$ is the support set and $\cQ$ is the query set.
For any input text $x \in \cS \cup \cQ$,
the attention generator produces word-level importance
given the support set and the source pool:
\[
  \alpha = \text{AttGen}(x \mid \cS, \cP).
\]
Now consider a word-substitution perturbation
$\sigma$ that replaces a word $w$ by $\sigma(w)$.\footnote{
We view this $\sigma$ as a mapping from the vocabulary set $V$ to
itself.}
If we arbitrarily swap words, we may encounter nonsensical outputs,
as important words may be substituted by common words,
like ``a'' or ``the.''
Thus, we consider perturbations that preserve the unigram probabilities
of words, estimated over the source pool: $\prob{w} = \prob{\sigma(w)}$ for all $w \in V$.
In this way, important words (for one class) are mapped to
similarly important words (perhaps for another class).
We use $\tilde{\cS}$ and $\tilde{\cQ}$ to denote the support and query sets
after applying $\sigma$ word by word.

\begin{theorem}
  Assume $\sigma: V \to V$ satisfies $\prob{w} = \prob{\sigma(w)}$ for all $w$.
  If $\sigma$ is a bijection,
  then for any input text $x \in \cS \cup \cQ$ and
  its perturbation $\tilde{x} \in \tilde{\cS} \cup \tilde{\cQ}$,
  the outputs of the attention generator are the same:
  \[
    \text{AttGen}(x\mid \cS, \cP) =
    \text{AttGen}(\tilde{x} \mid \tilde{\cS}, \cP).
  \]
\end{theorem}

The idea behind this proof is to show that
general word importance $s$ and class-specific word importance $t$
are invariant to such perturbations.
Details may be found in Appendix~\ref{app:thm}.

Alternatively, one can view this word-substitution perturbation
as a change to the input distribution.
Let $\prob{x \mid y}$ be the conditional probability of
the input $x$ given the class label $y$.
Assuming that $\sigma$ is a bijection,
the conditional after perturbation can be expressed as
$\tilde{\mathrm{P}}(x \mid y) = \prob{\sigma^{-1}(x) \mid y}$,
where $\sigma^{-1}(x)$ is the result of applying $\sigma^{-1}$ to every word in $x$.
Intuitively,
Theorem 1 tells us that if a word $w$ is a discriminative feature
for the original classification task,
then after the change of the input distribution,
the word $\sigma(w)$ should be a discriminative feature for the new task.
This property makes sense since
word-importance for classification should depends only on
the relative differences between classes.
In fact, the theorem holds when the input to the attention generator
is any function of unigram counts.
We also note that the classification performance on the query set
can be different,
as the pre-trained embedding function $f_\tx{ebd}(\cdot)$
is generally not invariant under such perturbations.
\section{Experimental setup}

\subsection{Datasets}
We evaluate our approach on five text classification datasets and one relation
classification dataset.\footnote{All processed datasets along with their splits are publicly available.}
(See Appendix~\ref{app:data} for more details.)

\textbf{20 Newsgroups } is comprised of informal discourse from news
discussion forums~\citep{lang1995newsweeder}.
Documents are organized under 20 topics.

\textbf{RCV1 } is a collection of Reuters newswire articles
from 1996 to 1997~\citep{lewis2004rcv1}.
These articles are written in formal speech and labeled with a set of topic codes.
We consider 71 second-level topics as our total class set and discard articles
that belong to more than one class.

\textbf{Reuters-21578 }
is a collection of shorter Reuters articles
from 1987~\citep{Lewis1997Reuters21578TC}.
We use the standard ApteMod version of the dataset.
We discard articles with more than one label
%and use the remaining
and consider
31 classes that have at least 20 articles.

\textbf{Amazon product data } contains customer reviews from 24
product categories~\citep{he2016ups}.
Our goal is to classify reviews into their respective product categories.
Since the original dataset is notoriously large (142.8 million reviews),
we select a more tractable subset by sampling 1000 reviews from each category.

\textbf{HuffPost headlines } consists of news headlines published on HuffPost
between 2012 and 2018~\citep{huffpost}.
These headlines split among 41 classes.
They are shorter and less grammatical than formal sentences.

\textbf{FewRel } is a relation classification dataset developed for
few-shot learning~\citep{han2018fewrel}.
Each example is a single sentence,
annotated with a head entity, a tail entity, and their relation.
The goal is to predict the correct relation between the head and tail.
The public dataset contains 80 relation types.

\subsection{Baselines}
We compare our model (denoted as \textsc{our}) to different combinations of
representations and learning algorithms.
Details of the baselines may be found in Appendix~\ref{app:baseline}.

\textbf{Representations }
We evaluate three representations.
\textsc{avg} represents each example
as the mean of its embeddings.
\textsc{idf} represents each example
as the weighted average of its word embeddings,
with weights given by inverse document frequency
over all training examples.
\textsc{cnn} applies 1D convolution over the input words
and obtains the representation
by max-over-time pooling~\citep{kim2014convolutional}.

\textbf{Learning algorithms }
In addition to the ridge regressor
(\textsc{rr})~\citep{bertinetto2018metalearning},
we evaluate two standard supervised learning algorithms
and two meta-learning algorithms.
\textsc{nn} is a 1-nearest-neighbor classifier
under Euclidean distance.
\textsc{ft} pre-trains a classifier over all training examples,
then finetunes the network using the support set~\citep{chen2018a}.
\textsc{maml} meta-learns a prior over model parameters,
so that the model can quickly adapt to new classes~\citep{finn2017model}.
Prototypical network (\textsc{proto})
meta-learns a metric space for few-shot classification by
minimizing the Euclidean distance between the centroid of each class
and its constituent examples~\citep{snell2017prototypical}.

\subsection{Implementation details}
We use pre-trained fastText embeddings~\citep{joulin2016fasttext}
for our model and all baselines.
For sentence-level datasets (HuffPost, FewRel), we also experiment with
pre-trained BERT 
embeddings~\citep{devlin2018bert} using HuggingFace’s codebase~\citep{Wolf2019HuggingFacesTS}.
For relation classification (FewRel),
we augment the input of our attention generator
with positional embeddings~\citep{zhang2017position}.\footnote{We also
provide the same positional embeddings to the baseline \textsc{cnn}.}

In the attention generator, we use a $\mathrm{biLSTM}$ with 50
hidden units and apply dropout of 0.1 on the output~\citep{srivastava2014dropout}.
In the ridge regressor,
we optimize meta-parameters
$\lambda$ and $a$ in the $\log$ space to maintain the
positivity constraint.
All parameters are optimized using Adam with a learning rate of
0.001~\citep{kingma2014adam}.

During meta-training, we sample 100 training episodes per epoch.
We apply early stopping
when the validation loss fails to improve for 20 epochs.
%We monitor validation loss for early stopping.
We evaluate test performance based on 1000 testing episodes
and report the average accuracy over 5 different random seeds.

\section{Results}
\label{sec:results}

\begin{table*}[t]
\small
\setlength{\tabcolsep}{1.1pt}
\begin{tabular}{llcccccccccccccc}
\toprule
\multicolumn{2}{c}{Method} &
\multicolumn{2}{c}{20 News} &
\multicolumn{2}{c}{Amazon} &
\multicolumn{2}{c}{HuffPost} &
\multicolumn{2}{c}{RCV1} &
\multicolumn{2}{c}{Reuters} &
\multicolumn{2}{c}{FewRel} &
\multicolumn{2}{c}{Average}
\\
\cmidrule(lr{0.5em}){1-2}\cmidrule(lr{0.5em}){3-4}\cmidrule(lr{0.5em}){5-6}\cmidrule(lr{0.5em}){7-8}\cmidrule(lr{0.5em}){9-10}\cmidrule(lr{0.5em}){11-12}\cmidrule(lr{0.5em}){13-14}
\cmidrule(lr{0.5em}){15-16}
Rep.\hspace{5mm} & Alg. & 1 shot & 5 shot & 1 shot & 5 shot & 1 shot & 5 shot & 1 shot & 5 shot & 1 shot & 5 shot & 1 shot & 5 shot & 1 shot & 5 shot \\
\midrule
\textsc{avg} & \textsc{nn}        & $33.9$ &  $45.8$ & $46.7$ &  $60.3$ & $31.4$ &  $41.5$ & $43.7$ &  $60.8 $ &  $56.5$ & $80.5$& $47.5$ &  $60.6$ & $43.3$ & $58.2$ \\
\textsc{idf} & \textsc{nn}        & $38.8$ &  $51.9$ & $51.4$ &  $67.1$ & $31.5$ &  $42.3$ & $41.9$ &  $58.2 $ &  $57.8$ & $82.9$& $46.8$ &  $60.6$ & $44.7$ & $60.5$ \\
\textsc{cnn} & \textsc{ft}       & $33.0$ &  $47.1$ & $45.7$ &  $63.9$ & $32.4$ &  $44.1$ & $40.3$ &  $62.3 $ &  $70.9$ & $91.0$& $54.0$ &  $71.1$ & $46.0$ & $63.2$ \\
\midrule
\textsc{avg} & \textsc{proto}     & $36.2$ &  $45.4$ & $37.2$ &  $51.9$ & $35.6$ &  $41.6$ & $28.4$ &  $31.2 $ &  $59.5$ & $68.1$& $44.0$ &  $46.5$ & $40.1$ & $47.4$ \\
\textsc{idf} & \textsc{proto}     & $37.8$ &  $46.5$ & $41.9$ &  $59.2$ & $34.8$ &  $50.2$ & $32.1$ &  $35.6 $ &  $61.0$ & $72.1$& $43.0$ &  $61.9$ & $41.8$ & $54.2$ \\
\textsc{cnn} & \textsc{proto}     & $29.6$ &  $35.0$ & $34.0$ &  $44.4$ & $33.4$ &  $44.2$ & $28.4$ &  $29.3 $ &  $65.2$ & $74.3$& $49.7$ &  $65.1$ & $40.1$ & $48.7$ \\
\midrule
\textsc{avg} & \textsc{maml}      & $33.7$ & $43.9$ & $39.3$ & $47.2$ & $36.1$ & $49.6$ & $39.9$ & $50.6$ &  $54.6$ & $62.5$& $43.8$ & $57.8$ & $41.2$ & $51.9$ \\
\textsc{idf} & \textsc{maml}      & $37.2$ & $48.6$ & $43.6$ & $62.4$ & $38.9$ & $53.7$ & $42.5$ & $54.1$ &  $61.5$ & $72.0$& $48.2$ & $65.8$ & $45.3$ & $59.4$ \\
\textsc{cnn} & \textsc{maml}      & $28.9$ & $36.7$ & $35.3$ & $43.7$ & $34.1$ & $45.8$ & $39.0$ & $51.1$ &  $66.6$ & $85.0$& $51.7$ & $66.9$ & $42.6$ & $54.9$ \\
\midrule
\textsc{avg} & \textsc{rr}        & $37.6$ &  $57.2$ & $50.2$ &  $72.7$ & $36.3$ &  $54.8$ & $48.1$ &  $72.6 $ &  $63.4$ & $90.0$& $53.2$ &  $72.2$ & $48.1$ & $69.9$ \\
\textsc{idf} & \textsc{rr}        & $44.8$ &  $64.3$ & $60.2$ &  $79.7$ & $37.6$ &  $59.5$ & $48.6$ &  $72.8 $ &  $69.1$ & $93.0$& $55.6$ &  $75.3$ & $52.6$ & $74.1$ \\
\textsc{cnn} & \textsc{rr}        & $32.2$ &  $44.3$ & $37.3$ &  $53.8$ & $37.3$ &  $49.9$ & $41.8$ &  $59.4 $ &  $71.4$ & $87.9$& $56.8$ &  $71.8$ & $46.1$ & $61.2$ \\
\midrule
\textsc{our} & & $\bm{52.1}$ & $\bm{68.3}$ & $\bm{62.6}$ & $\bm{81.1}$ & $\bm{43.0}$ & $\bm{63.5}$ & $\bm{54.1}$ & $\bm{75.3}$ & $\bm{81.8}$ & $\bm{96.0}$& $\bm{67.1}$ & $\bm{83.5}$ & $\bm{60.1}$ & $\bm{78.0}$\\
\midrule
\midrule
\multicolumn{2}{l}{\textsc{our}\text{ w/o }$\mathrm{t}(\cdot)$}        & $50.1$ & $67.5$ & $61.7$ & $80.5$ & $42.0$ & $60.8$ & $51.5$ & $75.1$ &  $76.7$ & $93.7$& $66.9$ & $83.2$ & $58.1$ & $76.8$ \\
\multicolumn{2}{l}{\textsc{our}\text{ w/o }$\mathrm{s}(\cdot)$}        & $41.9$ & $60.7$ & $51.1$ & $75.3$ & $40.1$ & $60.2$ & $48.5$ & $72.8$ &  $78.1$ & $94.8$& $65.8$ & $82.6$ & $54.2$ & $74.4$ \\
\multicolumn{2}{l}{\textsc{our}\text{ w/o }$\mathrm{bi}\textsc{lstm}$}            & $50.3$ & $66.9$ & $61.9$ & $80.9$ & $42.2$ & $63.0$ & $51.8$ & $74.1$ & $77.2$ & $95.4$ & $66.4$ & $82.9$ & $58.3$ & $77.2$ \\
\midrule
\multicolumn{2}{l}{\textsc{our}\text{ w }$\textsc{ebd}$}  &  $39.7$ & $57.5$ & $56.5$ & $76.3$ & $40.6$ & $58.6$ & $48.6$ & $71.5$ &  $81.7$ & $95.8$& $61.5$ & $80.9$ & $54.8$ & $73.4$ \\
\bottomrule
\end{tabular}
\centering
\caption{Results of 5-way 1-shot and 5-way 5-shot classification on six datasets.
The bottom four rows present our ablation study.
For complete results with standard deviations
see Table~\ref{tab:bigtable51} and \ref{tab:bigtable55} in Appendix~\ref{app:result}.}\label{table:bigtable}
\vspace{-5mm}
\end{table*}

\begin{figure}[t]
  \centering
  \vspace{-5mm}
  \begin{tabular}{c c }
    \includegraphics[width=.45\linewidth]{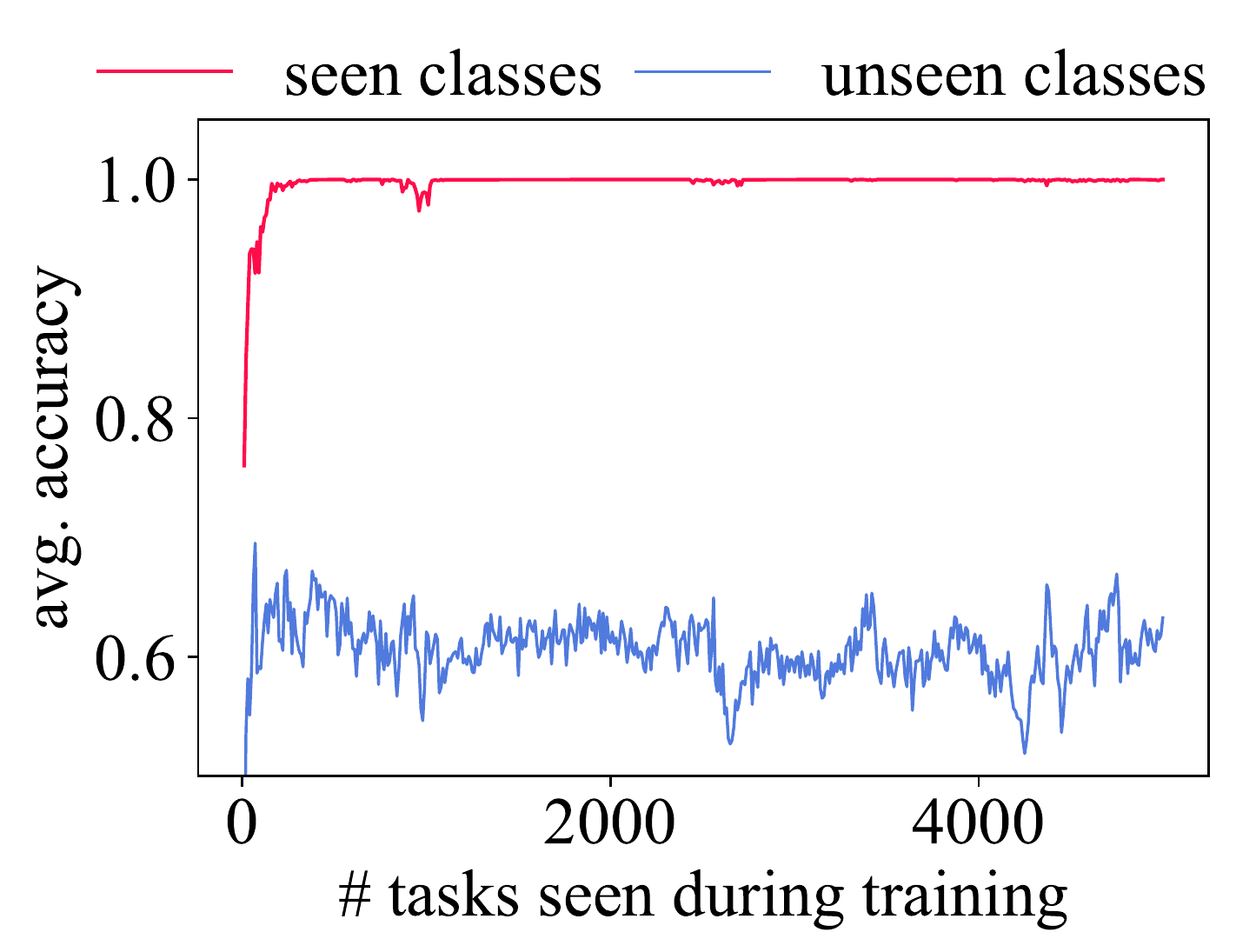}
    &
    \includegraphics[width=.45\linewidth]{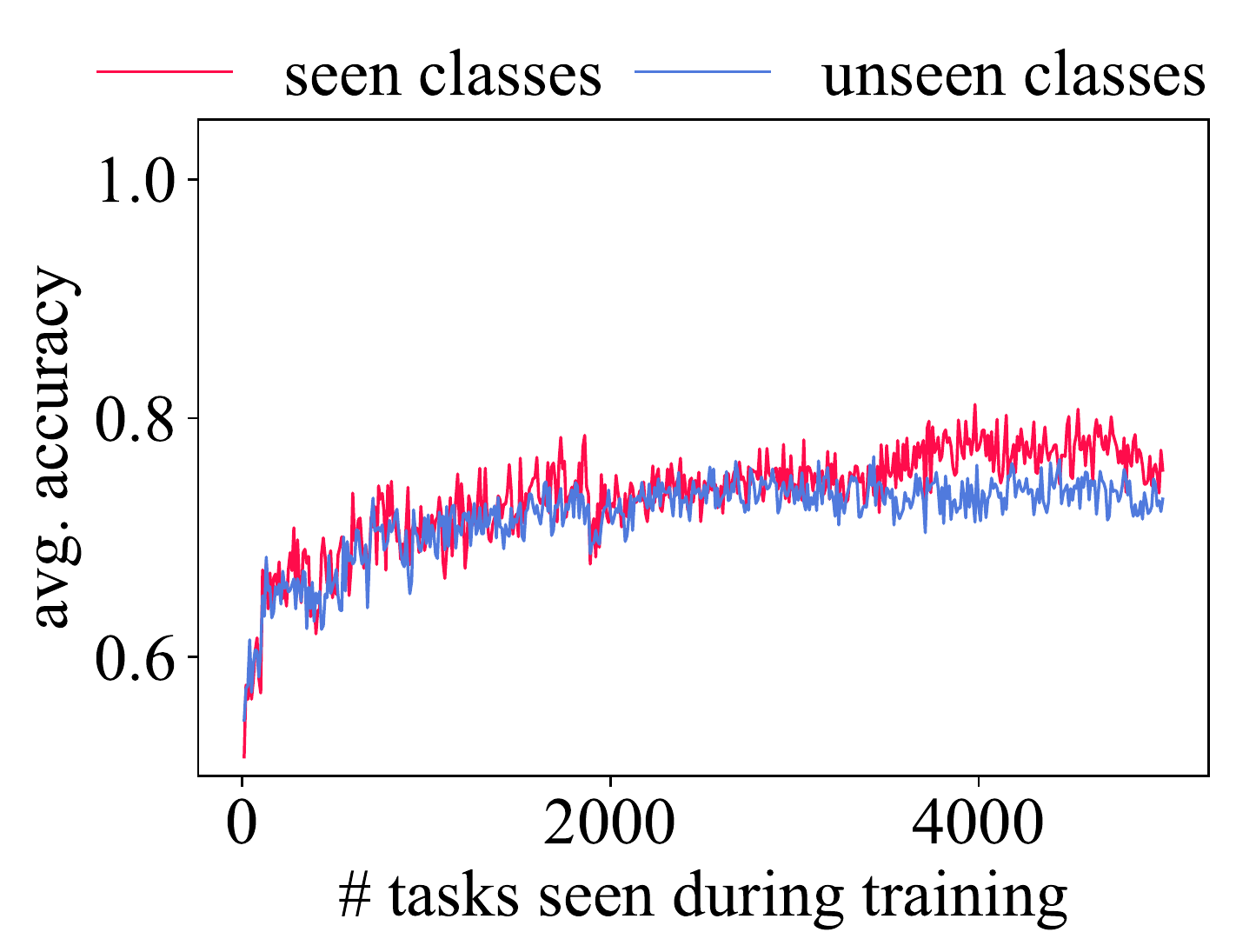}
    \\
    \fontsize{5}{4}\selectfont
    \footnotesize (a) \textsc{cnn}+\textsc{proto}
    &
    (b) \textsc{our}
  \end{tabular}
  \vspace{-3mm}
  \captionof{figure}{Learning curve of \textsc{cnn}+\textsc{proto} (left)
    v.s.  \textsc{our} (right) on the Reuters dataset.
    We plot average 5-way 1-shot accuracy over 50 episodes sampled from
    seen classes (blue) and unseen classes (red).
    While \textsc{our} has weaker representational power,
    it generalizes better to unseen classes.
  }
  \label{fig:learningcurve}
\end{figure}

We evaluated our model in both 5-way 1-shot and 5-way 5-shot settings.
These results are reported in Table~\ref{table:bigtable}.
Our model consistently achieves the best performance across all datasets.
On average, our model
improves 5-way 1-shot accuracy by 7.5\%
and 5-way 5-shot accuracy by 3.9\%,
against the best baseline for each dataset.
When comparing against \textsc{cnn}+\textsc{proto},
our model improves by 20.0\% on average in 1-shot classification.
The empirical results clearly demonstrate that
meta-learners privy to lexical information consistently fail,
while our model is able to generalize past
class-specific vocabulary.
Furthermore, Figure~\ref{fig:learningcurve} illustrates that
a lexicon-aware meta-learner (\textsc{cnn}+\textsc{proto})
is able to overfit the training data faster than our model,
but our model more readily generalizes to unseen classes.

% learning curves figure

\begin{table*}[t]
\small
\setlength{\tabcolsep}{10pt}
\begin{tabular}{llcccccc}
\toprule
\multicolumn{2}{c}{Method}& \multicolumn{2}{c}{HuffPost} & \multicolumn{2}{c}{FewRel}\\
\cmidrule(lr{0.5em}){1-2}\cmidrule(lr{0.5em}){3-4}\cmidrule(lr{0.5em}){5-6}
Rep. & Alg. & 1 shot & 5 shot & 1 shot & 5 shot \\
\midrule
\textsc{avg} &\textsc{nn} &$23.23 $ {\tiny $ \pm 0.20$} & $34.22 $ {\tiny $ \pm 0.28$} & $52.37 $ {\tiny $ \pm 0.27$} & $64.85 $ {\tiny $ \pm 0.25$} \\
\textsc{idf} &\textsc{nn} &$33.49 $ {\tiny $ \pm 0.18$} & $45.71 $ {\tiny $ \pm 0.17$} & $48.65 $ {\tiny $ \pm 0.18$} & $62.35 $ {\tiny $ \pm 0.18$} \\
\textsc{cnn} &\textsc{ft} &$37.30 $ {\tiny $ \pm 1.08$ } & $51.56 $ {\tiny $ \pm 1.28$} & $61.10$ {\tiny $\pm 2.54$ } & $80.04$ {\tiny $\pm 0.69$ }\\
\midrule
\textsc{avg} &\textsc{proto} &$34.21 $ {\tiny $ \pm 0.56$} & $49.77 $ {\tiny $ \pm 1.90$} & $50.27 $ {\tiny $ \pm 0.98$} & $66.24 $ {\tiny $ \pm 2.01$} \\
\textsc{idf} &\textsc{proto} &$36.06 $ {\tiny $ \pm 0.84$} & $54.58 $ {\tiny $ \pm 0.99$} & $48.23 $ {\tiny $ \pm 0.58$} & $67.82 $ {\tiny $ \pm 0.72$} \\
\textsc{cnn} &\textsc{proto} &$36.17 $ {\tiny $ \pm 1.00$} & $50.55 $ {\tiny $ \pm 0.96$} & $57.08 $ {\tiny $ \pm 5.52$} & $75.01 $ {\tiny $ \pm 2.21$} \\
%\midrule
%\textsc{avg} &\textsc{maml} &$35.68 $ {\tiny $ \pm 2.25$} & $47.50 $ {\tiny $ \pm 0.52$} & $37.68 $ {\tiny $ \pm 2.04$} & $50.04 $ {\tiny $ \pm 3.17$} \\
%\textsc{idf} &\textsc{maml} &$38.74 $ {\tiny $ \pm 0.71$} & $50.34 $ {\tiny $ \pm 0.57$} & $40.25 $ {\tiny $ \pm 2.06$} & $51.84 $ {\tiny $ \pm 2.94$} \\
%\textsc{cnn} &\textsc{maml} &$36.46 $ {\tiny $ \pm 0.88$} & $45.19 $ {\tiny $ \pm 1.22$} & $53.14 $ {\tiny $ \pm 1.26$} & $70.32 $ {\tiny $ \pm 1.73$} \\
\midrule
\textsc{avg} &\textsc{maml}
             &$38.58$ {\tiny $\pm 1.56$ }
             &$55.32$ {\tiny $\pm 1.42$ }
             &$47.18$ {\tiny $\pm 3.49$ }
             &$64.50$ {\tiny $\pm 2.72$ }\\
\textsc{idf} &\textsc{maml} 
             &$34.22$ {\tiny $ \pm 0.74$}
             &$56.50$ {\tiny $\pm 1.50$}
             &$50.06$ {\tiny $\pm 2.88$}
             &$68.43$ {\tiny $\pm 2.50$}\\
\textsc{cnn} &\textsc{maml}
             &$38.39$ {\tiny $\pm 1.68$}
             &$53.86$ {\tiny $\pm 0.76$}
             &$47.68$ {\tiny $\pm 1.66$}
             &$71.56$ {\tiny $\pm 4.75$}\\
\midrule
\textsc{avg} &\textsc{rr}& $25.34 $ {\tiny $ \pm 0.14$} & $51.52 $ {\tiny $ \pm 0.14$} & $55.65 $ {\tiny $ \pm 0.27$} & $73.91 $ {\tiny $ \pm 0.77$} \\
\textsc{idf} &\textsc{rr} & $40.38 $ {\tiny $ \pm 0.11$} & $61.72 $ {\tiny $ \pm 1.03$} & $54.48 $ {\tiny $ \pm 0.26$} & $73.48 $ {\tiny $ \pm 0.72$} \\
\textsc{cnn} &\textsc{rr} &$41.37 $ {\tiny $ \pm 0.54$} & $53.10 $ {\tiny $ \pm 0.76$} & $65.65 $ {\tiny $ \pm 5.70$} & $78.65 $ {\tiny $ \pm 4.24$} \\
\midrule
\textsc{our} & & $\bm{42.12} $ {\tiny $ \pm 0.15$} & $\bm{62.97} $ {\tiny $ \pm 0.67$} & $\bm{70.08} $ {\tiny $ \pm 0.56$} & $\bm{88.07} $ {\tiny $ \pm 0.27$} \\
\bottomrule
\end{tabular}
\centering
\caption{5-way 1-shot and 5-way 5-shot classification on HuffPost and FewRel
using BERT.}\label{tab:bert-std}
\vspace{-5mm}
\end{table*}

\begin{figure}[t]
  \centering
  \begin{subfigure}[t]{.32\linewidth}
    \centering
    \includegraphics[width=\linewidth]{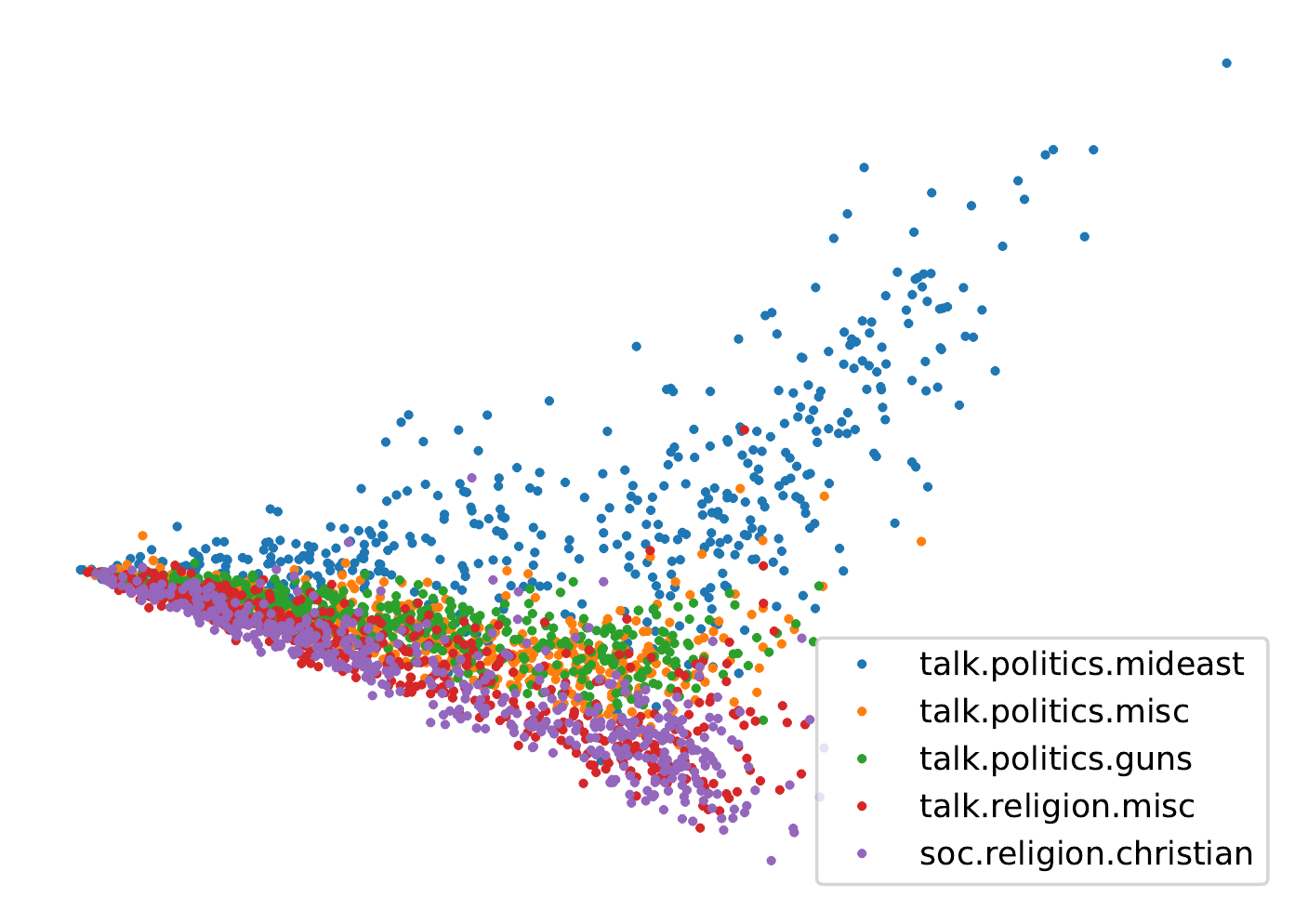}
    \caption{$\mathrm{s}(\cdot)$}
  \end{subfigure}
  \begin{subfigure}[t]{.32\linewidth}
    \centering
    \includegraphics[width=\linewidth]{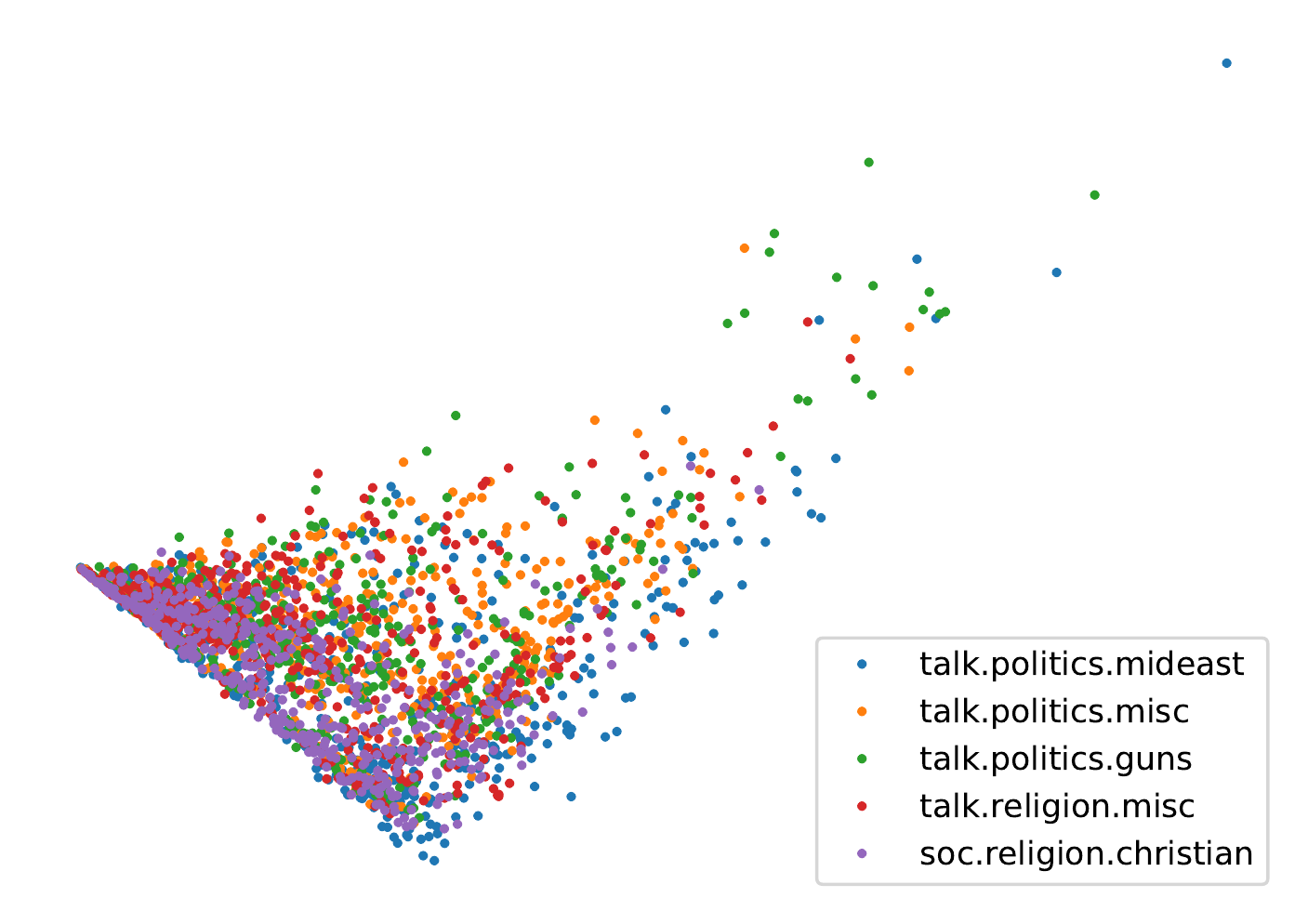}
    \caption{$\mathrm{t}(\cdot)$}
  \end{subfigure}
  \begin{subfigure}[t]{.32\linewidth}
    \centering
    \includegraphics[width=\linewidth]{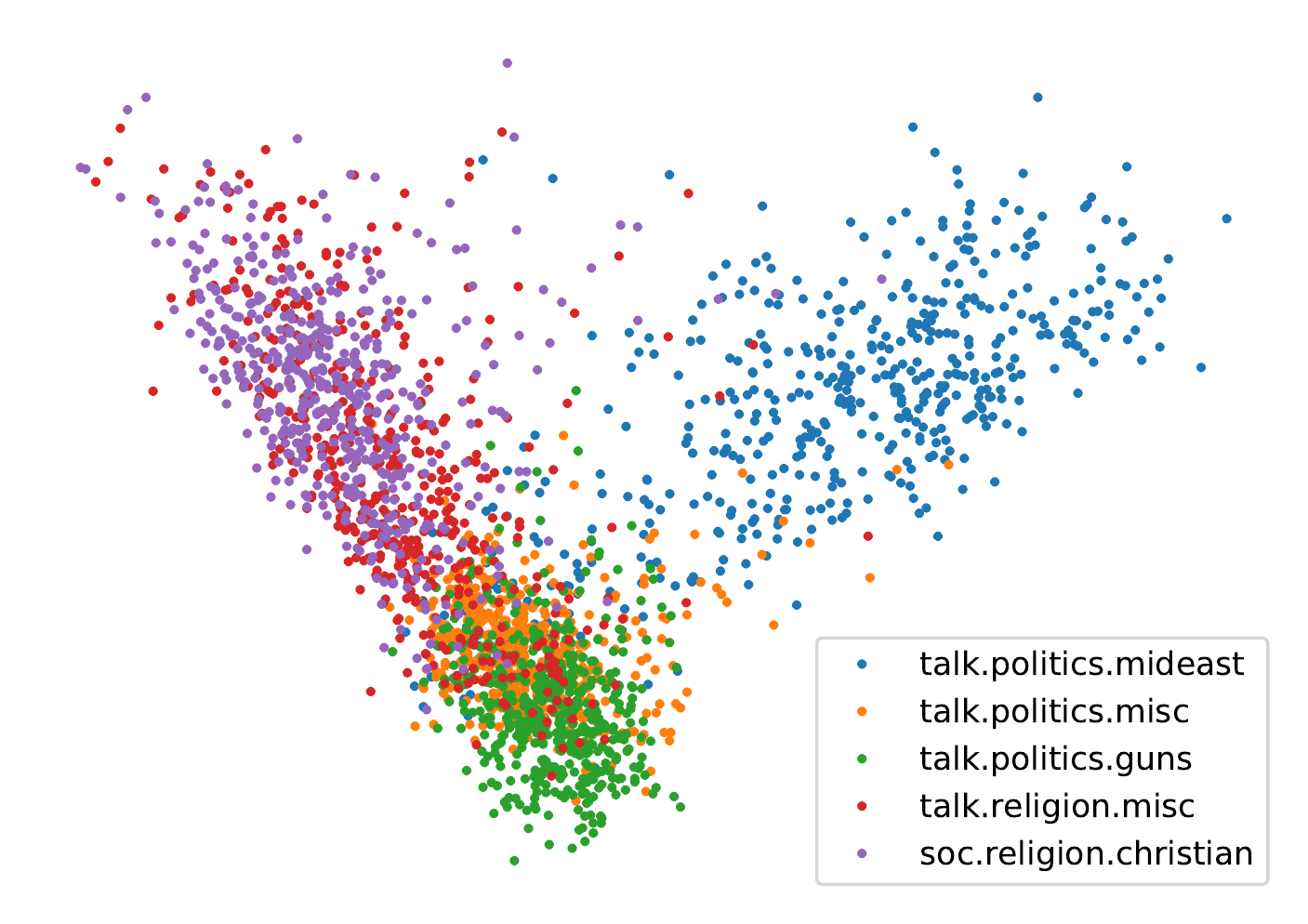}
    \caption{\textsc{our}}
  \end{subfigure}
  \caption{PCA visualization of the input representation for
    the query set of a testing episode ($N=5$, $K=5$, $L=500$)
    sampled from 20~Newsgroups.
    Weighted averages of word embeddings given by
    (a) $\mathrm{s}(\cdot)$,
    (b) $\mathrm{t}(\cdot)$, and
    (c) the attention generator
    meta-trained on a disjoint set of training classes.}\label{fig:pca_20news}
\end{figure}

\noindent\textbf{Ablation study }
We perform ablation studies on the attention generator.
These results are shown at the bottom of Table~\ref{table:bigtable}.
We observe that
both statistics $\mathrm{s}(\cdot)$ and $\mathrm{t}(\cdot)$
contribute to the performance,
though the former has a larger impact.
We also note that instead of
computing word importance independently for each word using a multilayer perceptron (\textsc{our} w/o bi\textsc{lstm}),
fusing information across the input with an $\mathrm{biLSTM}$
improves performance slightly.

Finally, we observe that restricting meta-knowledge to distributional signatures is essential to performance: when both lexical word embeddings and distributional signatures are fed to the attention generator (\textsc{our} w \textsc{ebd}), the performance drops consistently.

\noindent\textbf{Contextualized representations }
For sentence-level datasets (FewRel, HuffPost),
we also experiment with contextualized representations,
given by BERT~\citep{devlin2018bert}.
These results are shown in Table~\ref{tab:bert-std}.
While BERT
significantly improves classification performance on FewRel,
we observe no performance boost on HuffPost.
We postulate that this discrepancy
arises because relation classification is highly contextual,
while news classification is mostly keyword-based.

\noindent\textbf{Analysis }
We visualize our attention-weighted representation $\phi(x)$
in Figure~\ref{fig:pca_20news}.
Compared to directly using general word importance $\mathrm{s}(x)$
or class-specific word importance $\mathrm{t}(x)$,
our method produces better separation,
which enables effective learning from a few examples.
This highlights that the power of our approach
lies not in the distributional signatures themselves,
but rather in the representations learned on top of them.

{
\renewcommand{\hlr}[2]{\setlength{\fboxsep}{0.3pt}\colorbox{red!#2}{\rule[-0.2\baselineskip]{0pt}{0.8\baselineskip}{#1}}}
\setlength{\tabcolsep}{8pt}
\begin{table*}[t!]
    \small
    \centering
    \begin{tabular}{r p{0.42\linewidth} p{0.42\linewidth}}
        %\toprule
        & \multicolumn{1}{c}{Fine-grained classification} &
        \multicolumn{1}{c}{Coarse-grained classification} \\
        \cmidrule(lr){2-2}
        \cmidrule(lr){3-3} \vspace{-6pt} \\
        \arrayrulecolor{black}
        \emph{$\mathrm{s}(x)$} &
        \hlr{finnish}{0} \hlr{unemployment}{100} \hlr{was}{2} \hlr{6.7}{0} \hlr{pct}{3} \hlr{in}{0} \hlr{december}{25} \hlr{last}{5} \hlr{year}{3} \hlr{compared}{19} \hlr{with}{2} \hlr{6.8}{0} \hlr{pct}{3} \hlr{in}{0} \hlr{november}{42} \hlr{and}{0} \hlr{6.1}{0} \hlr{pct}{3} \hlr{in}{0} \hlr{december}{25} \hlr{1985}{17} \hlr{,}{0} \hlr{the}{0} \hlr{central}{34} \hlr{statistical}{76} \hlr{office}{76} \hlr{said}{0}
        &
        \hlr{finnish}{0} \hlr{unemployment}{100} \hlr{was}{2} \hlr{6.7}{0} \hlr{pct}{3} \hlr{in}{0} \hlr{december}{25} \hlr{last}{5} \hlr{year}{3} \hlr{compared}{19} \hlr{with}{2} \hlr{6.8}{0} \hlr{pct}{3} \hlr{in}{0} \hlr{november}{42} \hlr{and}{0} \hlr{6.1}{0} \hlr{pct}{3} \hlr{in}{0} \hlr{december}{25} \hlr{1985}{17} \hlr{,}{0} \hlr{the}{0} \hlr{central}{34} \hlr{statistical}{76} \hlr{office}{76} \hlr{said}{0}
        \vspace{0.3cm}\\
        \arrayrulecolor{black}
        \emph{$\mathrm{t}(x)$} &
        \hlr{finnish}{0} \hlr{unemployment}{100} \hlr{was}{1} \hlr{6.7}{0} \hlr{pct}{24} \hlr{in}{1} \hlr{december}{4} \hlr{last}{3} \hlr{year}{7} \hlr{compared}{4} \hlr{with}{2} \hlr{6.8}{0} \hlr{pct}{24} \hlr{in}{1} \hlr{november}{4} \hlr{and}{2} \hlr{6.1}{0} \hlr{pct}{24} \hlr{in}{1} \hlr{december}{4} \hlr{1985}{0} \hlr{,}{3} \hlr{the}{1} \hlr{central}{1} \hlr{statistical}{7} \hlr{office}{12} \hlr{said}{1}
        &
        \hlr{finnish}{0} \hlr{unemployment}{100} \hlr{was}{3} \hlr{6.7}{0} \hlr{pct}{87} \hlr{in}{1} \hlr{december}{4} \hlr{last}{5} \hlr{year}{15} \hlr{compared}{4} \hlr{with}{2} \hlr{6.8}{0} \hlr{pct}{87} \hlr{in}{1} \hlr{november}{5} \hlr{and}{2} \hlr{6.1}{0} \hlr{pct}{87} \hlr{in}{1} \hlr{december}{4} \hlr{1985}{0} \hlr{,}{1} \hlr{the}{1} \hlr{central}{4} \hlr{statistical}{28} \hlr{office}{17} \hlr{said}{2}
        \vspace{0.3cm}\\
        \emph{\textsc{Ours}} &
        \hlr{finnish}{0} \hlr{unemployment}{84} \hlr{was}{0} \hlr{6.7}{0} \hlr{pct}{0} \hlr{in}{0} \hlr{december}{0} \hlr{last}{0} \hlr{year}{0} \hlr{compared}{0} \hlr{with}{0} \hlr{6.8}{0} \hlr{pct}{0} \hlr{in}{0} \hlr{november}{0} \hlr{and}{0} \hlr{6.1}{0} \hlr{pct}{0} \hlr{in}{0} \hlr{december}{0} \hlr{1985}{0} \hlr{,}{0} \hlr{the}{0} \hlr{central}{2} \hlr{statistical}{13} \hlr{office}{5} \hlr{said}{0} &
        \hlr{finnish}{0} \hlr{unemployment}{100} \hlr{was}{0} \hlr{6.7}{0} \hlr{pct}{0} \hlr{in}{0} \hlr{december}{0} \hlr{last}{0} \hlr{year}{0} \hlr{compared}{0} \hlr{with}{0} \hlr{6.8}{0} \hlr{pct}{0} \hlr{in}{0} \hlr{november}{0} \hlr{and}{0} \hlr{6.1}{0} \hlr{pct}{0} \hlr{in}{0} \hlr{december}{0} \hlr{1985}{0} \hlr{,}{0} \hlr{the}{0} \hlr{central}{6} \hlr{statistical}{80} \hlr{office}{8} \hlr{said}{0}
    \end{tabular}
    \captionof{figure}{
      Attention weights generated by our model are specific to task.
      We visualize our model's inputs $\mathrm{s}(x)$ (top), $\mathrm{t}(x)$ (middle),
      and output (bottom) for one query example from class \emph{jobs} in Reuters dataset.
      Word ``statistical'' is downweighed for \emph{jobs} when compared to
      other economics classes (left), but it becomes important when considering
      dissimilar classes (right).
      Fine-grained classes: \emph{jobs}, \emph{retail}, \emph{industrial production index}, \emph{wholesale production index}, \emph{consumer production index}.
      Coarse-grained classes: \emph{jobs}, \emph{cocoa}, \emph{aluminum}, \emph{copper}, \emph{reserves}.
    }\label{fig:wpi}
\end{table*}
}

Figure~\ref{fig:wpi} visualizes the model's input and output
on the same query example in two testing episodes.
The example belongs to the class \emph{jobs} in the Reuters dataset.
First, we observe that our model
generates meaningful attention
from noisy distributional signatures.
Furthermore, the generated attention is \emph{task-specific}:
in the depicted example,
if the episode contains other economics-related classes,
the word ``statistical'' is downweighed by our model.
Conversely, ``statistical'' is upweighted
when we compare \emph{jobs} to other distant classes.

\section{Conclusion}
In this paper,
we propose a novel meta-learning approach
that capitalizes on the connection between
word importance and distributional signatures
to improve few-shot classification.
Specifically, we learn an attention generator
that translates distributional statistics into
high-quality attention.
This generated attention then provides guidance
for fast adaptation to new classification tasks.
Experimental results on both text and relation classification
validate that our model identifies important words
for new classes.
The effectiveness of our approach demonstrates
the promise of meta learning with distributional signatures.

\section*{Acknowledgments}
We thank the MIT NLP group and the reviewers for their helpful discussion and comments. This work is supported by MIT-IBM Watson AI Lab and Facebook Research. Any opinions, findings, conclusions, or recommendations expressed in this paper are those of the authors, and do not necessarily reflect the views of the funding organizations.
%In this paper, we propose a novel approach
%to meta-learning for NLP using distributional signatures.
%We learn an attention-generator
%using distributional statistics of our input,
%then leverage that attention
%to improve few-shot classification performance.
%Experimental results on both domain and relation classification
%validate that our model consistently outperforms
%lexical meta-learning methods for NLP.
%The effectiveness of our approach demonstrates
%the promise of meta-learning with distributional characteristics,
%for NLP and other areas.

\newpage

\bibliography{iclr2020_conference}
\bibliographystyle{iclr2020_conference}

\appendix
\clearpage
\section{Supplemental Material}
\subsection{Regularized linear classifier}\label{app:softmax}
Given an $N$-way $K$-shot classification task,
the goal of the regularized linear classifier is to
approximate task-specific word importance
using the support set.

Let $x = \{x_1,\ldots x_T\}$ be an input example,
and let $f_\tx{ebd}(\cdot)$ be an embedding function that maps
each word $x_i$ into $\RR^E$.
We compute the representation of $x$ by the average of its embeddings:
\[
  \psi(x) \coloneqq \frac{1}{T} \sum_i f_\tx{ebd}(x_i).
\]
Since the support set only contains a few examples,
we adopt a simple linear classifier to reduce overfitting:
\[
  \hat{p} \coloneqq \mathrm{softmax}(W \psi(x))
\]
where $W \in \RR^{N \times E}$ is the weight matrix to learn.
We minimize the cross entropy loss
between the prediction $\hat{p}$ and the ground truth label
while penalizing the Frobenius norm of $W$.
We stop training
once the gradient norm is less than $0.1$.
Finally, given a word $x_i$, we estimate its conditional probability via
$\mathrm{softmax}(W \psi(x_i))$.

\textbf{Time efficiency }
Since the support set is very small (less than 25 examples)
and the loss function is strongly convex,
this linear classifier converges very fast in practice.\footnote{less than 1
second on a single GeForce GTX TITAN X}
Note that for larger problems,
we can speed up computation by formulating this procedure as a regression problem
and solving for its closed-form solution (as in Section~\ref{sec:r2d2}).

\subsection{Proof of Theorem 1}\label{app:thm}
\begin{customthm}{1}
  Assume $\sigma: V \to V$ satisfies $\prob{w} = \prob{\sigma(w)}$ for all $w$.
  If $\sigma$ is a bijection,
  then for any input text $x \in \cS \cup \cQ$ and
  its perturbation $\tilde{x} \in \tilde{\cS} \cup \tilde{\cQ}$,
  the outputs of the attention generator are the same:
  \[
    \text{AttGen}(x\mid \cS, \cP) =
    \text{AttGen}(\tilde{x} \mid \tilde{\cS}, \cP).
  \]
\end{customthm}

\begin{proof}
It suffices to show that
the general word importance $\mathrm{s}(\cdot)$
and the class-specific word importance $\mathrm{t}(\cdot)$
are the same under the perturbation $\phi$.
For the former,
since $\phi$ preserves unigram probability estimated over the source pool,
we have
\[
\mathrm{s}(x_i) =
\frac{\varepsilon}{\varepsilon + \prob{x_i}}
=
\frac{\varepsilon}{\varepsilon + \prob{\sigma(x_i)}}
=
\mathrm{s}(\sigma(x_i)).
\]
To prove the latter,
we need to show that the conditional probability $\prob{y \mid x_i}$
estimated over the support set
is also invariant under $\phi$.
Let $\#(x_i \land y \mid \cS)$ denote the occurrences of the word $x_i$ in
examples with class label $y$
over the support set $\cS$.
Using maximum likelihood estimation, we have
\[
\hat{\mathrm{P}}(y \mid x_i, \cS)
= \frac{\#(x_i \land y \mid \cS)}{\sum_{y'} \#(x_i \land y' \mid \cS)}
= \frac{\#(\sigma(x_i) \land y \mid \tilde{\cS})}{\sum_{y'} \#(\sigma(x_i) \land y' \mid \tilde{\cS})}
=\hat{\mathrm{P}}(y \mid \sigma(x_i), \tilde{\cS}),
\]
where the second equality is derived from the fact that $\sigma$ is a bijection.
\end{proof}

\subsection{Learning procedure}

Algorithm~\ref{alg:train} contains the pseudo code for our learning procedure.
We apply early stopping when validation loss does not improve for 20 epochs.

\renewcommand{\algorithmicrequire}{\textbf{Input:}}
\renewcommand{\algorithmicensure}{\textbf{Hyperparameters:}}
\begin{algorithm*}[h]\small
 \begin{algorithmic}
    \Require Training set $\cD = \set{(x_1,y_1),(x_2,y_2),\dots}$
    where each $y_i\in\cY^\tx{train} = \set{1,\dots,N_\tx{train}}$
    \Ensure parameters of $\phi$; coefficients for losses $\lambda, a, b$

    \Repeat

     \State $\cY \lto\textsc{Sample}(\cY^\tx{train}, N)$
     \Comment sample target classes

     \State $\cD^\tx{src} \lto \cD(\cY^\tx{train} \setminus \cY)$
     \Comment create support pool

     \State $\cT^\tx{S}, \cT^\tx{Q} \lto \emptyset, \emptyset$
     \Comment sample support set and query set

     \For{$y\in\cY$}
        \State $\cT^\tx{S}\lto \cT^\tx{S}\cup \textsc{Sample}(\mathcal{D}(\set{y}), K)$
        \State $\cT^\tx{Q}\lto \cT^\tx{Q}\cup \textsc{Sample}(\mathcal{D}(\set{y}) \setminus \cT^\tx{S}, L)$
     \EndFor
     
     \State Generate attention scores using $\cT^\tx{S}, \cT^\tx{Q}, \cD^\tx{src}$
     \Comment Eq.~\ref{eq:alpha}
     
     \State Construct attention-weighted representations $\Phi_\tx{S}, \Phi_\tx{Q}$
     using $\phi(\cdot)$
     \Comment Eq.~\ref{eq:phi}

     \State Compute closed-form solution $W$ of ridge regression 
     from $\lambda$, $\Phi_S$, and support labels $Y_\tx{S}$
     \Comment Eq.~\ref{eq:w}

     \State Given $W, a, b$, compute cross entropy loss $\mathcal{L^\tx{CE}}$ on the query set $\Phi_\tx{Q}, Y_\tx{Q}$

     \State Update meta parameters (attention generator and $a, B$) using $\mathcal{L}^\tx{CE}$

    \Until stopping criterion is met

 \end{algorithmic}

 \caption{Meta-training procedure.
 $N_\tx{train} = \left| \cY^\tx{train} \right|$ is the number of training classes.
 $N < N_\tx{train}$ is the number of classes of each few-shot task.
 $K, L$ are the number of support and query examples
 per target class, respectively.
 To ease notation, we use
 $\textsc{Sample}(S,N)$ to denote a subset of $S$ with $N$ elements,
 chosen uniformly at random.
 We use $\cD(\cY)\subseteq\cD$ to denote the set of all elements $(x_i,y_i)$
 for which $y_i\in\cY$.}
 \label{alg:train}
\end{algorithm*}

\subsection{Datasets}\label{app:data}
To reliably test our model's ability to generalize across classes,
we consider two data splitting mechanisms in our experiments:
1) \emph{easy split}:
we randomly permute all classes and split them into train/val/test;
2) \emph{hard split}:
we select train/val/test based on the class hierarchy
such that train classes are distant to val and test.
We applied the easy split to
one sentence-level dataset (HuffPost)
and one document-level dataset (Reuters-21578).
Hard split is used for the other four datasets (details below).
This setting tests the generalization capacity of the algorithm, following~\cite{xian2017zeroshot}.

%We postulate that existing meta-learning methods
%work poorly when the train and test datasets contain
%dissimilar low-level features.
%Thus, we split our datasets into train, validation, and test
%to minimize the overlap of train and test vocabulary.

\textbf{20 Newsgroups}
Each class in 20 Newsgroups belongs to one of six top-level categories,
which roughly correspond to computers, recreation, science, politics,
religion, and for-sale.
We partition the set of labels so that no top-level category spans two splits.
Train contains ``sci'' and ``rec,''
val contains ``comp,'' and test contains all other labels.

\textbf{Amazon}
The Amazon dataset does not come with predefined top-level categories.
To generate a hard split,
we first apply spectral clustering to classes based on their word distributions.
Then we select train/val/test from different clusters.

\textbf{RCV1}
We apply the same approach as above.

\textbf{FewRel}
While FewRel does not provide higher-level categories,
we observe that most relations occur between named entities of similar types.
Thus, we extract the named entity type of the head and tail for each example
using a pretrained spaCy parser.\footnote{\url{https://spacy.io/}}
For each class, we determine the most common head and tail entity types.
Test contains all classes for which the most common head entity type is
\textsc{Work\_of\_Art}.
Train and validation were arbitrarily split to contain the remaining relations.

{
\setlength{\tabcolsep}{3pt}
\begin{table}[t]
  \centering
  \small
  \begin{tabular}{lrrrrrr}
    \toprule
    Dataset
    & \# tokens / example
    & vocab size
    & \# examples / cls
    & \# train cls
    & \# val cls
    & \# test cls\\
    \midrule
    20 Newsgroups    & 340& 32137 & 941  & 8  & 5  & 7  \\
    RCV1    & 372& 7304  & 20   & 37 & 10 & 24 \\
    Reuters & 168& 2234  & 20   & 15 & 5  & 11 \\
    Amazon  & 140& 17062 & 1000 & 10 & 5  & 9  \\
    HuffPost & 11 & 8218  & 900  & 20 & 5  & 16 \\
    FewRel      & 24 & 16045 & 700  & 65 & 5  & 10 \\
    \bottomrule
  \end{tabular}
  \caption{Dataset statistics. See Appendix~\ref{app:data} for details.}\label{table:dataset}
  \vspace{-3mm}
\end{table}
}

\subsection{Other baselines}

We also consider two other baselines: induction network~\citep{geng2019few} and P-MAML~\citep{zhang2019improving}.
Implementation details may be found in Appendix~\ref{app:baseline}.

\begin{itemize}
    \item
    \textbf{Induction network} encodes input examples using a biLSTM with self-attentive pooling~\citep{lin2017structured}.
    Based on the encoded representation,
    it then computes a prototype for each class through dynamic routing over the support set~\citep{sabour2017dynamic}.
    Finally, it uses a neural tensor layer~\citep{socher2013reasoning} to predict the relation between each query example and the class prototypes.
    \item
    \textbf{P-MAML} combines pre-training with MAML.
    It first finetunes pre-trained BERT representation on the meta-training data using masked language modeling~\citep{devlin2018bert}.
    Based on this finetuned representation,
    it trains first-order MAML~\citet{finn2017model} to enable fast adaptation.
\end{itemize}

\begin{figure}[t]
  \centering
  \begin{subfigure}[t]{.45\linewidth}
    \centering
    \includegraphics[width=\linewidth]{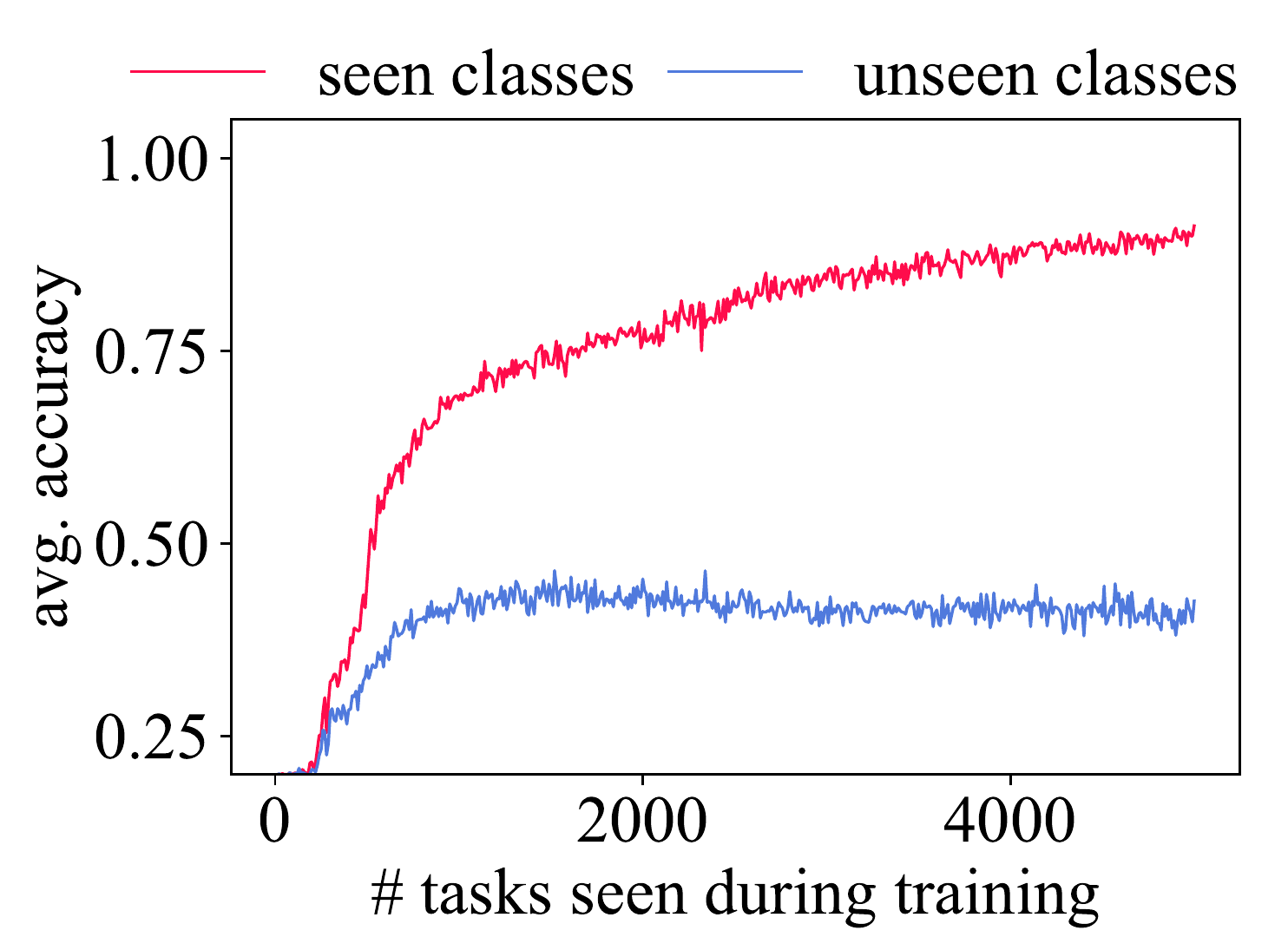}
    \vspace{-0.6cm} % account for pdf space
    \caption{HuffPost}
  \end{subfigure}
  \begin{subfigure}[t]{.45\linewidth}
    \centering
    \includegraphics[width=\linewidth]{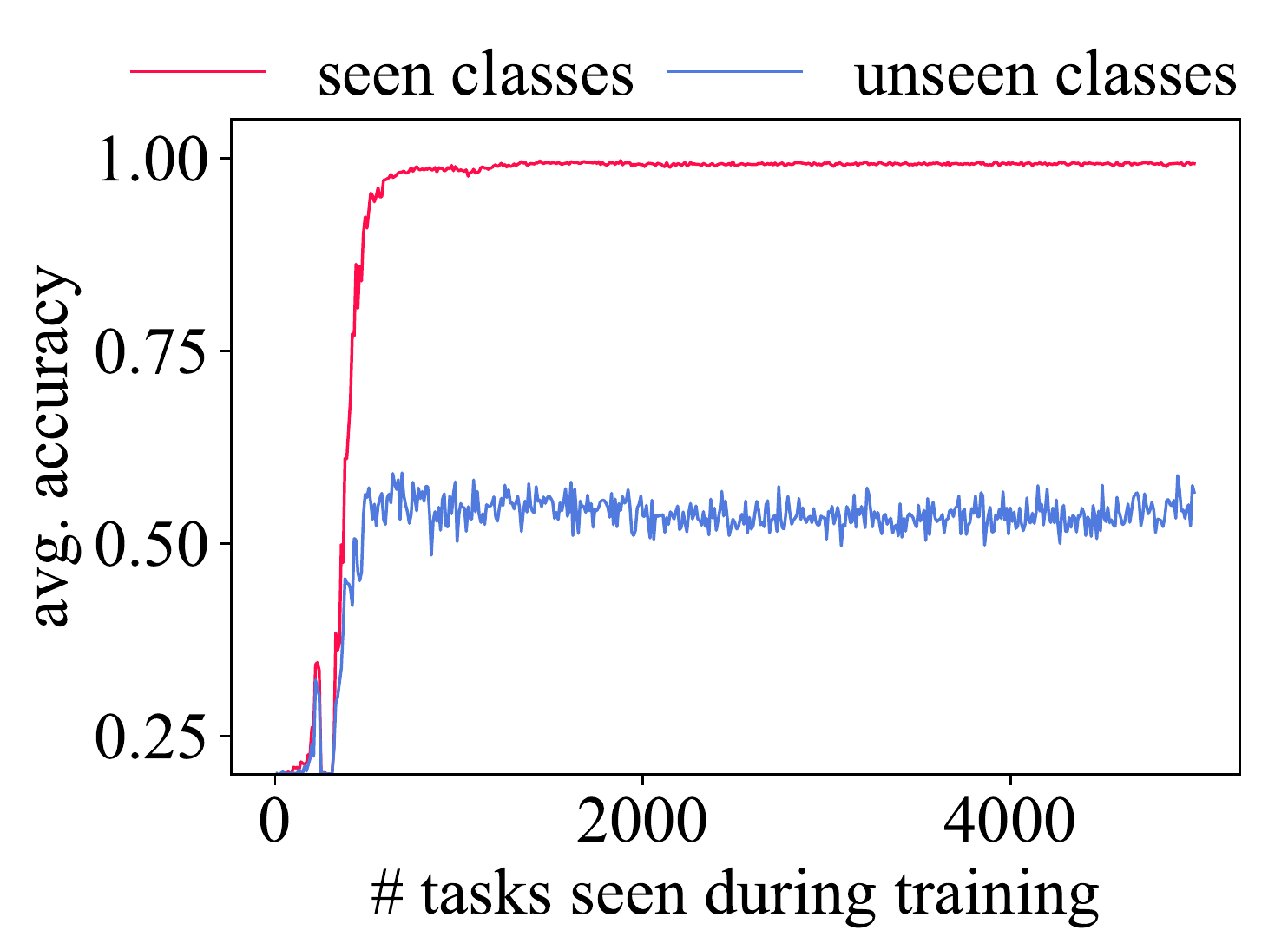}
    \vspace{-0.6cm} % account for pdf space
    \caption{Reuters}
  \end{subfigure}
  \caption{Learning curves of \textsc{induction net} on Huffpost and Reuters. We plot average 5-way 5-shot accuracy over 50 episodes sampled from seen classes (blue) and unseen classes (red).}\label{fig:learningcurve-other}
\end{figure}

\begin{figure}[t]
  \centering
%   \begin{subfigure}[t]{.32\linewidth}
%     \centering
%     \includegraphics[width=\linewidth]{figures/perplexity/amazon_perplexity.pdf}
%     \vspace{-0.6cm} % account for pdf space
%     \caption{Amazon}
%   \end{subfigure}
  \begin{subfigure}[t]{.45\linewidth}
    \centering
    \includegraphics[width=\linewidth]{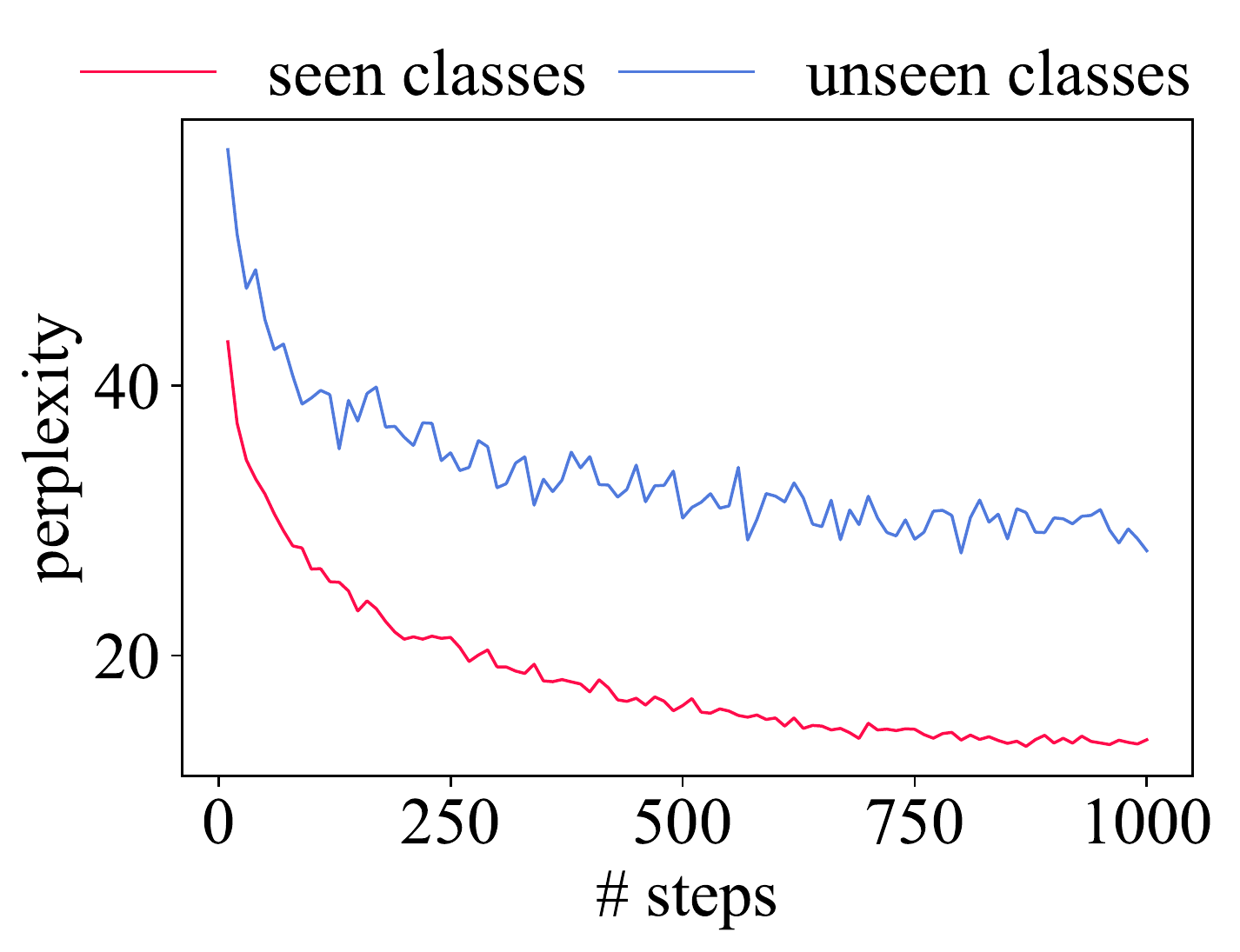}
    \vspace{-0.6cm} % account for pdf space
    \caption{HuffPost}
  \end{subfigure}
  \begin{subfigure}[t]{.45\linewidth}
    \centering
    \includegraphics[width=\linewidth]{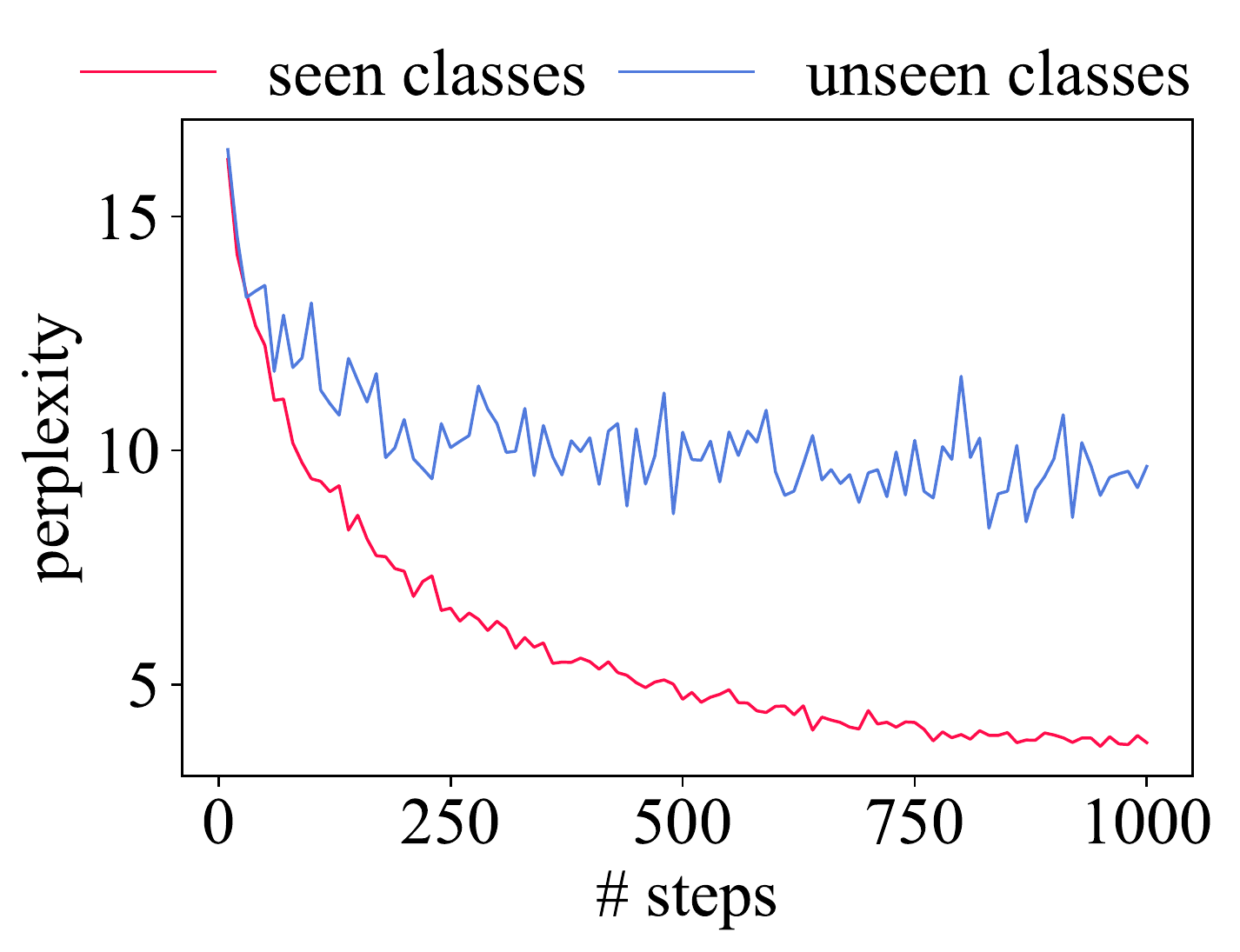}
    \vspace{-0.6cm} % account for pdf space
    \caption{Reuters}
  \end{subfigure}
  \caption{Perplexity during BERT language model finetuning in \textsc{p-maml}. The lexical distributions mismatch significantly  between meta-train and meta-test classes.}\label{fig:perplexity}
\end{figure}

Table~\ref{table:shabi} shows that learning with distributional signatures (\textsc{our}) significantly outperforms the two baselines across all datasets.
This is not surprising: both baselines build meta-knowledge on top of lexical representations, which may not generalize when the lexical distributions are vastly different between seen classes and unseen classes.
Figure~\ref{fig:learningcurve-other} shows the learning curve of \textsc{induction net}. Similar to Figure~\ref{fig:learningcurve}, we see that \textsc{induction net} overfits to meta-train classes with poor generalization to meta-test classes.
Figure~\ref{fig:perplexity} depicts the perplexity of BERT's masked language model objective, during finetuning for \textsc{p-maml}. Again, we see that the meta-train and meta-test classes exhibit vast lexical mismatch, as the perplexity improves only slightly on meta-test classes.

\begin{table*}[t]
\small
\setlength{\tabcolsep}{2.5pt}
\begin{tabular}{lcccccccccccc}
\toprule
\multirow{2}{*}{Method}&
\multicolumn{2}{c}{20 News} &
\multicolumn{2}{c}{Amazon} &
\multicolumn{2}{c}{HuffPost} &
\multicolumn{2}{c}{RCV1} &
\multicolumn{2}{c}{Reuters} &
\multicolumn{2}{c}{FewRel} \\
\cmidrule(lr{0.5em}){2-3}\cmidrule(lr{0.5em}){4-5}\cmidrule(lr{0.5em}){6-7}\cmidrule(lr{0.5em}){8-9}\cmidrule(lr{0.5em}){10-11}\cmidrule(lr{0.5em}){12-13}
&  1 shot & 5 shot & 1 shot & 5 shot & 1 shot & 5 shot & 1 shot & 5 shot & 1 shot & 5 shot & 1 shot & 5 shot \\
\midrule
\textsc{induction net} &  $27.6$ & $32.1$ & $30.6$ & $37.1$ & $34.9$ & $44.0$ & $32.3$ & $37.3$ & $58.3$ & $66.9$ & $50.4$ & $56.1$ \\
\textsc{p-maml} & --- & --- & $47.1$ & $58.4$ & $31.0$ & $51.3$ & --- & --- & $53.0$ & $72.9$ & --- & --- \\
\midrule
\textsc{our} & $\bm{52.1}$ & $\bm{68.3}$ & $\bm{62.6}$ & $\bm{81.1}$ & $\bm{43.0}$ & $\bm{63.5}$ & $\bm{54.1}$ & $\bm{75.3}$ & $\bm{81.8}$ & $\bm{96.0}$& $\bm{67.1}$ & $\bm{83.5}$ \\
\bottomrule
\end{tabular}
\centering
\caption{Comparison against Induction Net~\citep{geng2019few} and P-MAML~\citep{zhang2019improving}. We run P-MAML on text classification datasets with shorter documents, for which it is feasible to finetune BERT (after WordPiece tokenization, documents from "longer" datasets exceed BERT's max length of 512 tokens).}\label{table:shabi}
\end{table*}

\begin{figure}[t]
  \centering
  \begin{subfigure}[t]{.45\linewidth}
    \centering
    \includegraphics[width=\linewidth]{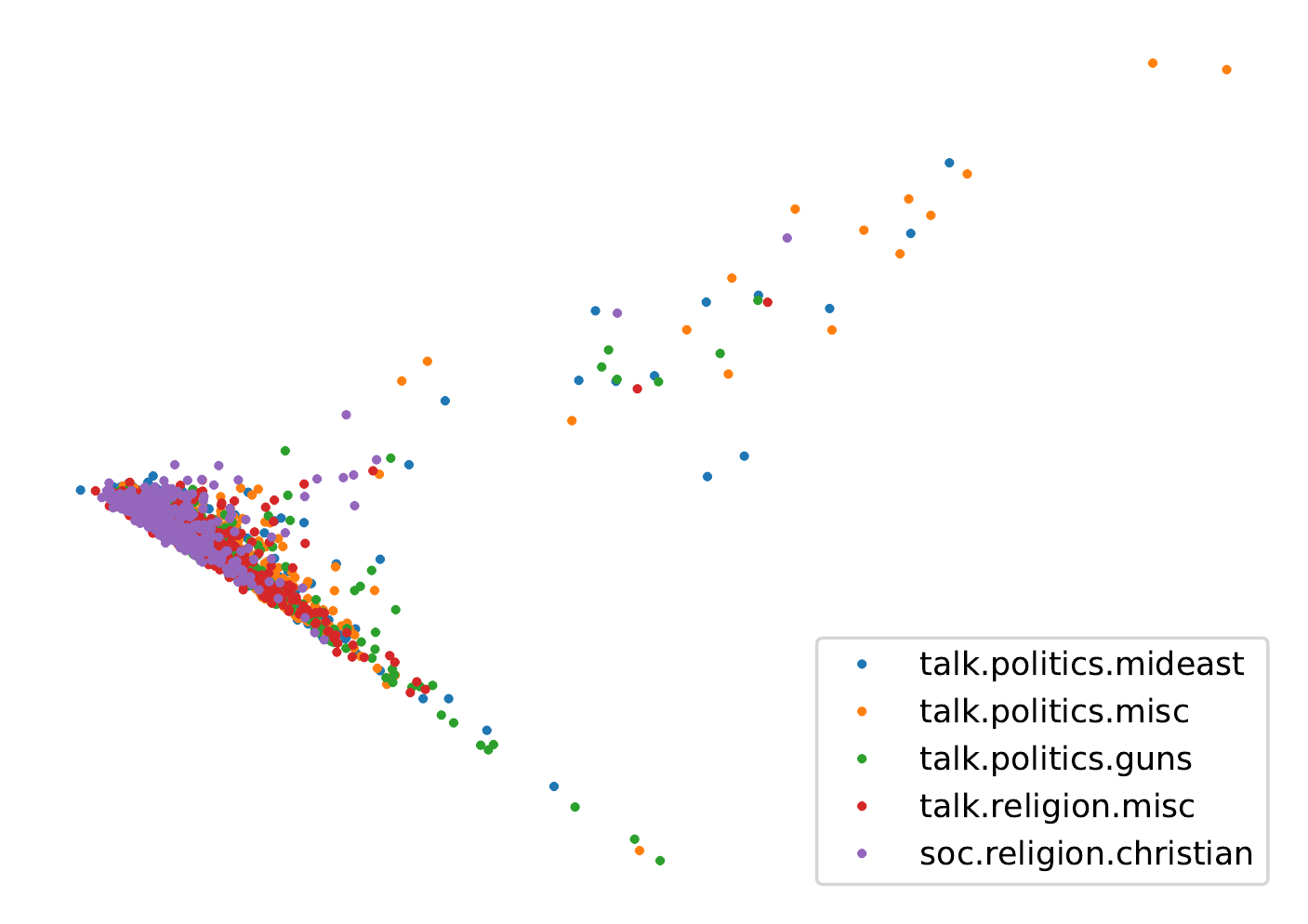}
    \caption{\textsc{avg}}
  \end{subfigure}
  \begin{subfigure}[t]{.45\linewidth}
    \centering
    \includegraphics[width=\linewidth]{figures/pca_sebd.pdf}
    \caption{$\mathrm{s}(\cdot)$}
  \end{subfigure}
  \begin{subfigure}[t]{.45\linewidth}
    \centering
    \includegraphics[width=\linewidth]{figures/pca_tebd.pdf}
    \caption{$\mathrm{t}(\cdot)$}
  \end{subfigure}
  \begin{subfigure}[t]{.45\linewidth}
    \centering
    \includegraphics[width=\linewidth]{figures/pca_meta.pdf}
    \caption{\textsc{our}}
  \end{subfigure}
  \caption{PCA visualization of the input representation for
    a testing episode in 20~Newsgroups with $N=5$, $K=5$, $L=500$
    (the query set has 500 examples per class).
    \textsc{avg}: average word embeddings.
    $\mathrm{s}(\cdot)$:
    weighted average of word embeddings
    with weights given by $\mathrm{s}(\cdot)$.
    $\mathrm{t}(\cdot)$:
    weighted average of word embeddings
    with weights given by $\mathrm{t}(\cdot)$.
    \textsc{our}:
    weighted average of word embeddings with weights
    given by the attention generator
    meta-trained on a disjoint set of training
  classes.}\label{fig:pca_20news_full}
\end{figure}

\begin{figure}[t]
  \centering
  \begin{subfigure}[t]{.45\linewidth}
    \centering
    \includegraphics[width=\linewidth]{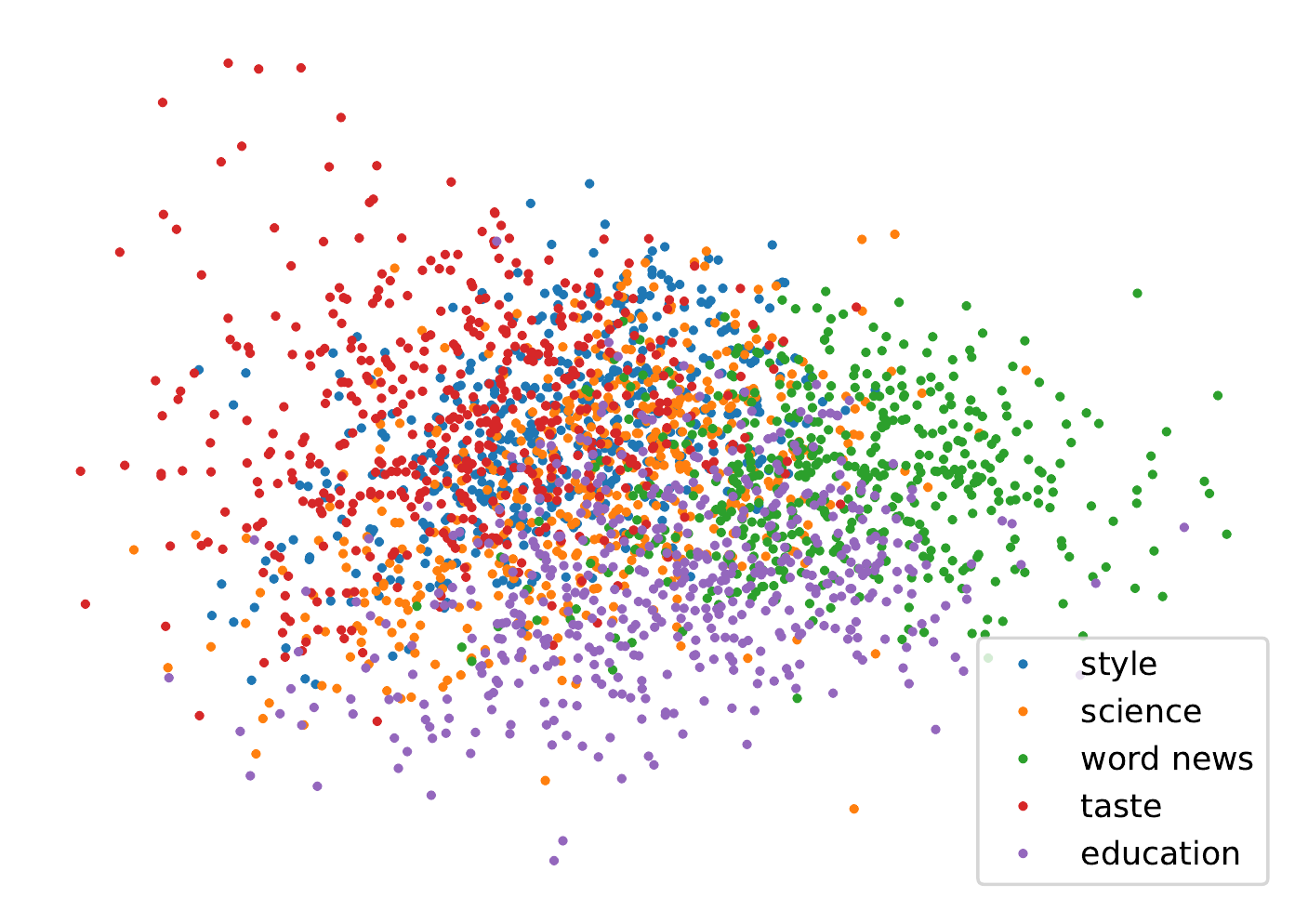}
    \caption{\textsc{avg}}
  \end{subfigure}
  \begin{subfigure}[t]{.45\linewidth}
    \centering
    \includegraphics[width=\linewidth]{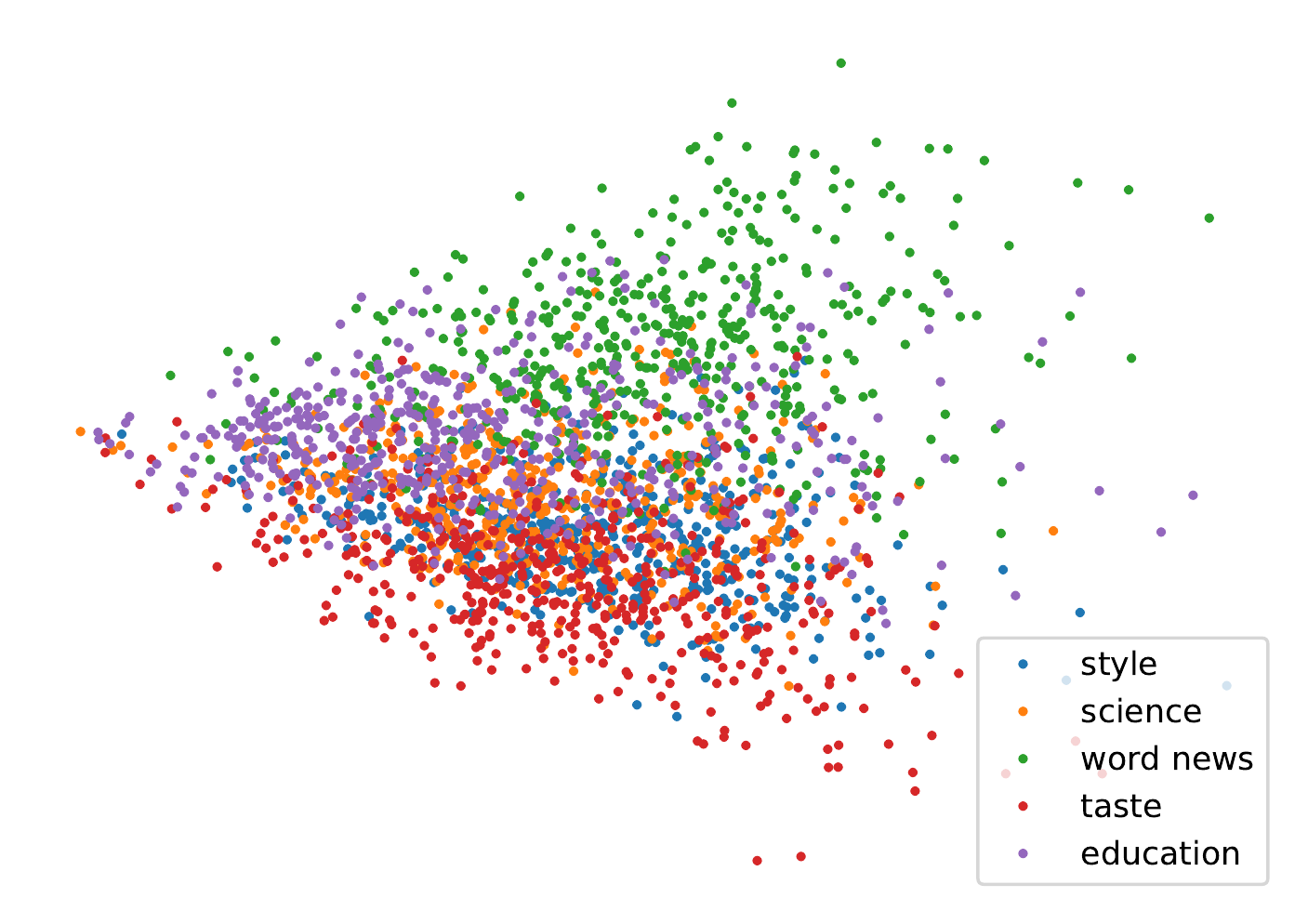}
    \caption{$\mathrm{s}(\cdot)$}
  \end{subfigure}
  \begin{subfigure}[t]{.45\linewidth}
    \centering
    \includegraphics[width=\linewidth]{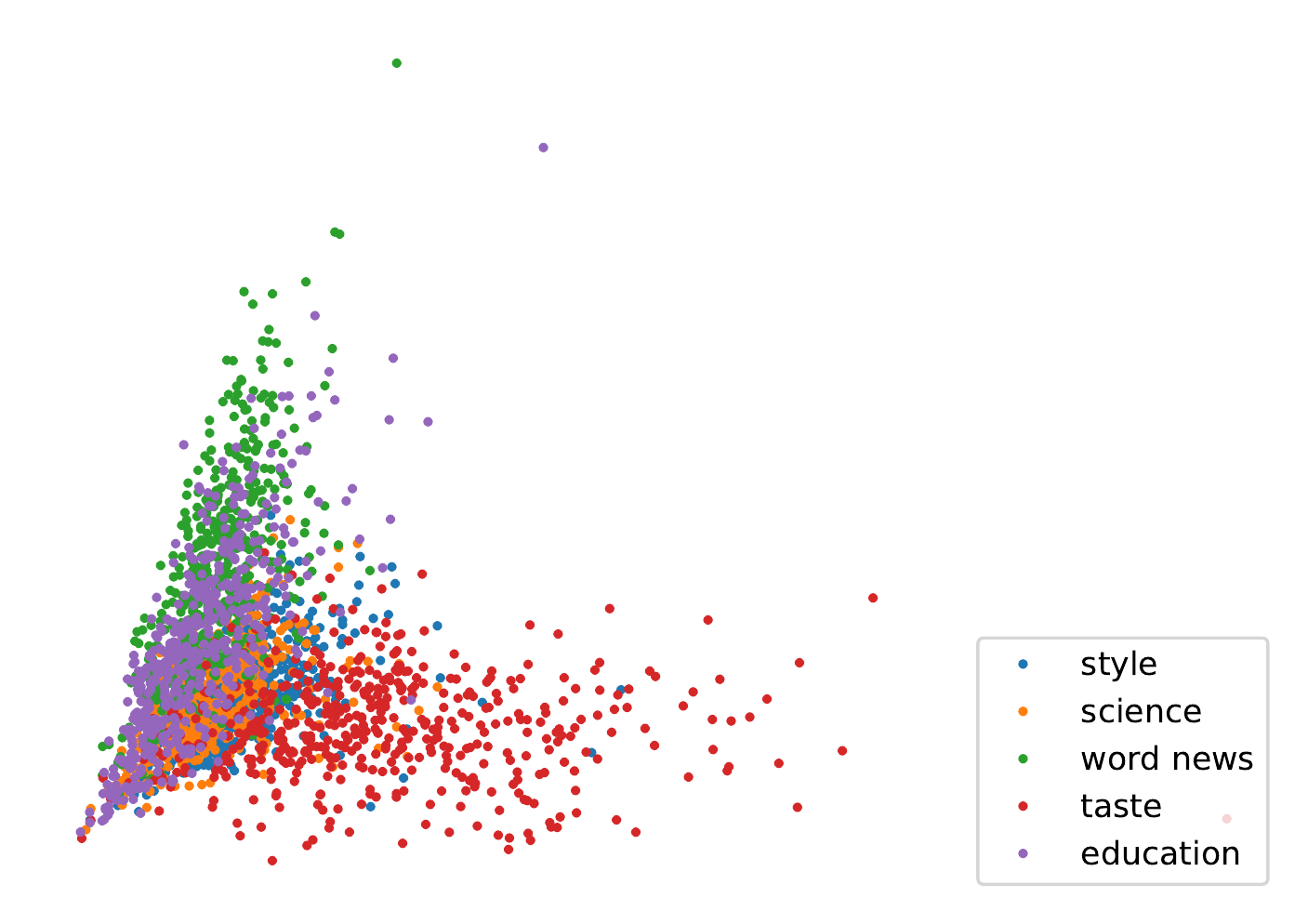}
    \caption{$\mathrm{t}(\cdot)$}
  \end{subfigure}
  \begin{subfigure}[t]{.45\linewidth}
    \centering
    \includegraphics[width=\linewidth]{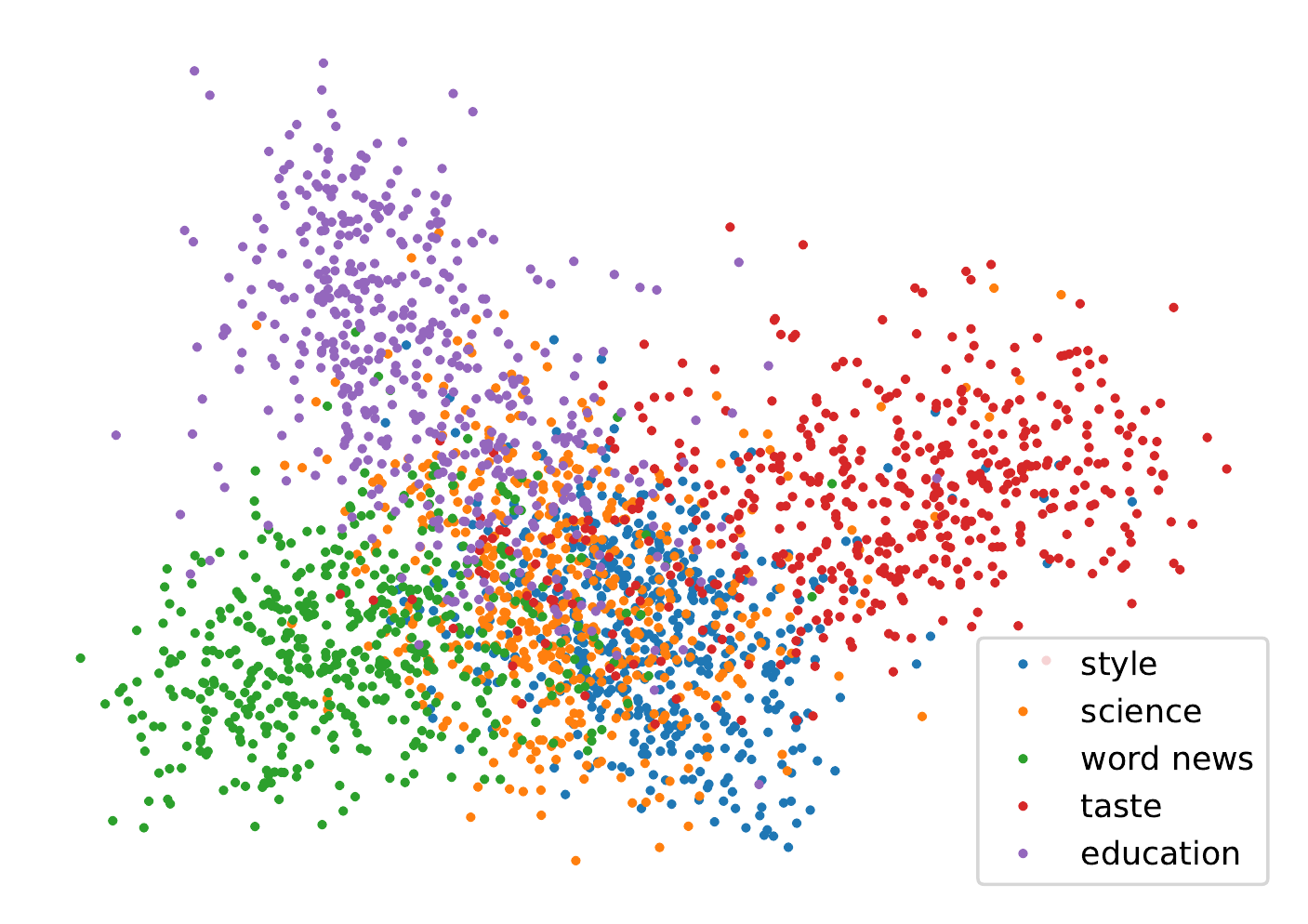}
    \caption{\textsc{our}}
  \end{subfigure}
  \caption{PCA visualization of the input representation for 
    a testing episode in HuffPost Headlines with $N=5$, $K=5$, $L=500$.
    \textsc{avg}: average word embeddings.
    $\mathrm{s}(\cdot)$:
    weighted average of word embeddings
    with weights given by $\mathrm{s}(\cdot)$.
    $\mathrm{t}(\cdot)$:
    weighted average of word embeddings
    with weights given by $\mathrm{t}(\cdot)$.
    \textsc{our}:
    weighted average of word embeddings with weights
    given by the attention generator
    meta-trained on a disjoint set of training classes.
}\label{fig:pca_huffpost}
\end{figure}

\begin{figure}[t]
  \centering
  \pgfplotsset{
    /pgfplots/bar cycle list/.style={/pgfplots/cycle list={
        {bar1,fill=bar1!80!white,mark=none},
        {bar2,fill=bar2!80!white,mark=none},
        {bar3,fill=bar3!80!white,mark=none},
        {bar4,fill=bar4!80!white,mark=none}
        }
    }
  }
  \begin{tikzpicture}
    \footnotesize
    \begin{axis}[
        ybar=0pt,
        ymin=0,
        ymax=0.4,
        width=6cm,
        height=6cm,
        bar width=0.5cm,
        enlarge x limits=0.45,
        axis lines*=left,
        ymajorgrids=true,
        xlabel={20 Newsgroups},
        ylabel={avg cosine similarity},
        xlabel near ticks,
        ylabel near ticks,
        symbolic x coords={avg,s,t,our},
        xticklabels={\textsc{avg},$\mathrm{s}(\cdot)$,$\mathrm{t}(\cdot)$,\textsc{our}},
        xtick={avg,s,t,our},
        typeset ticklabels with strut,
        nodes near coords,
        nodes near coords align={vertical},
        every node near coord/.append style={color=black},
        every axis plot/.append style={
            bar shift=0
        }
    ]
        \addplot coordinates {(avg,0.11)};
        \addplot coordinates {(s,0.21)};
        \addplot coordinates {(t,0.15)};
        \addplot coordinates {(our,0.32)};
    \end{axis}
  \end{tikzpicture}
  ~~~
  \begin{tikzpicture}
    \footnotesize
    \begin{axis}[
        ybar=0pt,
        ymin=0,
        ymax=0.8,
        width=6cm,
        height=6cm,
        bar width=0.5cm,
        enlarge x limits=0.45,
        axis lines*=left,
        ymajorgrids=true,
        xlabel={HuffPost Headlines},
        ylabel={avg cosine similarity},
        xlabel near ticks,
        ylabel near ticks,
        symbolic x coords={avg,s,t,our},
        xticklabels={\textsc{avg},$\mathrm{s}(\cdot)$,$\mathrm{t}(\cdot)$,\textsc{our}},
        xtick={avg,s,t,our},
        typeset ticklabels with strut,
        nodes near coords,
        nodes near coords align={vertical},
        every node near coord/.append style={color=black},
        every axis plot/.append style={
            bar shift=0
        }
    ]
        \addplot coordinates {(avg,0.51)};
        \addplot coordinates {(s,0.59)};
        \addplot coordinates {(t,0.70)};
        \addplot coordinates {(our,0.74)};
    \end{axis}
  \end{tikzpicture}
  \caption{Average cosine similarity to the \emph{oracle} word importance
    over the query set of a testing episode with $N=5$, $K=5$, $L=500$.
    This oracle is estimated using all labeled examples from
    the $N$ classes.
    Since examples in HuffPost Headlines are 30 times shorter,
    the cosine similarities are higher in this corpora.
    \textsc{avg}: uniform distribution over the words.
    $\mathrm{s}(\cdot)$: word importance estimated
    directly by $\mathrm{s}(\cdot)$.
    $\mathrm{t}(\cdot)$: word importance estimated
    directly by $\mathrm{t}(\cdot)$.
    \textsc{our}: word importance estimated by the meta-learned
    attention generator.
  }\label{fig:cos}
\end{figure}
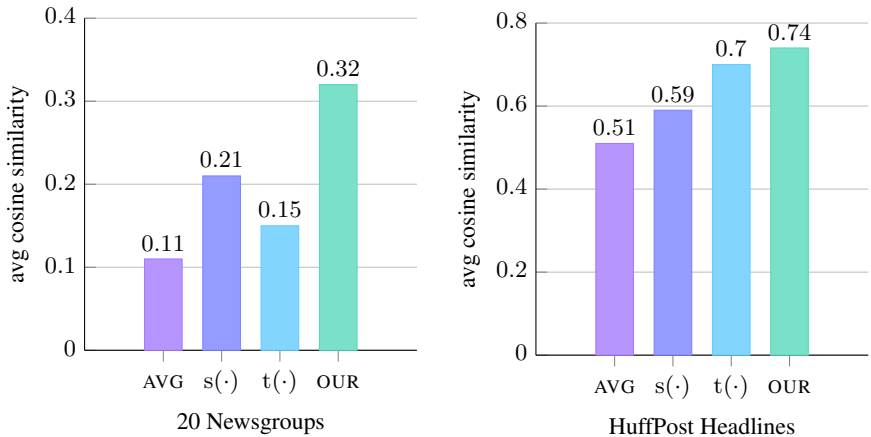

\subsection{Analysis of learned representation}\label{app:analysis}
To further understand the rationale behind our performance boost,
we provide a more detailed analysis on our model's empirical behavior.

\textbf{Visualizing the embedding space }
We visualize our attention-weghted representation $\phi(x)$
in 20~Newsgroups (Figure~\ref{fig:pca_20news_full}) and
HuffPost (Figure~\ref{fig:pca_huffpost}).
We observe that our model produces better separation
than the unweighted average \textsc{avg}
and directly using the distributional statistics, $\mathrm{s}(x)$
or $\mathrm{t}(x)$.
For instance, in 20~Newsgroups, our model recognizes three clusters:
\{\emph{talk.religion.misc, soc.religion.christian}\},
\{\emph{talk.politics.mideast}\} and
\{\emph{talk.politics.misc, talk.politics.guns}\},

\textbf{Cosine similarity to \emph{oracle} word importance }
We also quantitatively analyze the generated attention in
in 20~Newsgroups and HuffPost (Figure~\ref{fig:cos}).

To obtain a more reliable estimate of word importance,
we train an \emph{oracle} model
over all examples from the $N$ target classes.
This oracle model uses a $\mathrm{biLSTM}$
to encode each example from its word embeddings.
It then generates an attention score based on this encoding.
In order to estimate the importance of individual unigram,
we use this attention score to weight the ``original'' word embeddings
(not the output of the $\mathrm{biLSTM}$).
A MLP with one hidden layer is used to perform $N$-way classification
from the attention-weighted representation.

Compared to both $\mathrm{s}(\cdot)$ and $\mathrm{t}(\cdot)$,
the attention generated by our model is much closer to the
oracle's,
which explains our model's large performance gains.
Note that the attention generator
does not see any examples from the target classes during meta-training.

%\textbf{Adding lexical information may harm generalization }
%Figure~\ref{fig:learningcurve_rcv1} compares
%the learning curve of our model with its lexicalized counterparts.
%We observe that
%when word embeddings are accessible to the attention generator,
%it can utilize them to overfit the training classes easily.
%However, this knowledge fails to generalize to new classes.

%\begin{figure*}[t]
%  \centering
%  \begin{subfigure}[t]{.32\linewidth}
%    \includegraphics[width=\linewidth]{figures/rcv1_learningcurve_our.pdf}
%    \caption{Input: $\mathrm{s}(x)$, $\mathrm{t}(x)$}
%  \end{subfigure}
%  \begin{subfigure}[t]{.32\linewidth}
%    \includegraphics[width=\linewidth]{figures/rcv1_learningcurve_our_w_ebd.pdf}
%    \caption{Input: $\mathrm{s}(x)$, $\mathrm{t}(x)$, $f_\mathrm{ebd}(x)$}
%  \end{subfigure}
%  \begin{subfigure}[t]{.32\linewidth}
%    \includegraphics[width=\linewidth]{figures/rcv1_learningcurve_ebd.pdf}
%    \caption{Input: $f_\mathrm{ebd}(x)$}
%  \end{subfigure}
%  \caption{
%    Comparison between our delexicalized attention generator (a) and
%    its lexicalized counterparts (bc) in RCV1.
%    The subcaption details the input to the attention generator.
%    We plot the avg. 5-way 1-shot accuracy over 50 episodes sampled from
%    seen classes (blue) and unseen classes (red).
%    The meta-learner learns a more generalizable representation
%    when its input is delexicalized.
%  }
%  \label{fig:learningcurve_rcv1}
%\end{figure*}

\textbf{Visualizing the generated attention }
Figure~\ref{fig:app_huffpost} visualizes
the generated attention for a testing episode in HuffPost.
We observe that our model identifies meaningful keywords from the sentence.

\begin{table*}[t!]
	\centering
	\small
	\renewcommand{\hlr}[2]{\setlength{\fboxsep}{.3pt}\colorbox{red!#2}{\rule[-1mm]{0pt}{4mm}{#1}}}
	\renewcommand{\arraystretch}{1.5}
	\begin{tabular}{ll}
	  \toprule
	  class & input example\\
		\midrule
		taste& \hlr{you}{13} \hlr{wo}{33} \hlr{n't}{13} \hlr{even}{14} \hlr{miss}{14} \hlr{the}{1} \hlr{meat}{89} \hlr{with}{16} \hlr{these}{13} \hlr{delicious}{92} \hlr{vegetarian}{100} \hlr{sandwiches}{86} \\
		taste& \hlr{these}{2} \hlr{cookies}{100} \hlr{are}{0} \hlr{spot}{6} \hlr{-}{0} \hlr{on}{0} \hlr{copies}{21} \hlr{of}{0} \hlr{the}{0} \hlr{oscars}{13} \hlr{dresses}{12} \\
		\midrule
		word news& \hlr{prime}{12} \hlr{minister}{16} \hlr{saad}{0} \hlr{hariri}{0} \hlr{'s}{0} \hlr{return}{5} \hlr{to}{0} \hlr{lebanon}{100} \hlr{:}{0} \hlr{a}{0} \hlr{moment}{7} \hlr{of}{1} \hlr{truth}{13} \\
		word news& \hlr{new}{7} \hlr{zealand}{100} \hlr{just}{2} \hlr{became}{12} \hlr{the}{0} \hlr{11th}{0} \hlr{country}{30} \hlr{to}{0} \hlr{send}{6} \hlr{a}{0} \hlr{rocket}{11} \hlr{into}{4} \hlr{orbit}{0} \\
		\midrule
		style& \hlr{beyonc\'{e}}{100} \hlr{dressed}{17} \hlr{like}{5} \hlr{the}{0} \hlr{queen}{7} \hlr{she}{7} \hlr{is}{0} \hlr{at}{2} \hlr{the}{0} \hlr{grammys}{41} \\
		style& \hlr{tilda}{0} \hlr{swinton}{0} \hlr{,}{1} \hlr{is}{0} \hlr{that}{1} \hlr{a}{0} \hlr{jacket}{100} \hlr{or}{7} \hlr{a}{0} \hlr{dress}{37} \hlr{?}{9} \\
		\midrule
		science& \hlr{the}{4} \hlr{world}{35} \hlr{of}{6} \hlr{science}{100} \hlr{has}{14} \hlr{a}{1} \hlr{lot}{30} \hlr{to}{0} \hlr{look}{19} \hlr{forward}{44} \hlr{to}{1} \hlr{in}{11} \hlr{2016}{40} \\
		science& \hlr{dione}{0} \hlr{crosses}{0} \hlr{saturn}{100} \hlr{'s}{1} \hlr{disk}{0} \hlr{in}{2} \hlr{spectacular}{39} \hlr{new}{21} \hlr{image}{37} \\
		\midrule
		education& \hlr{the}{0} \hlr{global}{14} \hlr{search}{22} \hlr{for}{2} \hlr{education}{100} \hlr{:}{0} \hlr{ }{2} \hlr{just}{2} \hlr{imagine}{59} \hlr{secretary}{26} \hlr{hargreaves}{0} \\
		education& \hlr{thinking}{20} \hlr{at}{1} \hlr{harvard}{19} \hlr{:}{0} \hlr{what}{1} \hlr{is}{0} \hlr{the}{0} \hlr{future}{5} \hlr{of}{1} \hlr{learning}{100} \hlr{?}{11} \\
		\bottomrule
	\end{tabular}
	\captionof{figure}{
	  Visualization of the attention generated by our model
	  on 10 query examples from
	  a $5$-way $5$-shot testing episode
	  in Huffpost Headlines.
	}\label{fig:app_huffpost}
\end{table*}

\subsection{Effect of softmax calibration}

In Section~\ref{sec:r2d2}, we use ridge regression with softmax calibration as our downstream predictor due to its efficiency and effectiveness.
To study the effect of the softmax calibration, we follow~\citet{bertinetto2018metalearning} and compare ridge regression against direct optimization of binary logistic classifiers (\textsc{lr}) with Newton's method (five iterations).
Since Newton's method admits a closed-form solution at each iteration,
the final solution is also end-to-end differentiable.
For each $N$-way $K$-shot episode,
to adapt the binary logistic classifier for multi-class prediction,
we train $N$ binary (one-vs-all) classifiers using the support set.
The output of the $N$ binary classifiers are combined to make the final predictions on the query set~(\citeauthor{bishop2006pattern}, \citeyear{bishop2006pattern}, Chapter 4.1.2).

The results are shown in Table~\ref{table:rr-vs-lr}.
Overall, \textsc{rr} and \textsc{lr} produce similar results (78.0 vs. 77.1 on 5-way 5-shot, 60.1 vs. 58.5 on 5-way 1-shot, respectively),
though \textsc{rr} performs slightly better.
When using \textsc{lr} as the downstream predictor,
learning with distributional signatures improves 5-way 1-shot accuracy by
6.1\% and 5-way 5-shot accuracy by 3.2\% over the best baseline.
Note that \textsc{lr} requires solving $N$ binary classifications for each episode, which is less efficient than \textsc{rr}.

\begin{table*}[t]
\small
\setlength{\tabcolsep}{1.1pt}
\begin{tabular}{llcccccccccccccc}
\toprule
\multicolumn{2}{c}{Method} &
\multicolumn{2}{c}{20 News} &
\multicolumn{2}{c}{Amazon} &
\multicolumn{2}{c}{HuffPost} &
\multicolumn{2}{c}{RCV1} &
\multicolumn{2}{c}{Reuters} &
\multicolumn{2}{c}{FewRel} &
\multicolumn{2}{c}{Average}
\\
\cmidrule(lr{0.5em}){1-2}\cmidrule(lr{0.5em}){3-4}\cmidrule(lr{0.5em}){5-6}\cmidrule(lr{0.5em}){7-8}\cmidrule(lr{0.5em}){9-10}\cmidrule(lr{0.5em}){11-12}\cmidrule(lr{0.5em}){13-14}
\cmidrule(lr{0.5em}){15-16}
Rep.\hspace{5mm} & Alg. & 1 shot & 5 shot &1 shot & 5 shot &1 shot & 5 shot &1 shot & 5 shot &1 shot & 5 shot &1 shot & 5 shot & 1 shot & 5 shot \\
\midrule
\textsc{avg} & \textsc{rr}        & $37.6$ &  $57.2$ & $50.2$ &  $72.7$ & $36.3$ &  $54.8$ & $48.1$ &  $72.6 $ &  $63.4$ & $90.0$& $53.2$ &  $72.2$ & $48.1$ & $69.9$ \\
\textsc{idf} & \textsc{rr}        & $44.8$ &  $64.3$ & $60.2$ &  $79.7$ & $37.6$ &  $59.5$ & $48.6$ &  $72.8 $ &  $69.1$ & $93.0$& $55.6$ &  $75.3$ & $52.6$ & $74.1$ \\
\textsc{cnn} & \textsc{rr}        & $32.2$ &  $44.3$ & $37.3$ &  $53.8$ & $37.3$ &  $49.9$ & $41.8$ &  $59.4 $ &  $71.4$ & $87.9$& $56.8$ &  $71.8$ & $46.1$ & $61.2$ \\
\textsc{our} & \textsc{rr} & $\bm{52.1}$ & $\bm{68.3}$ & $\bm{62.6}$ & $\bm{81.1}$ & $\bm{43.0}$ & $\bm{63.5}$ & $54.1$ & $\bm{75.3}$ & $\bm{81.8}$ & $\bm{96.0}$& $\bm{67.1}$ & $\bm{83.5}$ & $\bm{60.1}$ & $\bm{78.0}$\\
\midrule
\textsc{avg} & \textsc{lr} &  $37.7$ & $56.8$ & $50.0$ & $71.6$ & $36.3$ & $56.0$ & $47.9$ & $72.5$ & $62.7$ & $89.3$ & $52.4$ & $72.9$ & $47.8$ & $69.9$ \\
\textsc{idf} & \textsc{lr} &  $46.1$ & $64.2$ & $60.5$ & $79.9$ & $38.7$ & $59.3$ & $47.3$ & $72.5$ & $68.1$ & $93.0$ & $55.5$ & $74.9$ & $52.7$ & $74.0$ \\
\textsc{cnn} & \textsc{lr} &  $32.6$ & $47.5$ & $41.0$ & $58.5$ & $37.6$ & $50.9$ & $44.2$ & $63.6$ & $75.0$ & $90.3$ & $58.7$ & $72.1$ & $48.2$ & $63.8$ \\
\textsc{our} & \textsc{lr} &  $49.7$ & $67.6$ & $61.6$ & $80.7$ & $41.9$ & $61.9$ & $\bm{54.4}$ & $74.1$ & $79.5$ & $95.6$ & $65.8$ & $82.8$ & $58.8$ & $77.1$ \\
\bottomrule
\end{tabular}
\centering
\caption{Results of 5-way 1-shot and 5-way 5-shot classification on six datasets using ridge regression vs. logistic regression with Newton's method.}
\label{table:rr-vs-lr}
\end{table*}

\subsection{Finetuning word embeddings during meta-training}\label{app:finetune}
In Table~\ref{table:bigtable},
we fix word embeddings during meta-training for all experiments. Table~\ref{tab:fix_vs_finetune} studies the effect of finetuning word embeddings during meta-training.
We observe that when the word embeddings are finetuned,
performance drops for nearly all models.

To intuitively understand this behavior,
we compare the meta-train vocabulary with the meta-test vocabulary.
On Amazon, 15095 of 28591 (52.8\%) tokens in the meta-train vocabulary are not present in meta-test, and 21049 of 34545 (60.9\%) tokens in the meta-test vocabulary are not present in meta-train. On Reuters, 3604 of 6372 (56.6\%) meta-train tokens are not in meta-test, while 2481 of 5249 (47.2\%) meta-test tokens are not in meta-test.
Due to this lexical mismatch, finetuning will destroy the original geometry of the pretrained word embeddings, leading to poor generalization on unseen tasks.

\begin{table*}[t]
\small
\setlength{\tabcolsep}{3pt}
\begin{tabular}{llcccc}
\toprule
\multicolumn{2}{c}{Method} & \multicolumn{2}{c}{5-way 5-shot Reuters}  & \multicolumn{2}{c}{5-way 5-shot Amazon} \\
\cmidrule(lr{0.5em}){1-2}\cmidrule(lr{0.5em}){3-4}\cmidrule(lr{0.5em}){5-6}
Rep.\hspace{5mm} & Alg. &  Fix embedding & Finetune embedding &  Fix embedding & Finetune embedding \\
\midrule
\textsc{avg} & \textsc{proto} & $68.15$ {\tiny $\pm 1.53$ } & $58.40$ {\tiny $\pm 2.84$ } & $51.99 $ {\tiny $ \pm 2.72$} & $41.34$ {\tiny $\pm 1.61$ } \\
\textsc{idf} & \textsc{proto} & $72.07$ {\tiny $\pm 2.69$ } & $65.93$ {\tiny $\pm 3.08$ } & $59.24 $ {\tiny $ \pm 1.10$} & $48.37$ {\tiny $\pm 2.41$ } \\
\textsc{cnn} & \textsc{proto} & $74.33$ {\tiny $\pm 1.86$ } & $74.20$ {\tiny $\pm 2.85$ } & $44.49 $ {\tiny $ \pm 1.53$} & $44.08$ {\tiny $\pm 2.21$ } \\
\midrule
\textsc{avg} & \textsc{maml} & $62.46$ {\tiny $\pm 0.58$ } & $50.59$ {\tiny $\pm 1.88$ } & $47.22$ {\tiny $\pm 1.99$ } & $30.95$ {\tiny $\pm 1.08$ } \\
\textsc{idf} & \textsc{maml} & $71.96$ {\tiny $\pm 1.48$ } & $68.76$ {\tiny $\pm 1.39$ } & $62.45$ {\tiny $\pm 1.33$ } & $50.63$ {\tiny $\pm 0.61$ } \\
\textsc{cnn} & \textsc{maml} & $85.00$ {\tiny $\pm 0.76$ } & $84.19$ {\tiny $\pm 0.41$ } & $43.70$ {\tiny $\pm 1.38$ } & $43.96$ {\tiny $\pm 1.19$ } \\
\midrule
\textsc{avg} & \textsc{rr} & $90.00$ {\tiny $\pm 0.49$ } &  $86.87$ {\tiny $\pm 0.60$ } & $72.78 $ {\tiny $ \pm 0.23$} & $75.65$ {\tiny $\pm 0.51$ } \\
\textsc{idf} & \textsc{rr} & $93.02$ {\tiny $\pm 0.45$ } &  $93.01$ {\tiny $\pm 0.47$ } & $79.78 $ {\tiny $ \pm 0.28$} & $78.73$ {\tiny $\pm 0.59$ } \\
\textsc{cnn} & \textsc{rr} & $87.93$ {\tiny $\pm 1.49$ }  & $85.47$ {\tiny $\pm 0.57$ } & $53.89 $ {\tiny $ \pm 1.54$} & $52.63$ {\tiny $\pm 2.02$ } \\
\midrule
\textsc{our} & \textsc{rr} & $96.00$ {\tiny $ \pm 0.27$} & $95.98$ {\tiny $\pm 0.39$ } & $81.16 $ {\tiny $ \pm 0.31$} & $78.94$ {\tiny $\pm 0.56$ }\\
\bottomrule
\end{tabular}
\centering
\caption{Fixed vs. finetuned word embeddings during meta-training. Since meta-train and meta-test classes exhibit different word distributions, fixing the word embedding yields better generalization. That is, we avoid overfitting to the meta-train vocabulary and destroying the geometry of the pre-trained embeddings.}\label{tab:fix_vs_finetune}
\end{table*}

\subsection{Distributional signatures with other classifiers}\label{app:combine}
In this paper, we demonstrate the advantage of learning meta-knowledge on top of distributional signatures.
We focus on a ridge regressor~\citep{bertinetto2018metalearning} as our downstream classifier due to its simplicity and effectiveness.
However, the idea of learning with distributional signatures is not limited to ridge regression;
rather, we can combine distributional signatures with other learning methods such as prototypical networks~\citep{snell2017prototypical} and induction networks~\citep{geng2019few}.

\textbf{Prototypical networks } To augment a prototypical network with features learned from
distributional signatures,
we can construct per-class prototypes based on the
attention-weighted representation $\phi(x)$ (Section \ref{sec:r2d2}).
From Table~\ref{tab:proto},
we see that learning with distributional signatures
improves 5-way 1-shot accuracy by $9.9$\%
and 5-way 5-shot accuracy by $16.7$\%,
against the best prototypical network baseline for each dataset.

\textbf{Induction networks} We can also augment induction networks with meta-knowledge learned from distributional signatures.
Specifically,
we directly feed our attention-weighted representation $\phi(x)$ (Section \ref{sec:r2d2})
to the induction module.
To avoid over-fitting on meta-train features,
we replace the meta-learned relation module (see ~\citet{geng2019few} for details) by a parameter-free nearest neighbour predictor, similar to that of prototypical networks.
From Table~\ref{tab:proto},
we again see that learning with distributional signatures
improves 5-way 1-shot accuracy by $14.3$\%
and 5-way 5-shot accuracy by $25.1$\% on average.

These empirical results clearly demonstrate the benefit of our approach.
However, both prototypical networks and induction networks
perform consistently worse than the ridge regressor across all experiments.
Our hypothesis is that the ridge regressor enables task-specific finetuning based on the support set,
while the prototypical network and induction network directly compute nearest neighbours from the meta-learned metric space.

\begin{table}[t]
\small
\setlength{\tabcolsep}{1.1pt}
\begin{tabular}{llcccccccccccccc}
\toprule
\multicolumn{2}{c}{Method} &
\multicolumn{2}{c}{20 News} &
\multicolumn{2}{c}{Amazon} &
\multicolumn{2}{c}{HuffPost} &
\multicolumn{2}{c}{RCV1} &
\multicolumn{2}{c}{Reuters} &
\multicolumn{2}{c}{FewRel} &
\multicolumn{2}{c}{Average}
\\
\cmidrule(lr{0.5em}){1-2}\cmidrule(lr{0.5em}){3-4}\cmidrule(lr{0.5em}){5-6}\cmidrule(lr{0.5em}){7-8}\cmidrule(lr{0.5em}){9-10}\cmidrule(lr{0.5em}){11-12}\cmidrule(lr{0.5em}){13-14}
\cmidrule(lr{0.5em}){15-16}
Rep.\hspace{2mm} & Alg. & 1 shot & 5 shot &1 shot & 5 shot &1 shot & 5 shot &1 shot & 5 shot &1 shot & 5 shot &1 shot & 5 shot & 1 shot & 5 shot \\
\midrule
\textsc{avg} &\textsc{proto}     &$36.2$ &  $45.4$ & $37.2$ &  $51.9$ & $35.6$ &  $41.6$ & $28.4$ &  $31.2 $ & $59.5$ & $68.1$ & $44.0$ &  $46.5$ & $40.1$ & $47.4$ \\
\textsc{idf} &\textsc{proto}     &$37.8$ &  $46.5$ & $41.9$ &  $59.2$ & $34.8$ &  $50.2$ & $32.1$ &  $35.6 $ & $61.0$ & $72.1$& $43.0$ &  $61.9$ & $41.8$ & $54.2$ \\
\textsc{cnn} &\textsc{proto}     &$29.6$ &  $35.0$ & $34.0$ &  $44.4$ & $33.4$ &  $44.2$ & $28.4$ &  $29.3 $ &  $65.2$ & $74.3$ & $49.7$ &  $65.1$ & $40.1$ & $48.7$  \\
\textsc{our}    & \textsc{proto}    &$\bf{42.4}$ & $\bf{59.5}$ & $\bf{52.6}$ & $\bf{73.9}$ & $\bf{40.6}$ & $\bf{58.9}$ & $\bf{45.1}$ & $\bf{63.6}$ & $\bf{72.2}$ & $\bf{93.9}$ & $\bf{57.5}$ & $\bf{75.3}$ & $\bf{51.7}$ & $\bf{70.9}$ \\
\midrule
\textsc{lstm} & \textsc{induct}  &  $27.6$ & $32.1$ & $30.6$ & $37.1$ & $34.9$ & $44.0$ & $32.3$ & $37.3$ & $58.3$ & $66.9$ & $50.4$ & $56.1$ & $39.0$ & $45.6$  \\
%\textsc{our} & \textsc{induct}  &  $\bf{31.5}$ & $\bf{41.3}$ & $\bf{34.8}$ & $\bf{45.2}$ & $\bf{34.5}$ & $\bf{52.1}$ & $\bf{38.7}$ & $\bf{54.0}$ & $\bf{74.5}$ & $\bf{93.1}$ & $\bf{50.2}$ & $\bf{66.2}$ & $\bf{44.0}$ & $\bf{58.7}$  \\
\textsc{our} & \textsc{induct} & $\bf{45.4}$ & $\bf{60.0}$ & $\bf{56.6}$ & $\bf{73.5}$ & $\bf{40.4}$ & $\bf{60.2}$ & $\bf{42.3}$ & $\bf{60.3}$ & $\bf{74.5}$ & $\bf{93.1}$ & $\bf{60.5}$ & $\bf{77.3}$ & $\bf{53.3}$ & $\bf{70.7}$ \\ 
\bottomrule
\end{tabular}
\centering
\caption{Performance of prototypical networks and induction networks learned on lexical information vs. distributional signatures (\textsc{our}). \textsc{lstm} with \textsc{induct} is the original induction network architecture.}
\label{tab:proto}
\end{table}

\subsection{Few-shot learning v.s. supervised learning}
\begin{figure}[t]
  \centering
  \begin{subfigure}[t]{.45\linewidth}
    \centering
    \includegraphics[width=\linewidth]{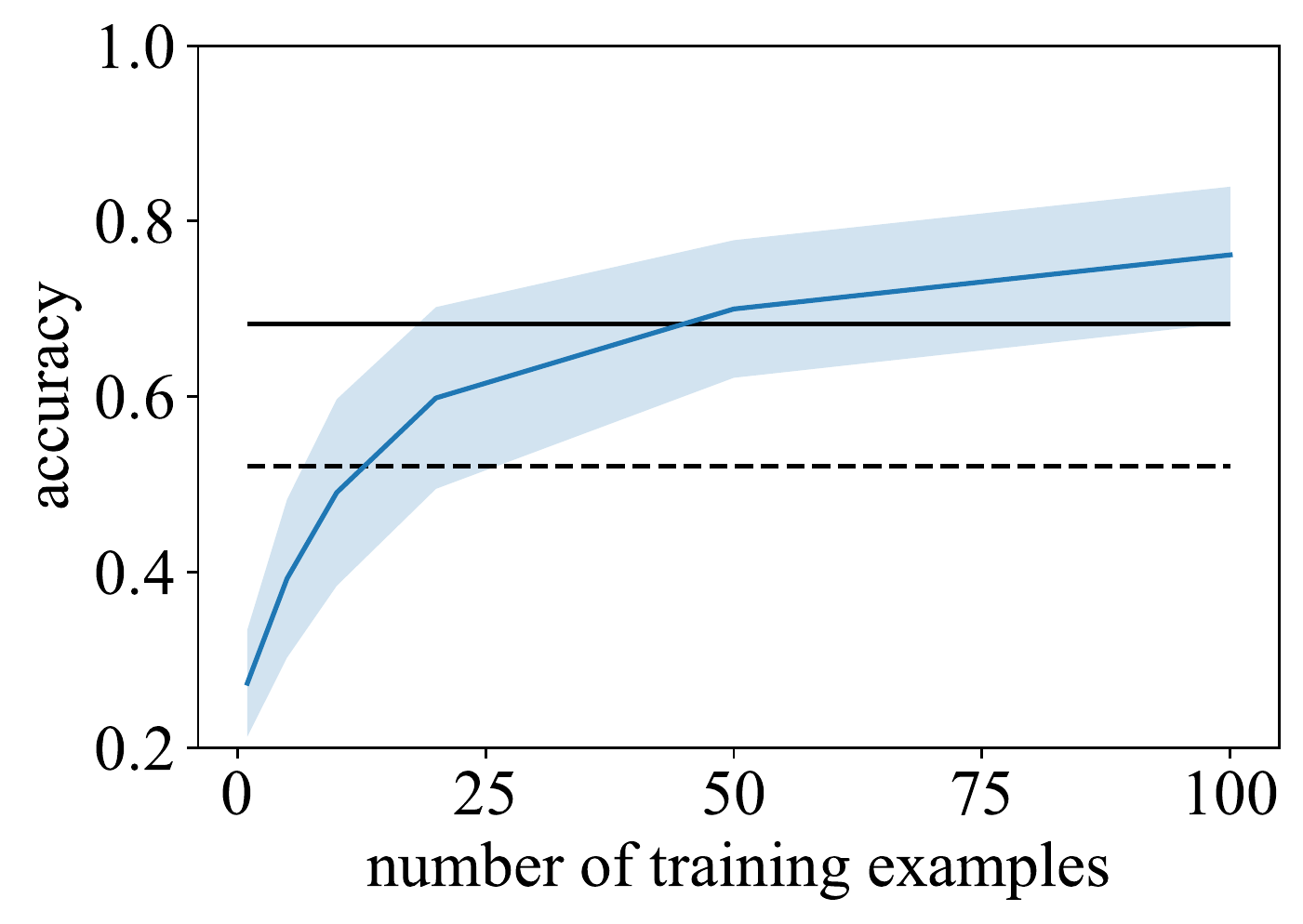}
    \vspace{-0.6cm} % account for pdf space
    \caption{20Newsgroups}
  \end{subfigure}
  \begin{subfigure}[t]{.45\linewidth}
    \centering
    \includegraphics[width=\linewidth]{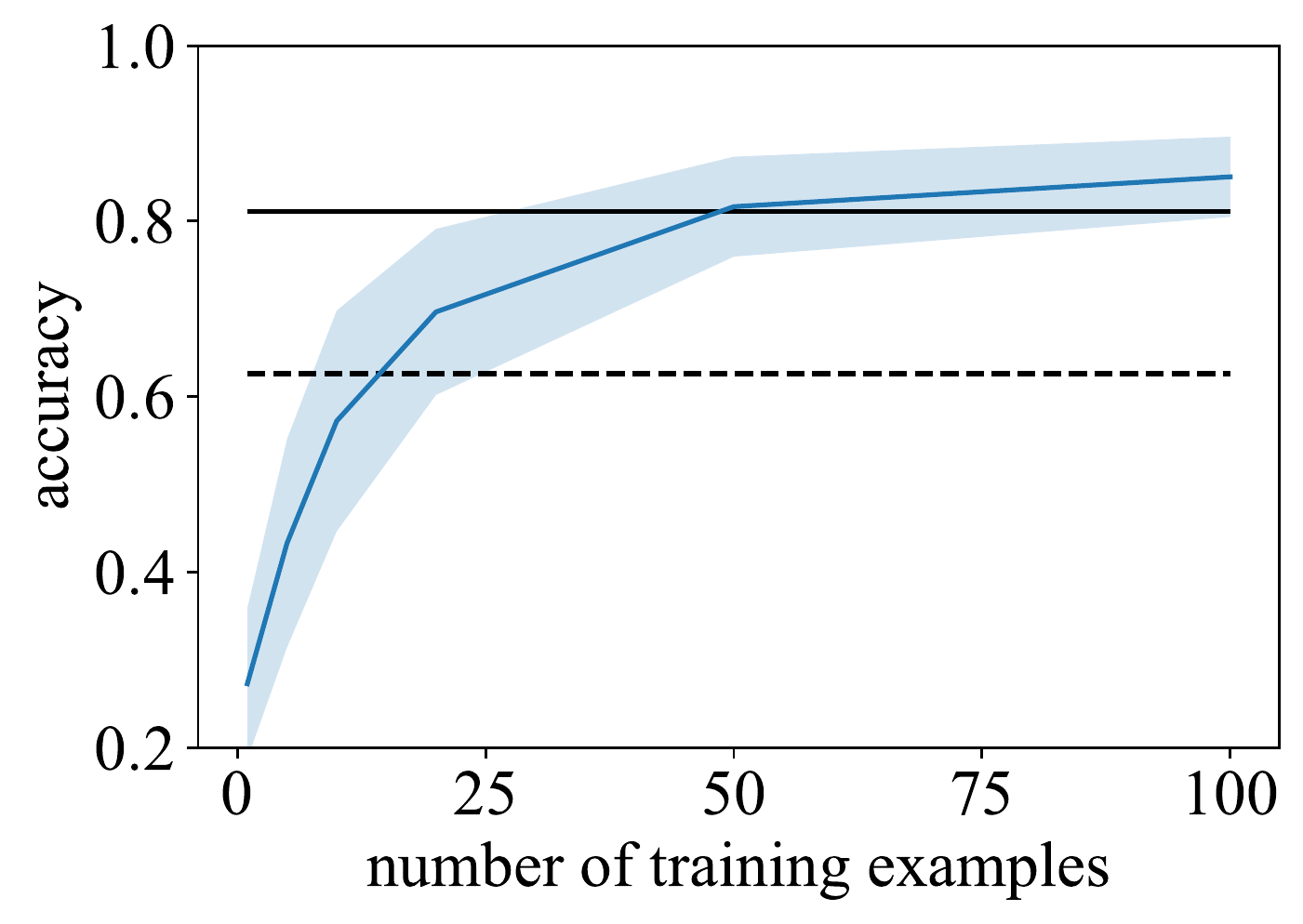}
    \vspace{-0.6cm} % account for pdf space
    \caption{Amazon}
  \end{subfigure}
  \begin{subfigure}[t]{.45\linewidth}
    \centering
    \includegraphics[width=\linewidth]{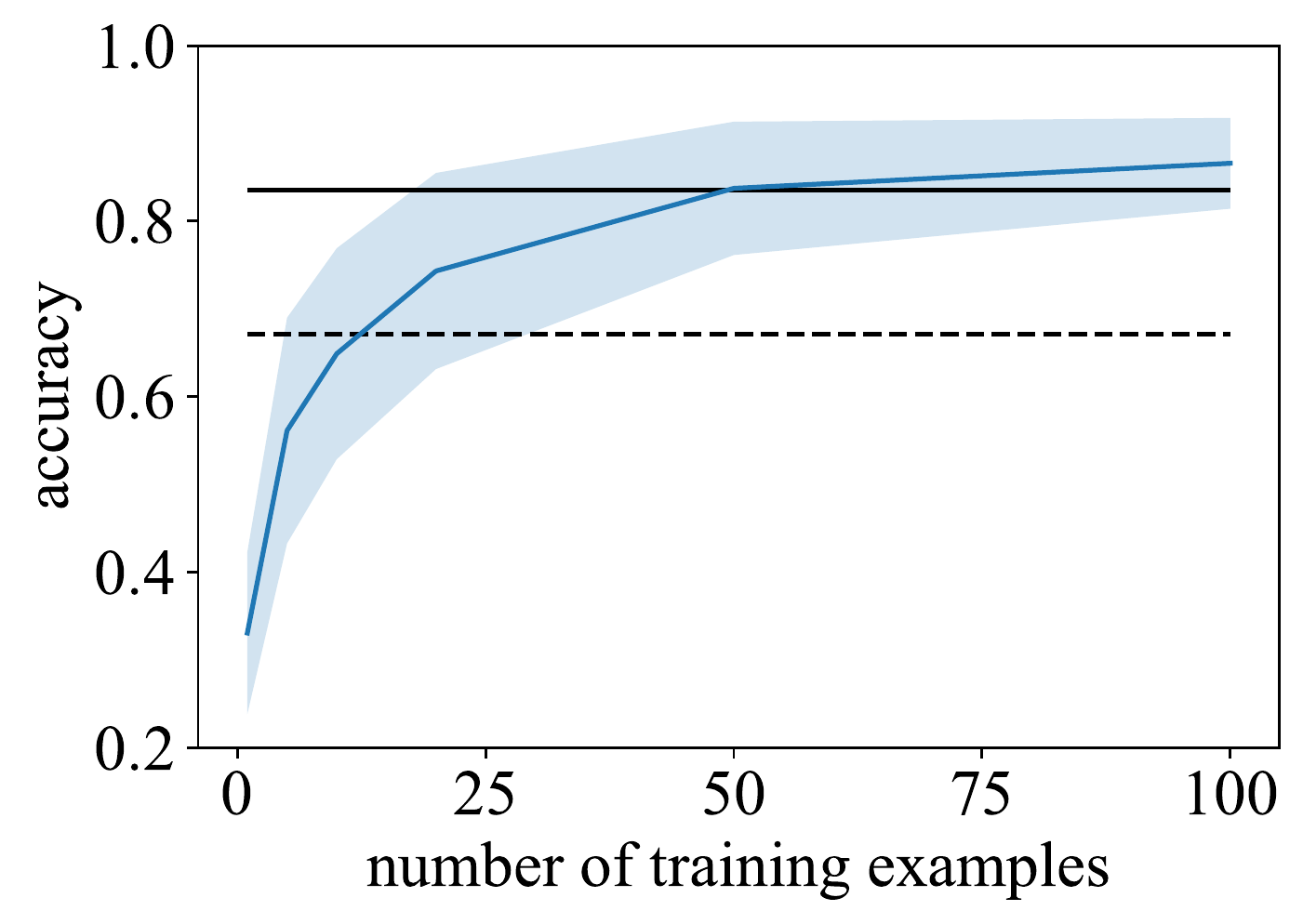}
    \vspace{-0.6cm} % account for pdf space
    \caption{FewRel}
  \end{subfigure}
  \begin{subfigure}[t]{.45\linewidth}
    \centering
    \includegraphics[width=\linewidth]{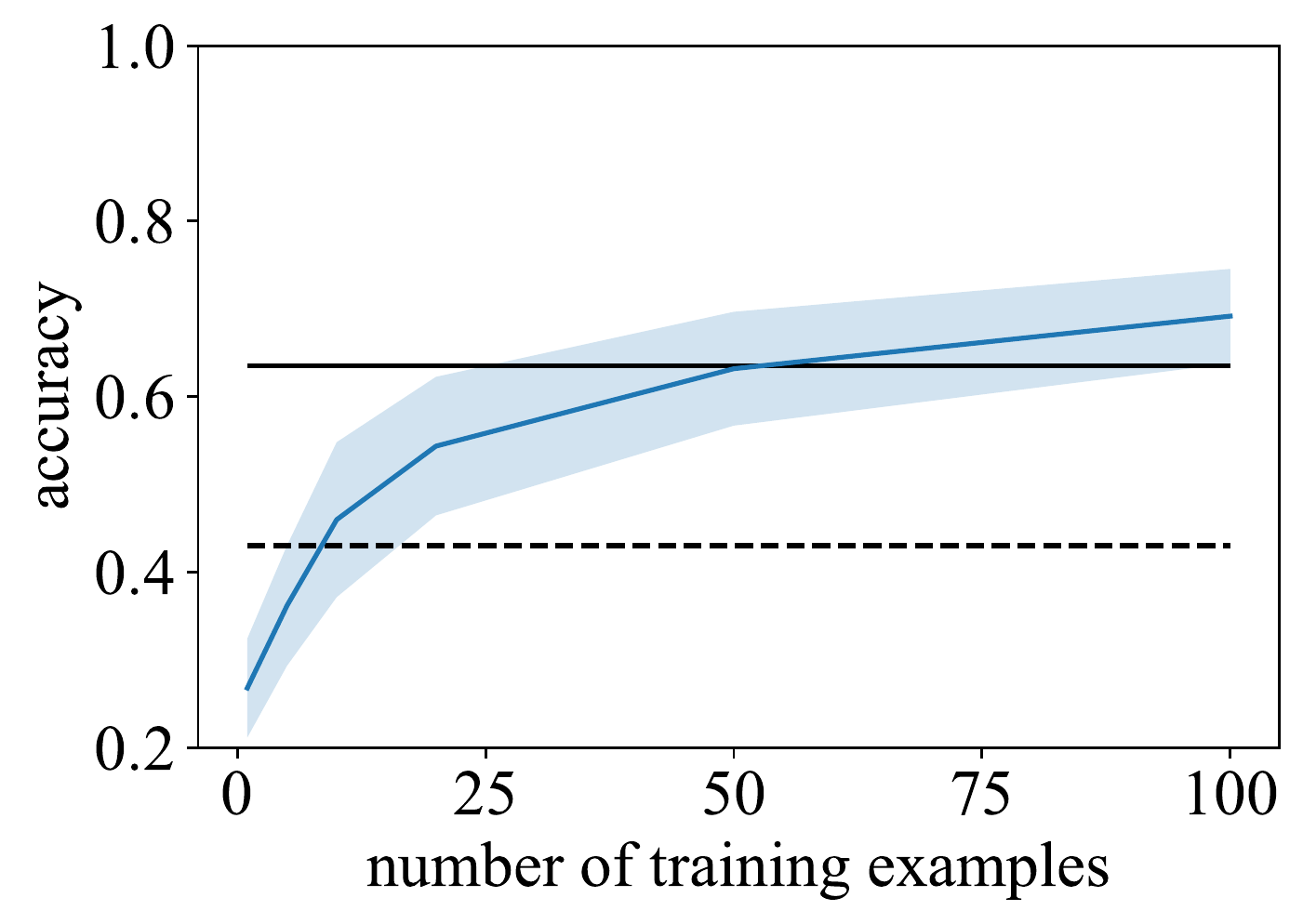}
    \vspace{-0.6cm} % account for pdf space
    \caption{HuffPost}
  \end{subfigure}
  \caption{Learning curve for fully-supervised CNN classifier vs \textsc{our}. Blue indicates CNN accuracy, with standard deviation shaded. Solid horizontal line is \textsc{our}'s 5-shot accuracy; dashed line is 1-shot accuracy. \textsc{our}'s 5-shot performance is competitive, especially when the total number of labeled examples is small.}\label{fig:fs_curve}
\end{figure}

To understand the extent of few-shot learning's utility,
we compare our results with that of a fully-supervised CNN classifier,
as we increase the number of training points.
From Figure~\ref{fig:fs_curve}, we see that our model's 5-shot accuracy is approximately equivalent to a fully-supervised classifier trained on 50 examples.

\subsection{Implementation Details}\label{app:baseline}
We detail the implementations of our baselines here. All baseline code are available in our repository.

\textbf{CNN}
For 1D convolution, we use filter windows of 3, 4, 5 with 50 feature maps each.
We applied $\mathrm{ReLU}$ after max-over-time pooling.

\textbf{Prototypical Network}
Prototypical network meta-learns
a multi-layer perceptron  to transform the input representation into
an embedding space that is suitable for few-shot classification.
If the input representation is learnable (e.g., \textsc{cnn}),
the parameters for the input representation are also updated
using meta-training.
In the experiments,
we use a MLP with one hidden layer and $\mathrm{ReLU}$ activation.
The dimensions of both the hidden layer and the output layer are 300.
We apply dropout with rate 0.1 to the hidden layer.

\textbf{MAML}
MAML meta-learns an initialization
such that the model can quickly adapt to new tasks after a few gradient steps.
For prediction on the input representation,
we use a MLP with one hidden layer of 300 $\mathrm{ReLU}$ units.
We apply dropout with rate 0.1 to the hidden layer.
During the MAML inner loop (adaptation stage),
we perform ten updates with step size 0.1
(we empirically found this outperform one-step MAML).
We backpropagate higher order gradients thoughout meta-training.
During the MAML outerloop,
we average the gradient across ten sampled tasks and
use Adam with learning rate $10^3$ to update the parameters.

\textbf{Finetune}
\citet{chen2018a} recently showed that
fine-tuning a properly pre-trained classifier
can achieve competitive performance when compared with
the state-of-the-art meta-learning.
Following their work,
we explicitly reduce the intra-class variation
during the pre-training stage.
Similar to MAML,
we use a MLP with one hidden layer (300 $\mathrm{ReLU}$ units)
to make predictions from the input representation (e.g., \textsc{cnn}).
During finetuning stage, we re-train the MLP from scratch
and fine-tune the learnable parameters of the input representation.
We stop fine-tuning once the gradient norm is less than $10^{-3}$.

\textbf{Induction network}
Induction network consists of three modules: encoder module, induction module and relation module.
The encoder module uses a biLSTM with self-attention to obtain a fix-length representation of each input example.
The induction module performs dynamic routing to compute the class-specific prototype.
The relation module uses a neural tensor layer to predict the relation association between each query example and the prototypes.
Following~\citet{geng2019few},
we set the hidden state size of biLSTM to 256 (128 for each direction) and the attention dimension to 64.
The iteration number in dynamic routing is set to $3$.
The dimension of the neural tensor layer is set to $100$.

\textbf{P-MAML}
P-MAML~\citep{zhang2019improving} has two phases: masked language model pretraining~\citep{devlin2018bert} and MAML~\citep{finn2017model}.
For pretraining, we used Hugging Face's language model finetuning code with default hyperparameters and BERT's pretrained base-uncased model. We applied early stopping when the validation perplexity failed to decrease for 2 epochs.
After pretraining, we add a softmax layer on top of the representation of the [CLS] token~\citep{devlin2018bert} to make predictions.
During the MAML inner loop (adaptation stage),
we perform ten updates with step size $10^{-3}$.
Following \citet{zhang2019improving},
we do not consider higher order gradients.
During the MAML outerloop,
we average the gradient across ten sampled tasks and use Adam with learning rate $10^{-5}$ to update the parameters.

\clearpage
\subsection{Results}\label{app:result}
This section contains experimental results with standard deviations.

\begin{table*}[h]
\small
\setlength{\tabcolsep}{1.5pt}
\begin{tabular}{llcccccc}
\toprule
Rep. &  Alg. &  20 News &  Amazon &  HuffPost &  RCV1 &  Reuters &  FewRel \\
\midrule
\textsc{avg} & \textsc{nn} & $33.95 $ {\tiny $ \pm 0.33$ } &  $46.76 $ {\tiny $ \pm 0.14$ } &  $31.45 $ {\tiny $ \pm 0.18$ } &  $43.76 $ {\tiny $ \pm 0.14$ } &   $56.48$ {\tiny $\pm 0.91$ } & $47.58 $ {\tiny $ \pm 0.38$ } \\
\textsc{idf} & \textsc{nn} & $38.88 $ {\tiny $ \pm 0.33$ } &  $51.43 $ {\tiny $ \pm 0.15$ } &  $31.53 $ {\tiny $ \pm 0.18$ } &  $41.96 $ {\tiny $ \pm 0.22$ } &   $57.76$ {\tiny $\pm 0.91$ } & $46.84 $ {\tiny $ \pm 0.31$ } \\
%\midrule
\textsc{cnn} &  \textsc{ft} &   $33.00 $ {\tiny $ \pm 0.74$ } &  $45.71 $ {\tiny $ \pm 0.86$ } &  $32.45 $ {\tiny $ \pm 0.54$ } &  $40.33 $ {\tiny $ \pm 1.47$ } &   $70.89$ {\tiny $\pm 4.30$ } & $54.08 $ {\tiny $ \pm 0.33$ } \\
\midrule
\textsc{avg} & \textsc{proto} & $36.25 $ {\tiny $ \pm 0.33$ } &  $37.26 $ {\tiny $ \pm 1.85$ } &  $35.68 $ {\tiny $ \pm 1.25$ } &  $28.48 $ {\tiny $ \pm 0.96$ } &   $59.54$ {\tiny $\pm 1.48$ } & $44.04 $ {\tiny $ \pm 0.80$ } \\
\textsc{idf} & \textsc{proto} & $37.86 $ {\tiny $ \pm 1.13$ } &  $41.91 $ {\tiny $ \pm 1.17$ } &  $34.88 $ {\tiny $ \pm 0.73$ } &  $32.14 $ {\tiny $ \pm 0.51$ } &   $61.00$ {\tiny $\pm 1.23$ } & $43.09 $ {\tiny $ \pm 1.15$ } \\
\textsc{cnn} & \textsc{proto} & $29.67 $ {\tiny $ \pm 1.02$ } &  $34.02 $ {\tiny $ \pm 1.48$ } &  $33.49 $ {\tiny $ \pm 0.79$ } &  $28.43 $ {\tiny $ \pm 0.68$ } &   $65.22$ {\tiny $\pm 1.52$ } & $49.78 $ {\tiny $ \pm 0.22$ } \\
\midrule
\textsc{avg} & \textsc{maml} &  $33.74$ {\tiny $\pm 0.32$ } &  $39.35$ {\tiny $\pm 1.38$ } &  $36.14$ {\tiny $\pm 1.23$ } &  $39.98$ {\tiny $\pm 1.83$ } &   $54.55$ {\tiny $\pm 1.08$ } & $43.83$ {\tiny $\pm 2.04$ } \\
\textsc{idf} & \textsc{maml} &  $37.26$ {\tiny $\pm 1.62$ } &  $43.63$ {\tiny $\pm 2.42$ } &  $38.95$ {\tiny $\pm 0.48$ } &  $42.58$ {\tiny $\pm 0.77$ } &   $61.46$ {\tiny $\pm 1.50$ } & $48.22$ {\tiny $\pm 1.20$ } \\
\textsc{cnn} & \textsc{maml} &  $28.98$ {\tiny $\pm 1.62$ } &  $35.30$ {\tiny $\pm 1.04$ } &  $34.12$ {\tiny $\pm 0.86$ } &  $39.03$ {\tiny $\pm 0.97$ } &   $66.62$ {\tiny $\pm 1.89$ } & $51.73$ {\tiny $\pm 3.98$ } \\
\midrule
\textsc{avg} & \textsc{rr} & $37.60 $ {\tiny $ \pm 0.10$ } &  $50.25 $ {\tiny $ \pm 0.23$ } &  $36.33 $ {\tiny $ \pm 0.36$ } &  $48.17 $ {\tiny $ \pm 0.17$ } &   $63.39$ {\tiny $\pm 0.95$ } & $53.25 $ {\tiny $ \pm 1.01$ } \\
\textsc{idf} & \textsc{rr} & $44.83 $ {\tiny $ \pm 1.07$ } &  $60.27 $ {\tiny $ \pm 1.33$ } &  $37.68 $ {\tiny $ \pm 0.96$ } &  $48.65 $ {\tiny $ \pm 0.57$ } &   $69.12$ {\tiny $\pm 1.87$ } & $55.65 $ {\tiny $ \pm 1.08$ } \\
\textsc{cnn} & \textsc{rr} & $32.25 $ {\tiny $ \pm 1.62$ } &  $37.30 $ {\tiny $ \pm 0.80$ } &  $37.32 $ {\tiny $ \pm 1.15$ } &  $41.81 $ {\tiny $ \pm 1.49$ } &   $71.40$ {\tiny $\pm 1.63$ } & $56.83 $ {\tiny $ \pm 2.30$ } \\
\midrule
\multicolumn{2}{l}{\textsc{our}} &   $\bm{52.17} $ {\tiny $ \pm 0.65$ } &  $\bm{62.66} $ {\tiny $ \pm 0.67$ } &  $\bm{43.09} $ {\tiny $ \pm 0.16$ } &  $\bm{54.15} $ {\tiny $ \pm 1.06$ } &  $\bm{81.81} $ {\tiny $ \pm 1.61$ } & $\bm{67.10} $ {\tiny $ \pm 0.93$ } \\ 
\midrule
\midrule
\multicolumn{2}{l}{\textsc{our}\text{ w/o }$\mathrm{t}(\cdot)$} &   $50.15 $ {\tiny $ \pm 1.62$ } &  $61.77 $ {\tiny $ \pm 0.73$ } &  $42.09 $ {\tiny $ \pm 0.37$ } &  $51.51 $ {\tiny $ \pm 0.75$ } &   $76.71$ {\tiny $\pm 1.44$ } & $66.93 $ {\tiny $ \pm 0.46$ } \\ 
\multicolumn{2}{l}{\textsc{our}\text{ w/o }$\mathrm{s}(\cdot)$} &   $41.99 $ {\tiny $ \pm 0.69$ } &  $51.12 $ {\tiny $ \pm 0.88$ } &  $40.16 $ {\tiny $ \pm 0.23$ } &  $48.59 $ {\tiny $ \pm 0.62$ } &   $78.15$ {\tiny $\pm 0.99$ } & $65.83 $ {\tiny $ \pm 0.33$ } \\ 
\multicolumn{2}{l}{\textsc{our}\text{ w/o }$\mathrm{biLSTM}$} &   $50.35 $ {\tiny $ \pm 0.73$ } &  $61.95 $ {\tiny $ \pm 0.40$ } &  $42.22 $ {\tiny $ \pm 0.66$ } &  $51.88 $ {\tiny $ \pm 0.84$ } & $77.17$ {\tiny $\pm 1.53$ } &  $66.42 $ {\tiny $ \pm 0.35$ } \\ 
\midrule
\multicolumn{2}{l}{\textsc{our}\text{ w }\textsc{ebd}} &   $39.68$ {\tiny $\pm 2.60$ } &  $56.48$ {\tiny $\pm 2.63$ } &  $40.64$ {\tiny $\pm 0.84$ } &  $48.64$ {\tiny $\pm 0.69$ } &   $81.68$ {\tiny $\pm 2.21$ } & $61.47$ {\tiny $\pm 1.89$ }\\ 
\bottomrule
\end{tabular}
\centering
\caption{5-way 1-shot classification.
The bottom four rows present our ablation study.}\label{tab:bigtable51}
\vspace{-2mm}
\end{table*}

\begin{table*}[h]
\small
\setlength{\tabcolsep}{1.5pt}
\begin{tabular}{llcccccc}
\toprule
Rep. & Alg. & 20 News & Amazon & HuffPost & RCV1 & Reuters & FewRel \\
\midrule
\textsc{avg} &\textsc{nn} &$45.87 $ {\tiny $ \pm 0.39$} & $60.36 $ {\tiny $ \pm 0.20$} & $41.55 $ {\tiny $ \pm 0.23$} & $60.84 $ {\tiny $ \pm 0.19$} &  $80.51$ {\tiny $\pm 0.66$ } & $60.67 $ {\tiny $ \pm 0.30$} \\
\textsc{idf} &\textsc{nn} &$51.94 $ {\tiny $ \pm 0.20$} & $67.15 $ {\tiny $ \pm 0.21$} & $42.35 $ {\tiny $ \pm 0.15$} & $58.27 $ {\tiny $ \pm 0.23$} &  $82.88$ {\tiny $\pm 0.57$ } & $60.62 $ {\tiny $ \pm 0.41$} \\
\textsc{cnn} & \textsc{ft} &  $47.17 $ {\tiny $ \pm 0.98$} & $63.91 $ {\tiny $ \pm 1.20$} & $44.13 $ {\tiny $ \pm 0.70$} & $62.34 $ {\tiny $ \pm 0.56$} &  $90.95$ {\tiny $\pm 3.59$ } & $71.19 $ {\tiny $ \pm 0.57$} \\
\midrule
\textsc{avg} &\textsc{proto} &$45.42 $ {\tiny $ \pm 1.36$} & $51.99 $ {\tiny $ \pm 2.72$} & $41.67 $ {\tiny $ \pm 1.13$} & $31.22 $ {\tiny $ \pm 1.53$} &  $68.15$ {\tiny $\pm 1.53$ } & $46.55 $ {\tiny $ \pm 1.58$} \\
\textsc{idf} &\textsc{proto} &$46.53 $ {\tiny $ \pm 1.44$} & $59.24 $ {\tiny $ \pm 1.10$} & $50.24 $ {\tiny $ \pm 0.94$} & $35.63 $ {\tiny $ \pm 0.83$} &  $72.07$ {\tiny $\pm 2.69$ } & $61.99 $ {\tiny $ \pm 1.82$} \\
\textsc{cnn} &\textsc{proto} &$35.09 $ {\tiny $ \pm 0.71$} & $44.49 $ {\tiny $ \pm 1.53$} & $44.21 $ {\tiny $ \pm 0.58$} & $29.33 $ {\tiny $ \pm 0.81$} &  $74.33$ {\tiny $\pm 1.86$ } & $65.16 $ {\tiny $ \pm 0.99$} \\
\midrule
\textsc{avg} & \textsc{maml} &  $43.92$ {\tiny $\pm 1.07$ } & $47.22$ {\tiny $\pm 1.99$ } & $49.69$ {\tiny $\pm 0.54$ } & $50.69$ {\tiny $\pm 1.03$ } &  $62.46$ {\tiny $\pm 0.58$ } & $57.87$ {\tiny $\pm 1.86$ } \\
\textsc{idf} & \textsc{maml} &  $48.62$ {\tiny $\pm 1.31$ } & $62.45$ {\tiny $\pm 1.33$ } & $53.70$ {\tiny $\pm 0.29$ } & $54.14$ {\tiny $\pm 0.72$ } &  $71.96$ {\tiny $\pm 1.48$ } & $65.80$ {\tiny $\pm 1.19$ } \\
\textsc{cnn} & \textsc{maml} & $36.79$ {\tiny $\pm 0.78$ } & $43.70$ {\tiny $\pm 1.38$ } & $45.89$ {\tiny $\pm 0.53$ } & $51.15$ {\tiny $\pm 0.39$ } &  $85.00$ {\tiny $\pm 0.76$ } & $66.90$ {\tiny $\pm 3.23$ } \\
\midrule
\textsc{avg} &\textsc{rr} &$57.24 $ {\tiny $ \pm 0.19$} & $72.78 $ {\tiny $ \pm 0.23$} & $54.86 $ {\tiny $ \pm 0.21$} & $72.62 $ {\tiny $ \pm 0.17$} &  $90.00$ {\tiny $\pm 0.49$ } & $72.22 $ {\tiny $ \pm 0.16$} \\
\textsc{idf} &\textsc{rr} &$64.35 $ {\tiny $ \pm 0.54$} & $79.78 $ {\tiny $ \pm 0.28$} & $59.56 $ {\tiny $ \pm 1.78$} & $72.85 $ {\tiny $ \pm 0.21$} &  $93.02$ {\tiny $\pm 0.45$ } & $75.30 $ {\tiny $ \pm 0.34$} \\
\textsc{cnn} &\textsc{rr} &$44.32 $ {\tiny $ \pm 0.44$} & $53.89 $ {\tiny $ \pm 1.54$} & $49.96 $ {\tiny $ \pm 0.20$} & $59.47 $ {\tiny $ \pm 0.58$} &  $87.93$ {\tiny $\pm 1.49$ } & $71.81 $ {\tiny $ \pm 1.25$} \\
\midrule
\textsc{our} & & $\bm{68.33} $ {\tiny $ \pm 0.17$} & $\bm{81.16} $ {\tiny $ \pm 0.31$} & $\bm{63.51} $ {\tiny $ \pm 0.10$} & $\bm{75.38} $ {\tiny $ \pm 1.12$} & $\bm{96.00} $ {\tiny $ \pm 0.27$} & $\bm{83.53} $ {\tiny $ \pm 0.27$} \\
\midrule
\midrule
\multicolumn{2}{l}{\textsc{our}\ \text{w/o}\ $\mathrm{t}(\cdot)$} &  $67.59 $ {\tiny $ \pm 0.57$} & $80.58 $ {\tiny $ \pm 0.19$} & $60.86 $ {\tiny $ \pm 0.19$} & $75.12 $ {\tiny $ \pm 0.90$} &  $93.71$ {\tiny $\pm 0.84$ } & $83.20 $ {\tiny $ \pm 0.39$} \\
\multicolumn{2}{l}{\textsc{our}\ \text{w/o}\ $\mathrm{s}(\cdot)$} &  $60.77 $ {\tiny $ \pm 0.52$} & $75.37 $ {\tiny $ \pm 0.27$} & $60.29 $ {\tiny $ \pm 0.35$} & $72.84 $ {\tiny $ \pm 0.23$} &  $94.79$ {\tiny $\pm 0.32$ } & $82.62 $ {\tiny $ \pm 0.20$} \\
\multicolumn{2}{l}{\textsc{our}\ \text{w/o}\ $\mathrm{biLSTM}$} &  $66.99 $ {\tiny $ \pm 0.32$} & $80.90 $ {\tiny $ \pm 0.36$} & $63.04 $ {\tiny $ \pm 0.20$} & $74.19 $ {\tiny $ \pm 0.22$} &  $95.36$ {\tiny $\pm 0.39$ } & $82.90 $ {\tiny $ \pm 0.21$} \\
\midrule
 \multicolumn{2}{l}{\textsc{our}\ \text{w}\ \textsc{ebd}}&  $57.52$ {\tiny $\pm 2.39$ } & $76.34$ {\tiny $\pm 0.65$ } & $58.58$ {\tiny $\pm 1.09$ } & $71.50$ {\tiny $\pm 0.28$ } &  $95.76$ {\tiny $\pm 0.53$ } & $80.88$ {\tiny $\pm 1.18$ }\\
\bottomrule
\end{tabular}
\centering
\caption{5-way 5-shot classification.
The bottom four rows present our ablation study.}\label{tab:bigtable55}
\end{table*}

\begin{table*}[t]
\small
\setlength{\tabcolsep}{4pt}
\begin{tabular}{cccccccccc}
\toprule
\emph{talk.politics.mideast}&\emph{sci.space}    &\emph{misc.forsale} &\emph{talk.politics.misc}&\emph{comp.graphics}\\
\midrule
israel       &space        &sale         &president    &image        \\
armenian     &nasa         &00           &cramer       &graphics     \\
turkish      &launch       &shipping     &mr           &jpeg         \\
armenians    &orbit        &offer        &stephanopoulos&images       \\
israeli      &shuttle      &condition    &people       &gif          \\
jews         &moon         &1st          &government   &format       \\
armenia      &henry        &price        &optilink     &file         \\
arab         &earth        &forsale      &myers        &3d           \\
people       &mission      &asking       &clayton      &ftp          \\
jewish       &solar        &comics       &gay          &color        \\
\bottomrule
\\
\toprule
\emph{sci.crypt}    &\emph{comp.windows.x}&\emph{comp.os.ms-windows.misc}&\emph{talk.politics.guns}&\emph{talk.religion.misc}\\
\midrule
key          &window       &ax           &gun          &god          \\
clipper      &xx           &max          &guns         &jesus        \\
encryption   &motif        &windows      &fbi          &sandvik      \\
chip         &server       &g9v          &firearms     &christian    \\
keys         &widget       &b8f          &atf          &bible        \\
security     &file         &a86          &batf         &jehovah      \\
privacy      &xterm        &145          &weapons      &christ       \\
government   &x11          &pl           &people       &lord         \\
escrow       &entry        &1d9          &waco         &kent         \\
des          &dos          &34u          &cdt          &brian        \\
\bottomrule
\\
\toprule
\emph{rec.autos}    &\emph{sci.med}      &\emph{comp.sys.mac.hardware}&\emph{sci.electronics}&\emph{rec.sport.hockey}\\
\midrule
car          &medical      &mac          &circuit      &hockey       \\
cars         &disease      &apple        &wire         &game         \\
engine       &msg          &centris      &ground       &team         \\
ford         &cancer       &quadra       &wiring       &nhl          \\
oil          &health       &lc           &voltage      &play         \\
dealer       &patients     &monitor      &battery      &season       \\
callison     &doctor       &duo          &copy         &games        \\
mustang      &hiv          &nubus        &amp          &25           \\
com          &food         &drive        &electronics  &ca           \\
autos        &diet         &simms        &audio        &pit          \\
\bottomrule
\\
\toprule
\emph{alt.atheism}
&\emph{rec.motorcycles}&\emph{comp.sys.ibm.pc.hardware}&\emph{rec.sport.baseball}&\emph{soc.religion.christian}\\
\midrule
god          &bike         &scsi         &baseball     &god          \\
atheism      &dod          &drive        &game         &church       \\
atheists     &ride         &ide          &year         &jesus        \\
keith        &bmw          &controller   &team         &christ       \\
livesey      &com          &card         &players      &sin          \\
morality     &riding       &bus          &games        &christians   \\
religion     &bikes        &drives       &hit          &christian    \\
moral        &motorcycle   &bios         &braves       &rutgers      \\
islamic      &dog          &disk         &runs         &bible        \\
say          &rider        &pc           &pitcher      &faith        \\
\bottomrule
\end{tabular}
\centering
\caption{Top 10 LMI-ranked words for each class in 20 Newsgroup.
  Class names are shown in italic.
Different class exhibit different salient features.}\label{tab:lmi_20news}
\end{table*}

\end{document}